\documentclass[transmag]{IEEEtran}
\usepackage{latexsym}
\usepackage{graphicx}
\usepackage{amsfonts,amssymb,amsmath}
\usepackage{hyperref}
\usepackage{multirow}
\usepackage{booktabs}
\usepackage{cite}
\usepackage{colortbl}
\usepackage[flushleft]{threeparttable}
\graphicspath{{figures/}}

\begin{document}

\title{RAPIQUE: Rapid and Accurate Video Quality Prediction of User Generated Content}

\author{Zhengzhong Tu, Xiangxu Yu, Yilin Wang, Neil Birkbeck, Balu Adsumilli, and Alan C. Bovik,
\IEEEmembership{Fellow, IEEE}
\thanks{Z. Tu, X. Yu, and A. C. Bovik are with Laboratory for Image and Video Engineering (LIVE), Department of Electrical and Computer Engineering, The University of Texas at Austin, Austin, TX, 78712, USA (emails: zhengzhong.tu@utexas.edu, yuxiangxu@utexas.edu, bovik@utexas.edu).}
\thanks{Y. Wang, N. Birkbeck, and B. Adsumilli are with YouTube Media Algorithms Team, Google LLC, Mountain View, CA, 94043, USA. (emails: yilin@google.com, birkbeck@google.com, 
badsumilli@google.com)}}

\IEEEtitleabstractindextext{\begin{abstract}Blind or no-reference video quality assessment of user-generated content (UGC) has become a trending, challenging, heretofore unsolved problem. Accurate and efficient video quality predictors suitable for this content are thus in great demand to achieve more intelligent analysis and processing of UGC videos. Previous studies have shown that natural scene statistics and deep learning features are both sufficient to capture spatial distortions, which contribute to a significant aspect of UGC video quality issues. However, these models are either incapable or inefficient for predicting the quality of complex and diverse UGC videos in practical applications. Here we introduce an effective and efficient video quality model for UGC content, which we dub the Rapid and Accurate Video Quality Evaluator (RAPIQUE), which we show performs comparably to state-of-the-art (SOTA) models but with orders-of-magnitude faster runtime. RAPIQUE combines and leverages the advantages of both quality-aware scene statistics features and semantics-aware deep convolutional features, allowing us to design the first general and efficient spatial and temporal (space-time) bandpass statistics model for video quality modeling. Our experimental results on recent large-scale UGC video quality databases show that RAPIQUE delivers top performances on all the datasets at a considerably lower computational expense. We hope this work promotes and inspires further efforts towards practical modeling of video quality problems for potential real-time and low-latency applications. To promote public usage, an implementation of RAPIQUE has been made freely available online: \url{https://github.com/vztu/RAPIQUE}.
\end{abstract}

}

\maketitle

\section{INTRODUCTION}
\label{sec:introduction}

Recent years have witnessed an explosion of user-generated content (UGC) captured and streamed over social media platforms such as YouTube, Facebook, TikTok, and Twitter. Thus, there is a great need to understand and analyze billions of these shared contents to optimize video pipelines of efficient UGC data storage, processing, and streaming. UGC videos, which are typically created by amateur videographers, often suffer from unsatisfactory perceptual quality, arising from imperfect capture devices, uncertain shooting skills, and a variety of possible content processes, as well as compression and streaming distortions. In this regard, predicting UGC video quality is much more challenging than assessing the quality of synthetically distorted videos in traditional video databases. UGC distortions are more diverse, complicated, commingled, and no ``pristine'' reference is available.

Traditional video quality assessment (VQA) models have been widely studied \cite{seshadrinathan2010study} as an increasingly important toolset used by the streaming and social media industries. While full-reference (FR) VQA research is gradually maturing and several algorithms \cite{wang2004image, li2016toward} are quite widely deployed, recent attention has shifted more towards creating better no-reference (NR) VQA models that can be used to predict and monitor the quality of authentically distorted UGC videos. One intriguing property of UGC videos, from the data compression aspect, is that the original videos to be compressed often already suffer from artifacts or distortions, making it difficult to decide the compression settings \cite{yu2020predicting}. Similarly, it is of great interest to be able to deploy flexible video transcoding profiles in industry-level applications based on measurements of input video quality to achieve even better rate-quality tradeoffs relative to traditional encoding paradigms \cite{Wang2020}. The decision tuning strategy of such an adaptive encoding scheme, however, would require the guidance of an accurate and efficient NR or blind video quality (BVQA) model suitable for UGC \cite{tu2020ugc}.

Many blind video quality models have been proposed to solve the UGC-VQA problem \cite{moorthy2011blind, mittal2012no, saad2014blind, kundu2017no,xue2014blind, ghadiyaram2017perceptual, korhonen2019two, ye2012unsupervised, pei2015image, tu2020ugc, ebenezer2020no, yu2020predicting, ying2019patches, ying2020patch, li2019quality}. Among these, BRISQUE \cite{mittal2012no}, GM-LOG \cite{xue2014blind}, FRIQUEE \cite{ghadiyaram2017perceptual}, V-BLIINDS \cite{saad2014blind}, and VIDEVAL \cite{tu2020ugc} have leveraged different sets of natural scene statistics (NSS)-based quality-aware features, using them to train shallow regressors to predict subjective quality scores. Another well-founded approach is to design a large number of distortion-specific features, whether individually \cite{feng2006measurement, tu2020bband, norkin2018film}, or combined, as is done in TLVQM \cite{korhonen2019two} to achieve a final quality prediction score.  Recently, convolutional neural networks (CNN) have been shown to deliver remarkable performance on a wide range of computer vision tasks \cite{chen2020proxiqa, wang2019going, chen2020learning}. Several deep CNN-based BVQA models have also been proposed \cite{kim2018deep, zhang2018blind, li2019quality, ying2020patch} by training them on recently created large-scale psychometric databases \cite{lin2018koniq, ghadiyaram2015massive}. These methods have yielded promising results on synthetic distortion datasets \cite{seshadrinathan2010study}, but still struggled on UGC quality assessment databases \cite{sinno2018large, hosu2017konstanz, wang2019youtube}.

Prior work has mainly focused on spatial distortions, which have been shown to indeed play a critical role in UGC video quality prediction \cite{tu2020ugc}. The exploration of the temporal statistics of natural videos, however, has been relatively limited. The authors of \cite{tu2020ugc} have shown that temporal- or motion-related features are essential components when analyzing the quality of mobile captured videos, as exemplified by those in the LIVE-VQC database \cite{sinno2018large}. Yet, previous BVQA models that account for temporal distortions, such as V-BLIINDS and TLVQM, generally involve expensive motion estimation models, which are not practical in many scenarios. Furthermore, while compute-efficient VQA models exist, simple BVQA models like BRISQUE \cite{mittal2012no}, NIQE \cite{mittal2012making}, GM-LOG \cite{xue2014blind} are incapable of capturing complex distortions that arise in UGC videos. Complex models like V-BLIINDS \cite{saad2014blind}, TLVQM \cite{korhonen2019two}, and VIDEVAL \cite{tu2020ugc}, on the contrary, perform well on existing UGC video databases, but are much less efficient, since they either involve intensive motion-estimation algorithms or complicated scene statistics features. A recent deep learning model, VSFA \cite{li2019quality}, which extracts frame-level ResNet-50 \cite{he2016deep} features followed by training a GRU layer, is also less practical due to the use of full-size, frame-wise image inputs and the recurrent layers. 


We have made recent progress towards efficient modeling of temporal statistics relevant to the video quality problem, by exploiting and combining spatial and temporal scene statistics, as well as deep spatial features of natural videos. We summarize our contributions as follows:

\begin{itemize}
    \item We created a rapid and accurate video quality predictor, called RAPIQUE, in an efficient manner, achieving superior performance that is comparable or better than state-of-the-art (SOTA) models, but with a relative 20x speedup on 1080p videos. The runtime of RAPIQUE also scales well as a function of video resolution, and is 60x faster than the SOTA model VIDEVAL on 4k videos.
    \item We built a first-of-its-kind BVQA model that combines novel, effective, and easily computed low-level scene statistics features with high-level deep learning features. Aggressive spatial and temporal sampling strategies are used, exploiting content and distortion redundancies, to increase efficiency without sacrificing performance.
    \item We created a new spatial NSS feature extraction module within RAPIQUE, which is a highly efficient and effective alternative to the popular but expensive feature-based BIQA model, FRIQUEE. The spatial NSS features used in RAPIQUE are suitable for inclusion as basic elements of a variety of perceptual transforms, leading to significant efficiencies which might also be useful when developing future BVQA models.
    \item We designed \textit{the first} general, effective and efficient temporal statistics model (beyond frame-differences) that is based on bandpass regularities of natural videos, and which can also be used as a standalone module to boost existing BVQA methods on temporally-distorted or motion-intensive videos.
\end{itemize}

The rest of this paper is organized as follows. Section \ref{sec:related_work} reviews previous literature relating to video quality assessment models, while Section \ref{sec:rapique} unfolds the details of the RAPIQUE model. Experimental results and concluding remarks are given in Section \ref{sec:experiments} and Section \ref{conclusion}, respectively.

\section{Related Work}
\label{sec:related_work}

\subsection{Traditional BVQA Models}
\label{ssec:traditiona_bvqa_models}

Many early BVQA/BIQA models have been `distortion specific' in that they were designed to quantify a specific type of distortion such as blockiness \cite{wang2000blind}, blur \cite{marziliano2002no}, ringing \cite{ feng2006measurement}, banding \cite{ wang2016perceptual, tu2020bband, tu2020adaptive}, or noise \cite{amer2005fast, norkin2018film} in compressed images and videos. Recent high-performing BIQA/BVQA models are almost exclusively learning-based, operating by training sets of generic quality-aware features, which are combined to conduct quality predictions \cite{ moorthy2011blind, mittal2012no, saad2014blind, kundu2017no, ghadiyaram2017perceptual, korhonen2019two, ye2012unsupervised, pei2015image}. Learning-based BVQA models are more versatile and generalizable than `distortion specific' models, in that the selected features are broadly perceptually relevant, while powerful regression models can adaptively map the features onto quality scores learned from the data in the context of a specific application.

The most popular BVQA algorithms deploy perceptually relevant, low-level features based on simple, yet highly regular parametric bandpass models of scene statistics \cite{ruderman1994statistics}. These natural scene statistics (NSS) models often are predictably altered by the presence of distortions \cite{sheikh2006image}, although they have more limited power to characterize complex, commingled distortions. Successful picture quality models of this type have been explored in the wavelet (BIQI \cite{moorthy2010two}, DIIVINE \cite{moorthy2011blind}, C-DIIVINE \cite{zhang2014c}), DCT (BLIINDS \cite{saad2010dct}, BLIINDS-II \cite{saad2012blind}) and spatial domains (NIQE \cite{mittal2012making}, BRISQUE \cite{mittal2012no}), and have been further extended to encompass natural bandpass space-time video statistics models \cite{li2016spatiotemporal,mittal2015completely,saad2014blind, sinno2019spatio}, among which the most well-known model is Video-BLIINDS \cite{saad2014blind}. Other extensions of empirical NSS include models of the joint statistics of the gradient magnitude and Laplacian of Gaussian (GM-LOG \cite{xue2014blind}), in log-derivative and log-Gabor spaces (DESIQUE \cite{zhang2013no}), as well as in the gradient domain of LAB color transforms (HIGRADE \cite{kundu2017no}). The FRIQUEE model \cite{ghadiyaram2017perceptual} achieves excellent performance on UGC/consumer video/picture databases  \cite{ghadiyaram2015massive, nuutinen2016cvd2014, hosu2017konstanz, hosu2020koniq, wang2019youtube} by leveraging a bag of NSS features drawn from diverse color spaces and perceptually motivated transform domains.

Time-domain behavior is the key attribute that differentiates videos from still pictures. The perception of video correlates highly with motion and temporal change \cite{born2005structure}. Amongst BVQA models, Video-BLIINDS \cite{saad2014blind} was the first to explore the use of (spatio-) temporal scene statistics of video using DCT coefficient statistics in the time-differenced domain. V-BLIINDS also involves calculating motion coherence and global motion features, which requires expensive motion estimation, to account for temporal masking effects. 

\begin{figure*}[!ht]
\centering
\includegraphics[width=0.98\textwidth]{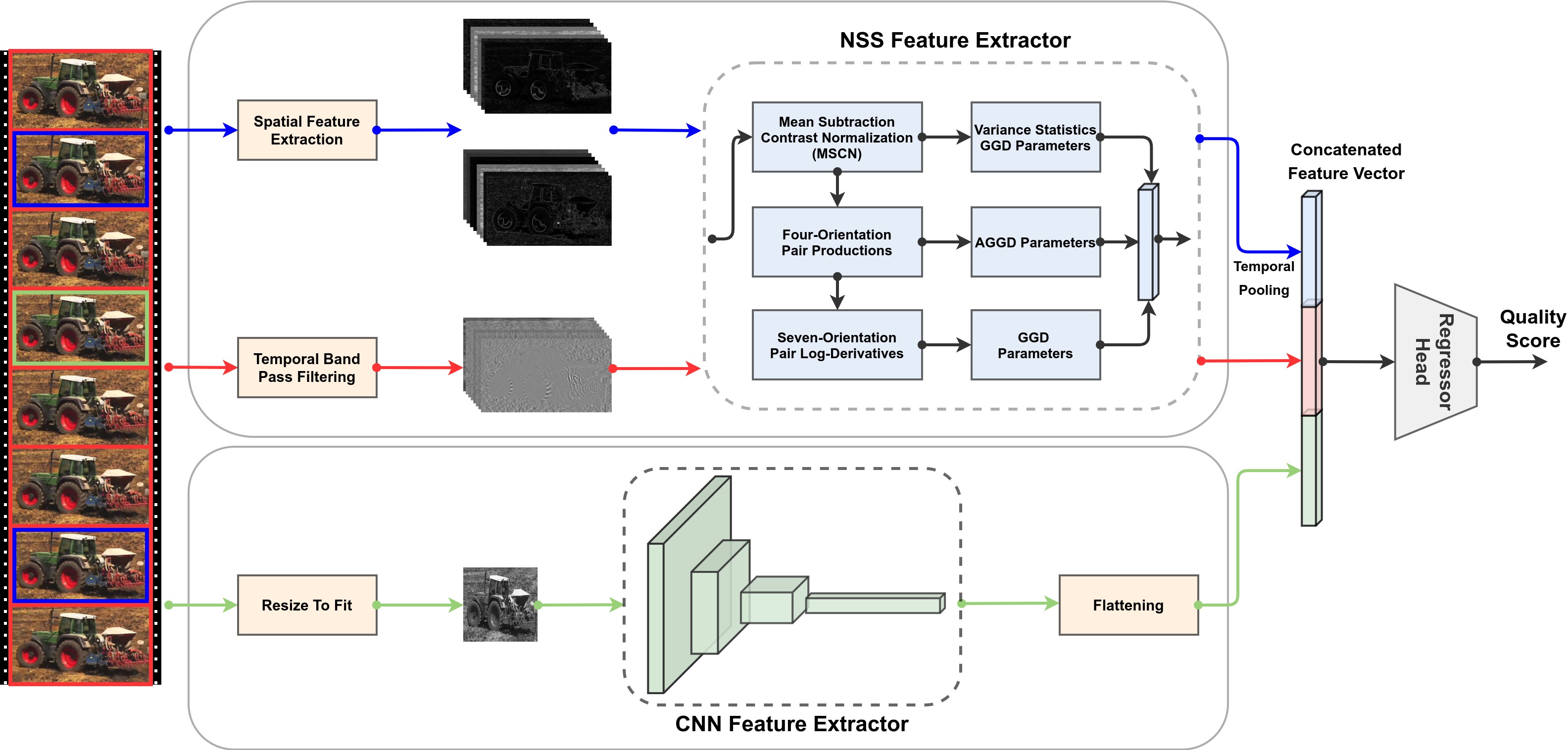}
\caption{Schematic overview of the proposed RAPIQUE model. Top block shows the spatial and temporal NSS feature extraction branch, while bottom block depicts the CNN feature extraction flow. The final feature vector is simply concatenated from the extracted spatial and temporal NSS and the CNN features, which is further used to train a regressor head.}
\label{fig:rapique_flowchart}
\end{figure*}

Instead of using DCT transforms, Mittal \textit{et al.} proposed a completely blind model called VIIDEO \cite{mittal2015completely}, which inspects the divisively normalized spatial statistics of frame differences. 
Bandpass filtering followed by divisive normalization was applied to frame differences, after which the inter-subband correlations are modeled over the temporal variation of the extracted generalized Gaussian parameters. 
As a highly experimental temporal-only model, VIIDEO includes no spatial features, hence does not perform well on natural UGC video datasets \cite{hosu2017konstanz, sinno2018large}.

Regarding the joint modeling of spatiotemporal statistics, Li \textit{et al.} proposed to adopt 3D-DCT transforms of local space-time regions from videos to extract quality-aware features \cite{li2016spatiotemporal}. More recently, the authors of \cite{dendi2020no} leveraged 3D divisive normalization transformed (DNT) and spatiotemporal Gabor-filtered responses of 3D-DNT coefficients of natural videos. The 3D transforms adopted therein, however, are too expensive for practical use; neither have these models been observed to perform well on UGC datasets \cite{sinno2018large, wang2019youtube}.

Another intriguing and more practical approach to integrating temporal features into BVQA models is to design separable spatial-temporal statistics \cite{sinno2019spatio, yu2020predicting, madhusudana2020st, chen2020chroma}. Spatial features can be modified to capture temporal effects within BIQA models like BRISQUE, whereby simple frame-differences or spatially displaced frame-differences are deployed \cite{sinno2019spatio,lee2020ipas, yu2020predicting, lee2020josa}.

A very recent feature-based BVQA model called TLVQM \cite{korhonen2019two} uses a two-level feature extraction mechanism to achieve efficient computation of a set of impairment/distortion-relevant features. Unlike NSS-based models, TLVQM relies on a comprehensive set of highly crafted features that measure motion, specific distortion artifacts, and aesthetics. TLVQM requires that a large number of parameters be specified by the user, which may affect its general performance on datasets or scenarios it has not been exposed to. The model currently achieves very good performance on natural video quality databases at a reasonable complexity.

VIDEVAL \cite{tu2020ugc} is currently the SOTA feature-based BVQA model on recent large-scale video dataset like KoNViD-1k \cite{hosu2017konstanz} and YouTube-UGC \cite{wang2019youtube}. It employs feature selection and fusion on top of efficacious NSS-based models as well as distortion-based features. It is also a very compact model as it only utilizes 60 features. However, it has not been observed to efficiently scale to high-resolution and high-framerate videos.

\subsection{Deep Learning-Based BVQA Models}
\label{ssec:deep_learning_bvqa_models}

Deep convolutional neural networks (CNNs) have been shown to deliver standout performance in a wide variety of low-level computer vision applications \cite{chen2020proxiqa, chen2020learning, ying2019patches, jiang2019enlightengan}. Recently, the release of several large-scale psychometric visual quality databases \cite{ghadiyaram2015massive, hosu2017konstanz, sinno2018large, hosu2020koniq, wang2019youtube} have sped the application of deep CNNs to perceptual video and image quality modeling. To conquer the limits of small data size, researchers have either proposed to conduct patch-wise data-augmentation during training \cite{kang2014convolutional, bosse2016deep, kim2017deep}, or to pretrain deep nets on larger visual sets like ImageNet \cite{deng2009imagenet}, then fine tune on target quality databases. Several authors report remarkable performance on synthetic distortion databases \cite{sheikh2006statistical,ponomarenko2013color} or on naturally distorted databases \cite{ghadiyaram2015massive, hosu2020koniq}.

Deep CNN models have also been employed for natural video quality
prediction. Kim \textit{et al.} \cite{kim2018deep} proposed a deep video quality assessor (DeepVQA) to learn spatio-temporal visual sensitivity maps via a deep CNN and a convolutional aggregation network. The V-MEON model \cite{liu2018end} leveraged a multi-task CNN framework which jointly optimizes a 3D-CNN for feature extraction and a codec classifier, and using fully-connected layers to predict video quality. Zhang \textit{et al.} \cite{zhang2018blind} leveraged transfer learning to develop a general-purpose BVQA framework based on weakly supervised learning and a resampling strategy. In the VSFA model \cite{li2019quality}, the authors applied a pre-trained image classification CNN as a deep feature extractor, then integrated the frame-wise deep features using a gated recurrent unit and a subjectively-inspired temporal pooling layer, reporting leading performance on several natural video databases \cite{nuutinen2016cvd2014, hosu2017konstanz, ghadiyaram2017capture}. The authors then built an enhanced version of VSFA, dubbed MDVSFA \cite{li2020unified}, by employing a mixed datasets training strategy on top, training a single VQA model on multiple datasets, and reporting superior performance on publicly available datasets. Several other popular CNN-based BVQA models \cite{kim2018deep, liu2018end, zhang2018blind, li2019quality, li2020unified} produce accurate quality predictions on legacy (single synthetic distortion) video datasets \cite{seshadrinathan2010study, vu2014vis3}, but struggle on recent in-the-wild UGC databases \cite{nuutinen2016cvd2014, ghadiyaram2017capture,hosu2017konstanz}.

\begin{figure}[!t]
\centering
\footnotesize
\def\xheight{0.485}
\def\xwidth{0.36}
\setlength{\tabcolsep}{1pt}
\begin{tabular}{cc}
\includegraphics[width=\xheight\linewidth]{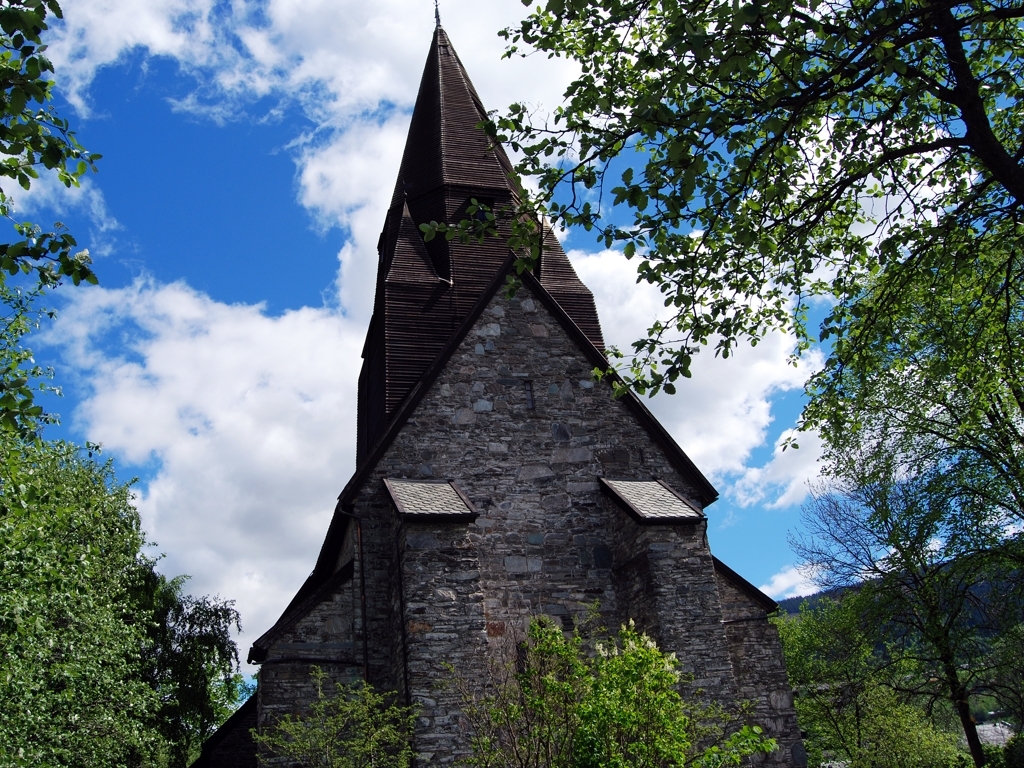} &
\includegraphics[width=\xheight\linewidth]{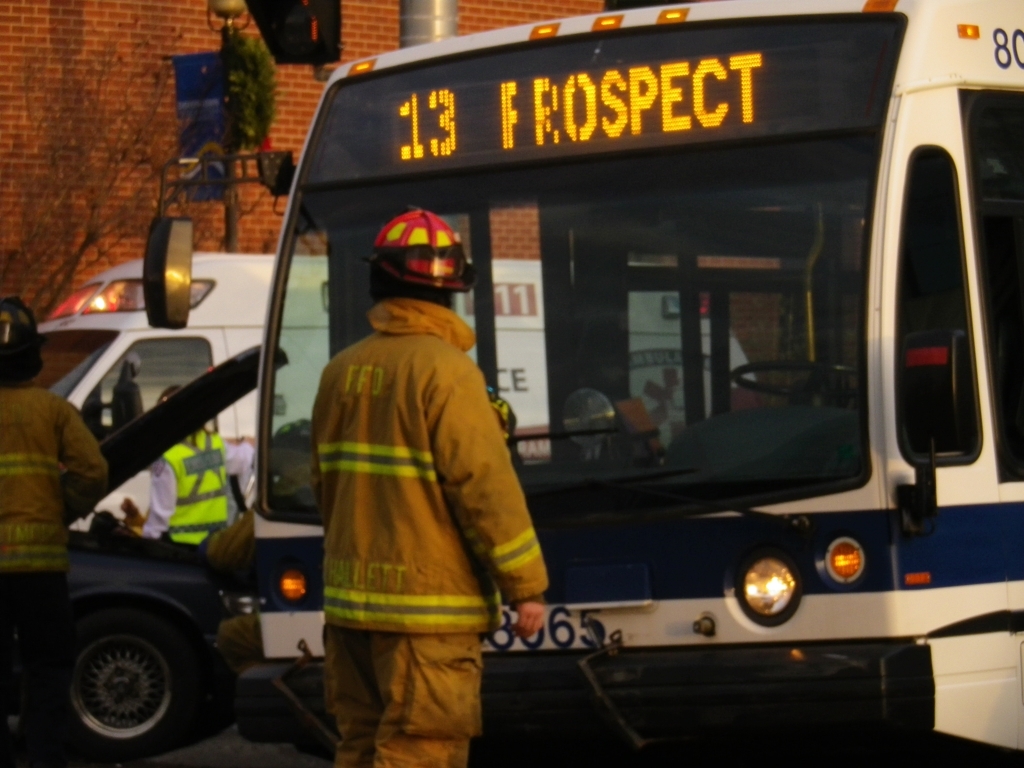} \\
(a) img1 (MOS=3.82, best quality) & (b) img2 (MOS=2.93, good quality) \\
\includegraphics[width=\xheight\linewidth,height=\xwidth\linewidth]{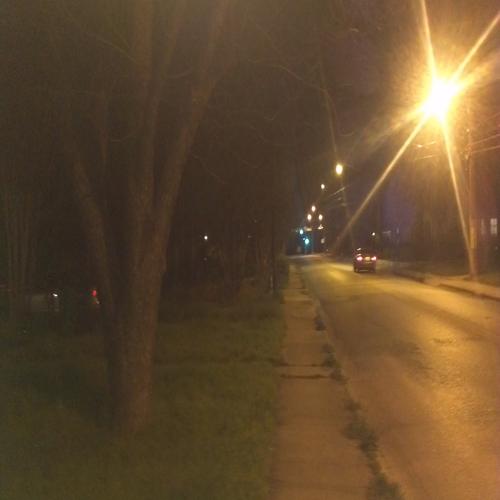} &
\includegraphics[width=\xheight\linewidth,height=\xwidth\linewidth]{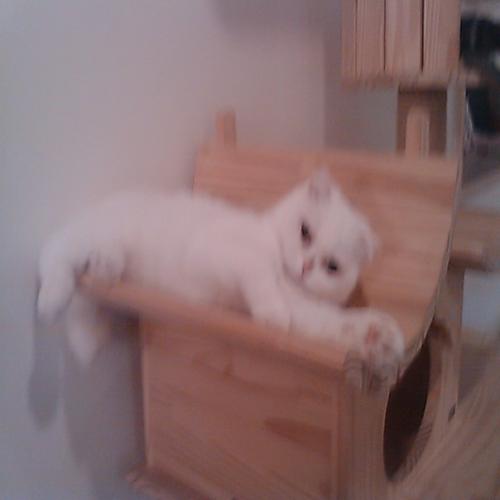} \\
(c) img3 (MOS=35.8, bad quality) & (d) img4 (MOS=15.9, worst quality) \\
\end{tabular}
\caption{Exemplar test images exhibiting four categories of quality: (a) img1 (best) and (b) img2 (good) are two good-quality images from KonIQ-10k (MOS range: $[1,5]$) \cite{hosu2020koniq}, while (c) img3 (bad) and (d) img4 (worst) are two bad-quality pictures from CLIVE (MOS range: $[0,100]$) \cite{ghadiyaram2015massive}.}
\label{fig:eg_imgs}
\end{figure}

\section{Rapid and Accurate Video Quality Evaluator (RAPIQUE)}
\label{sec:rapique}

Prior statistics-based video quality models have been shown to be capable of capturing complex UGC distortions, such as FRIQUEE \cite{ghadiyaram2017perceptual}, TLVQM \cite{korhonen2019two}, and VIDEVAL \cite{tu2020ugc}. However, these models are subject to time-consuming executions since either expensive motion estimation or high-order statistical features are required. CNN models are able to efficiently capture high-level features, which have also been observed to be useful quality indicators \cite{li2019quality}, albeit directly applying a CNN on high-resolution video frames is expensive. Here we propose an efficient two-branch framework, as depicted in Fig. \ref{fig:rapique_flowchart}, which combines quality-aware, low-level NSS features with high-level, semantics-aware CNN features. The NSS features operate on higher-resolution spatial and temporal bandpass feature maps, while the CNN feature extractor is applied on a resized low-resolution frames for practical considerations. We also adopt a sparse frame sampling strategy when extracting features, which further accelerates the runtime. We present the details of RAPIQUE in the following.

\begin{figure}[!t]
\centering
\footnotesize
\def\xheight{0.485}
\setlength{\tabcolsep}{1pt}
\begin{tabular}{cc}
\includegraphics[width=\xheight\linewidth]{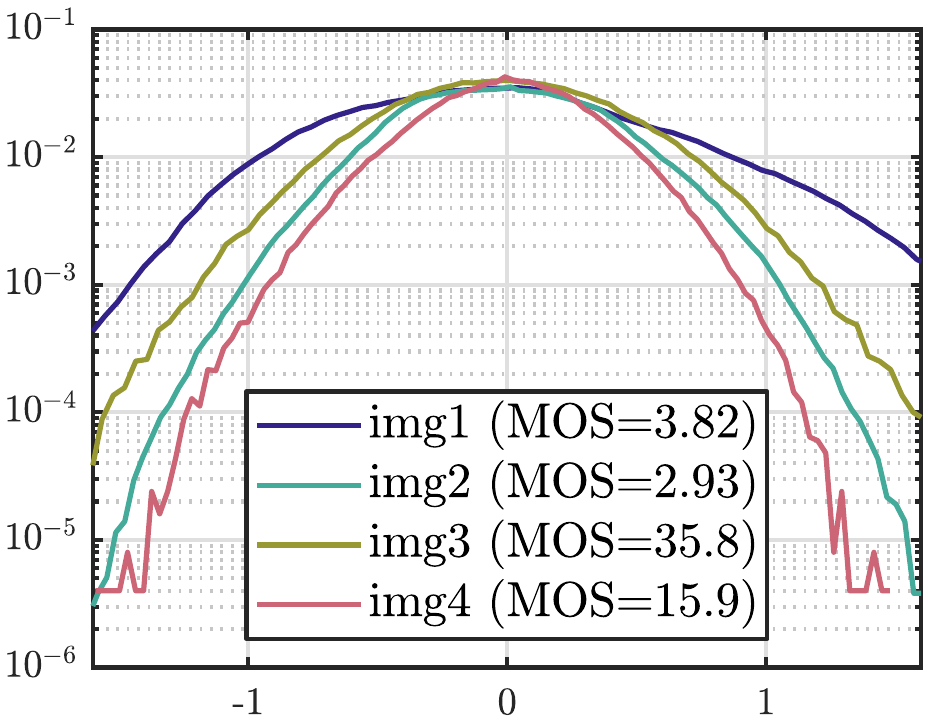} &
\includegraphics[width=\xheight\linewidth]{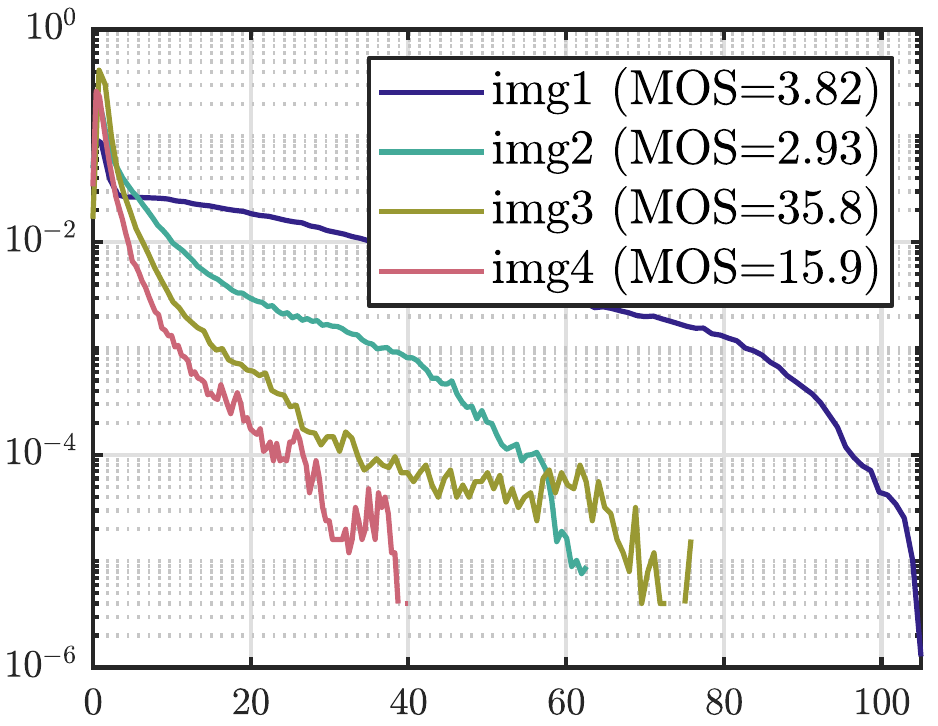} \\
\end{tabular}
\caption{Histograms of MSCN (left) and variance map (right) of the four images shown in Fig. \ref{fig:eg_imgs}.}
\label{fig:mscn_sigma}
\end{figure}

\begin{figure}[!t]
\centering
\footnotesize
\def\xheight{0.485}
\setlength{\tabcolsep}{1pt}
\begin{tabular}{cc}
\includegraphics[width=\xheight\linewidth]{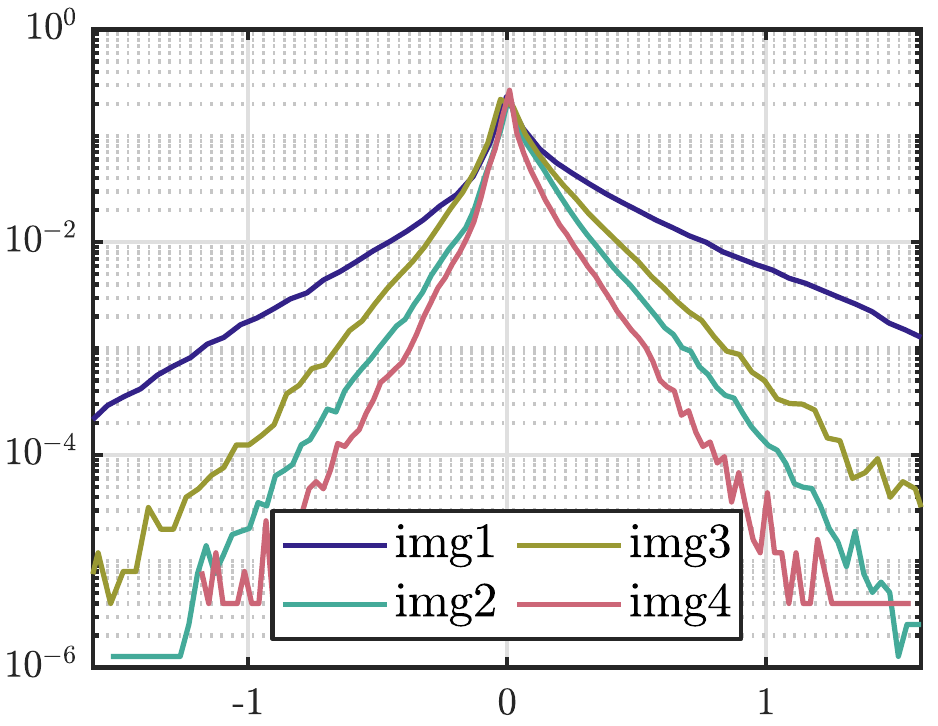} &
\includegraphics[width=\xheight\linewidth]{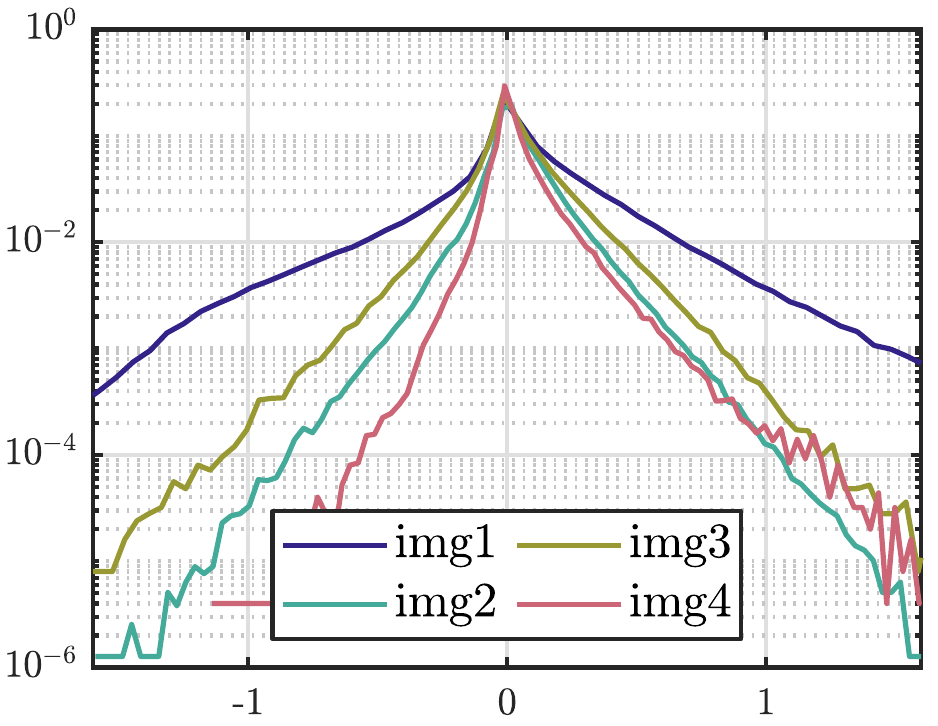} \\
\includegraphics[width=\xheight\linewidth]{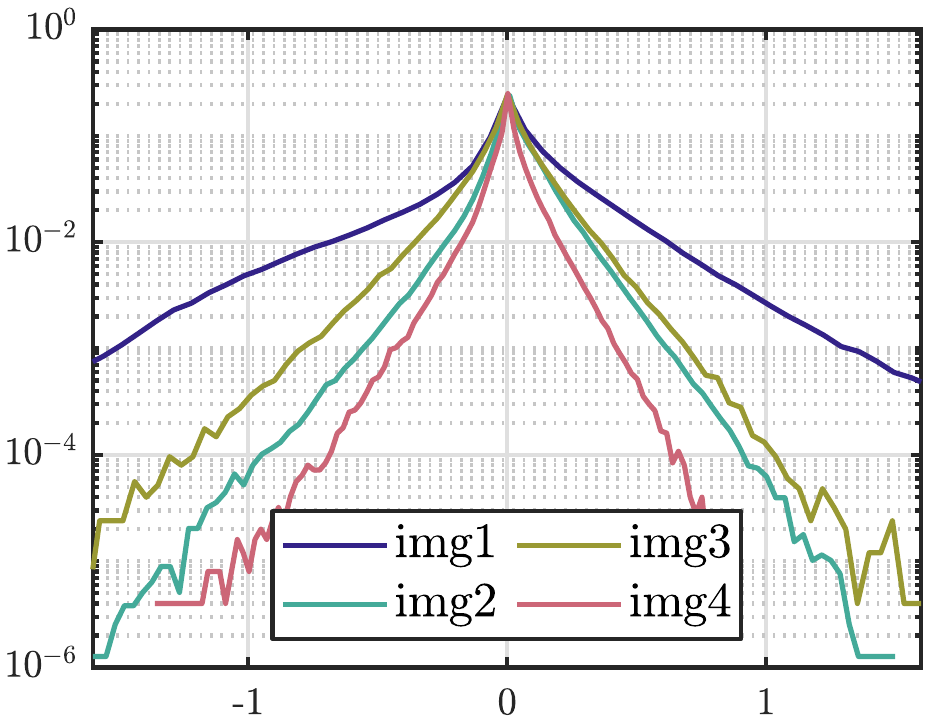} &
\includegraphics[width=\xheight\linewidth]{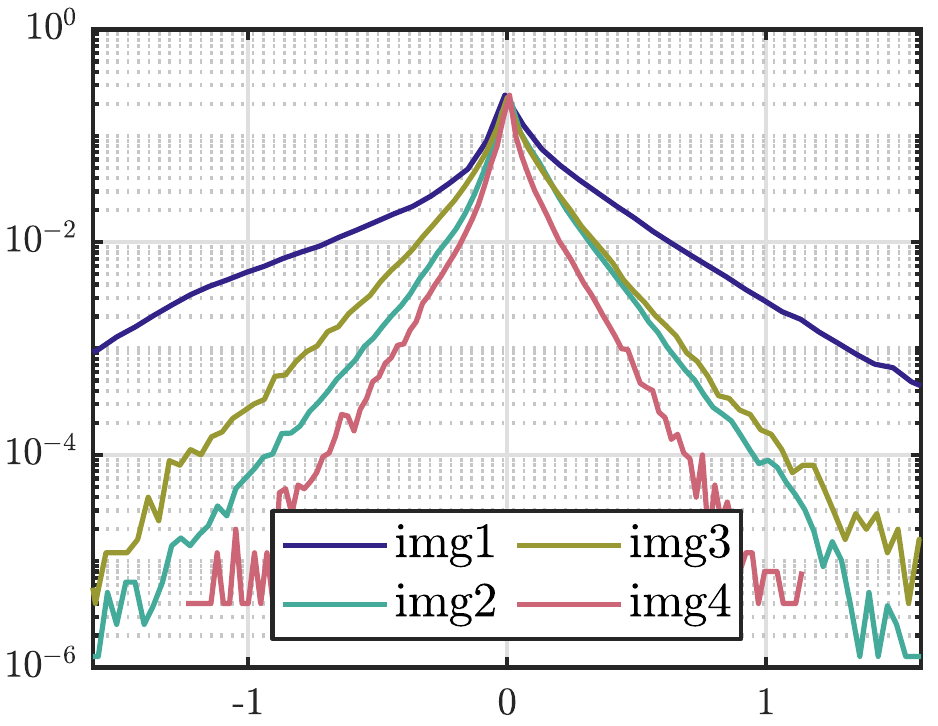} \\
\end{tabular}
\caption{Histograms of four-orientation (H, V, D1, D2) MSCN pair-production of the four images shown in Fig. \ref{fig:eg_imgs}.}
\label{fig:pair_prod}
\end{figure}

\subsection{Natural Scene Statistics}
\label{ssec:nss_model}

It has been observed that the spatial wavelet/subband coefficients of natural images exhibit strong regularities (Gaussianity) following a divisive normalization transform (DNT) \cite{moorthy2011blind}. 
A simple but effective form of divisive normalization, called mean subtraction and contrast normalization (MSCN), has been observed to accurately characterize image naturalness in multiple feature transforms \cite{mittal2012no, kundu2017no, ghadiyaram2017perceptual, zhang2013no}. We develop NSS-based features following the methodology of FRIQUEE \cite{ghadiyaram2017perceptual}, which leverages multiple perceptually-relevant feature transforms to extract a large number of statistical features. RAPIQUE uses simple yet effective low-order bandpass statistics, achieving comparable performance as the complex and time-consuming high-order features used in FRIQUEE. We were inspired by the successful and efficient basic features developed as products of spatially-adjacent MSCN responses, and log-derivative statistics in BRISQUE \cite{mittal2012no} and DESIQUE \cite{zhang2013no}, respectively. Specifically, let $Y(i,j)$ be a given intensity image or a transformed feature map. The MSCN operator is applied on $Y(i,j)$ to further decorrelate and Gaussianize the local pixels:
\begin{equation}
\label{eq:mscn}   
\hat{Y}(i,j)=\frac{Y(i,j)-\mu(i,j)}{\sigma(i,j)+C},
\end{equation}
where, $(i,j)$ are spatial indices and $C=1$ is a constant that prevents instabilities caused by having a small variance in the denominators. The factors $\mu(i,j)$ and $\sigma(i,j)$ are the weighted local mean and standard deviation within a spatial window centered at location $(i,j)$ calculated by:
\begin{equation}
\label{eq:mu}   
\mu(i,j)=\sum_{k=-K}^K\sum_{\ell=-L}^Lw_{k,l}Y(i-k,j-\ell)
\end{equation}
\begin{equation}
\label{eq:sigma}
\sigma(i,j)\!=\!\sqrt{\sum_{k=-K}^K\!\sum_{\ell=-L}^Lw_{k,\ell}[Y(i-k,j-\ell)-\mu(i,j)]^2},
\end{equation}
where $\boldsymbol{w}=\{w_{k,\ell}|k=-K,...,K,\ell=-L,...,L\}$ is a 2D isotropic, truncated, unit-volume Gaussian weighting function. We used $K=L=3$ in our implementations.

\begin{figure}[!t]
\centering
\footnotesize
\def\xheight{0.485}
\setlength{\tabcolsep}{1pt}
\begin{tabular}{cc}
\includegraphics[width=\xheight\linewidth]{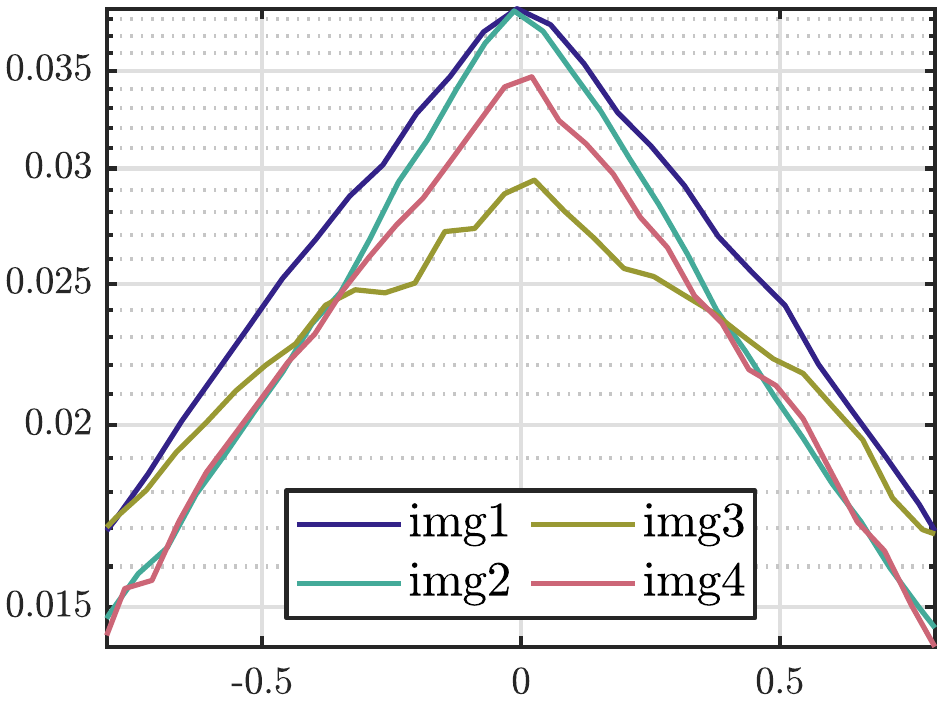} &
\includegraphics[width=\xheight\linewidth]{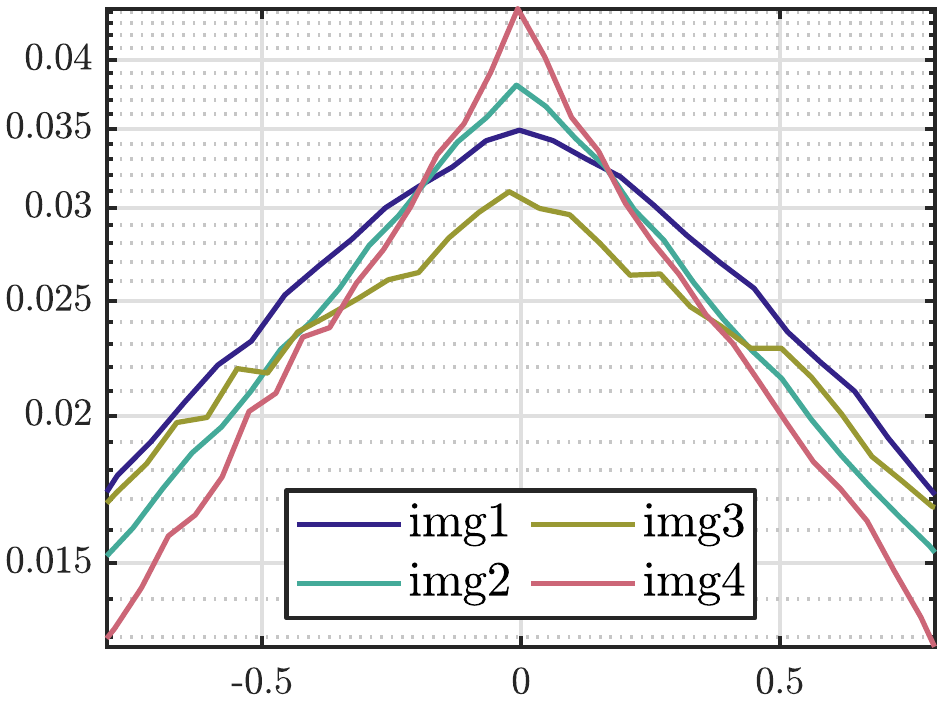} \\
\includegraphics[width=\xheight\linewidth]{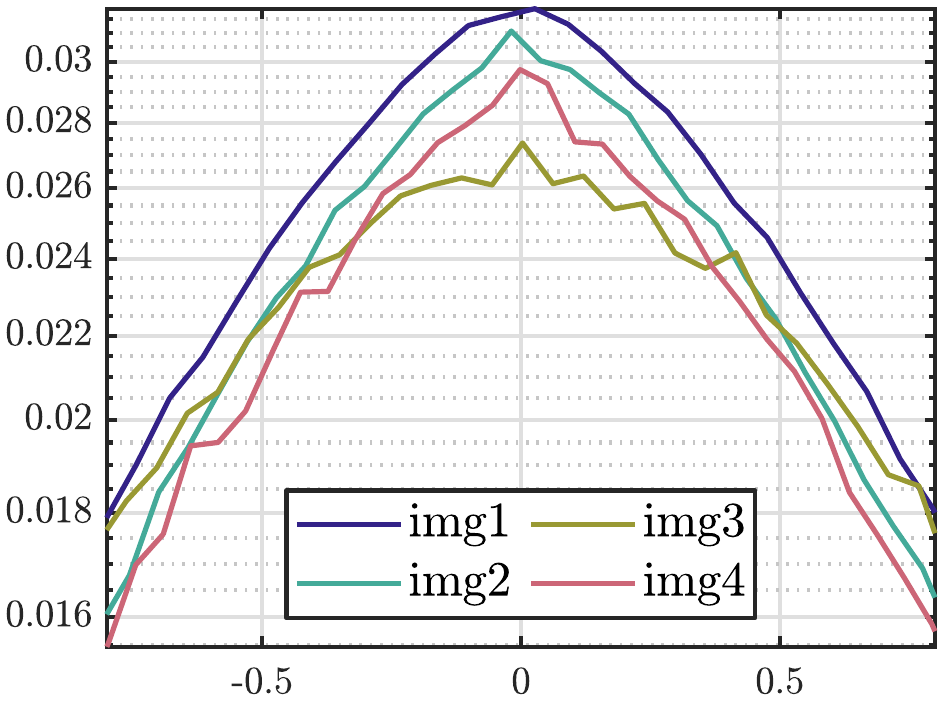} &
\includegraphics[width=\xheight\linewidth]{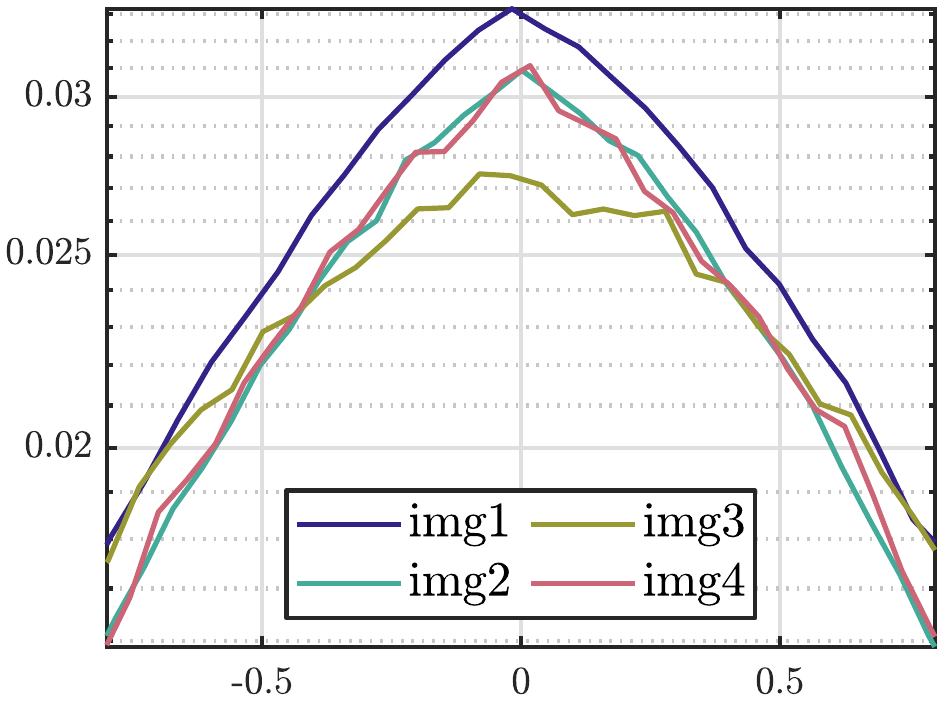} \\
\includegraphics[width=\xheight\linewidth]{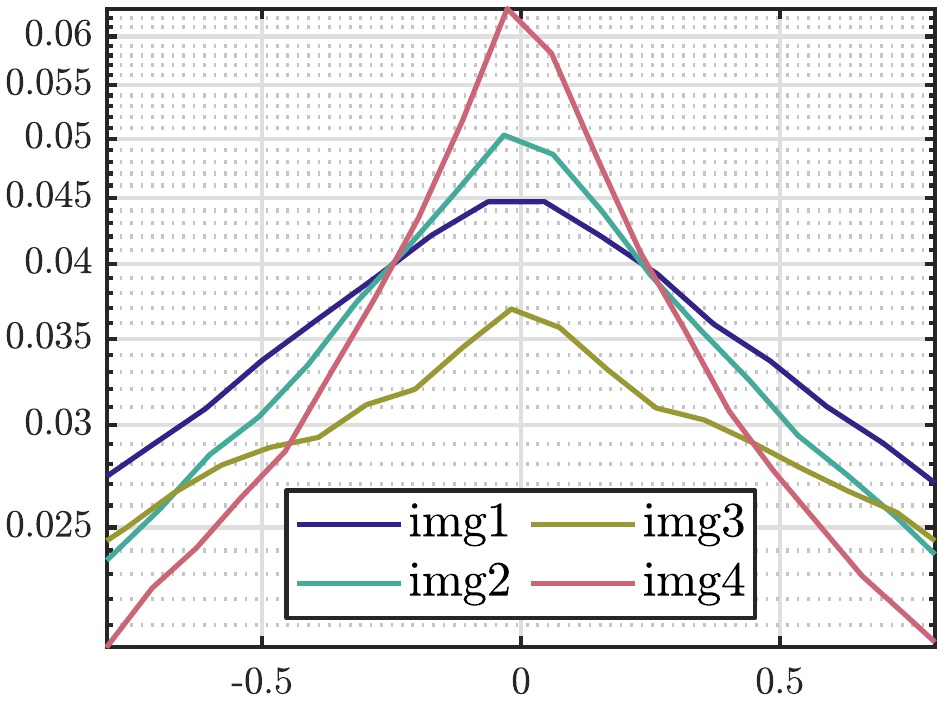} &
\includegraphics[width=\xheight\linewidth]{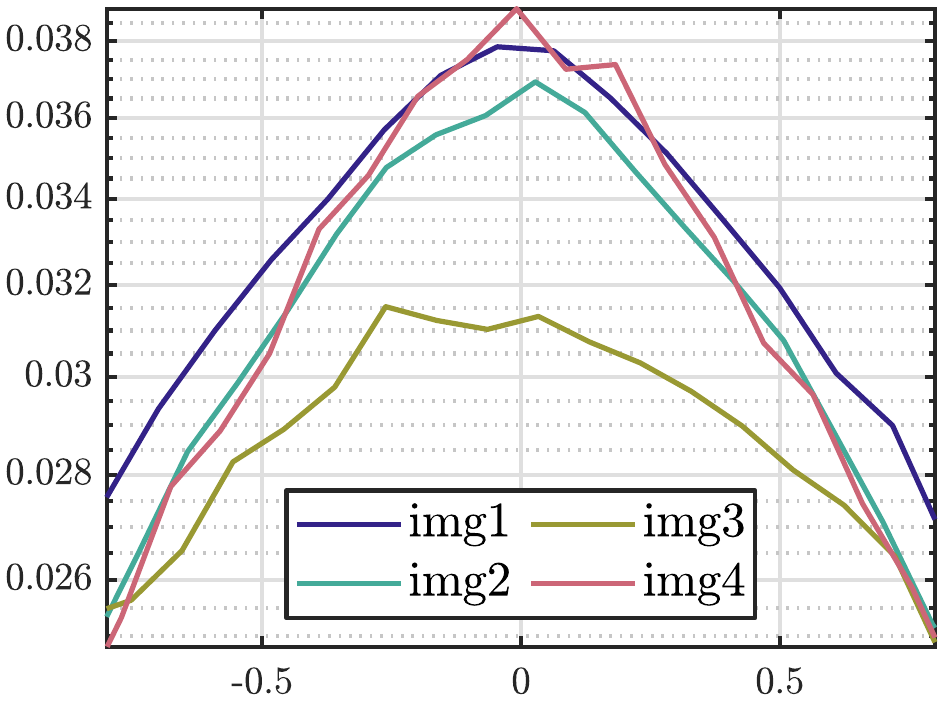} \\
\multicolumn{2}{c}{\includegraphics[width=\xheight\linewidth]{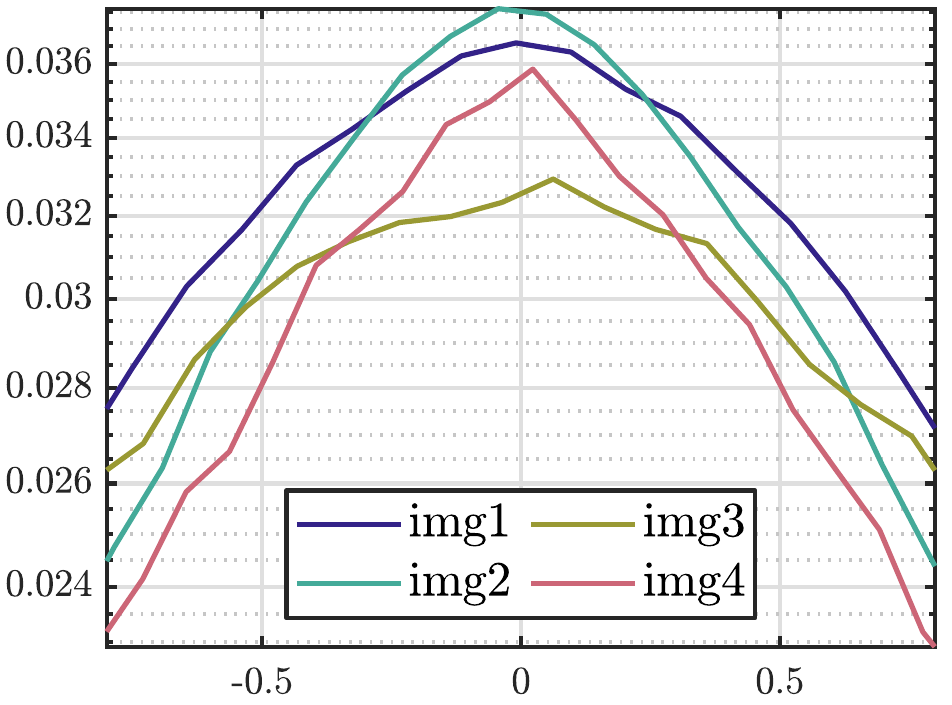}}  \\
\end{tabular}
\caption{Histograms of seven types of MSCN paired log-derivative (Eqs. (\ref{eq:log-deri})) of the four images shown in Fig. \ref{fig:eg_imgs}.}
\label{fig:log_deri}
\end{figure}

It has been empirically observed that the MSCN coefficients of images or video frames have characteristic statistical properties that are altered by the presence of distortion, and therefore, quantifying these deviations can help enable the prediction of perceived quality \cite{ruderman1994statistics, mittal2012no}. A well-known model is the generalized Gaussian distribution (GGD) with zero mean \cite{mittal2012no}:
\begin{equation}
\label{eq:ggd}
f(x;\alpha,\sigma^2)=\frac{\alpha}{2\beta\Gamma(1/\alpha)}\exp{\left(-\left(\frac{|x|}{\beta}\right)^{\alpha}\right)},
\end{equation}
where $\beta=\sigma\sqrt{\frac{\Gamma(1/\alpha)}{\Gamma(3/\alpha)}}$ and $\Gamma(\cdot)$ is the gamma function: $\Gamma(a)=\int_0^{\infty}t^{a-1}e^{-t}dt$. The two parameters are the shape $\alpha$ and the spread $\sigma$, of the zero-mean symmetric GGD, which are estimated using a popular moment-matching based method \cite{sharifi1995estimation}. These are used as features to predict perceptual quality.

Another statistical observation is that the sample distributions of products of pairs of neighboring pixels in the MSCN coefficient map along four directions - horizontal (H) ($\hat{Y}(i,j)\hat{Y}(i,j+1)$), vertical (V) ($\hat{Y}(i,j)\hat{Y}(i+1,j)$), main-diagonal (D1) ($\hat{Y}(i,j)\hat{Y}(i+1,j+1)$), and secondary-diagonal (D2) ($\hat{Y}(i,j)\hat{Y}(i+1,j-1)$) also exhibit a regular statistical structure, which are well-modeled as following a zero mode asymmetric generalized Gaussian distribution (AGGD) \cite{mittal2012no, mittal2015completely}:
\begin{equation}
\label{eq:aggd}
f(x;\nu,\sigma_l^2,\sigma_r^2)=\!
\left\{\!
\begin{array}{lll}

\!\cfrac{\nu}{(\beta_l\!+\!\beta_r)\Gamma(\frac{1}{\nu})}\exp\!{\left(\!-\!\left(\!\cfrac{\!-\!x\!}{\beta_l}\right)^{\!\nu}\right)}\! & \! x\!<\!0 
\\[3ex]
\!\cfrac{\nu}{(\beta_l\!+\!\beta_r)\Gamma(\frac{1}{\nu})}\exp\!{\left(\!-\!\left(\!\cfrac{x}{\beta_r}\right)^{\!\nu}\right)} \! & \! x\!>\!0,
\end{array}
\right.
\end{equation}
where 
\begin{equation}
\label{eq:beta_lr}
\beta_l=\sigma_l\sqrt{\cfrac{\Gamma(1/\nu)}{\Gamma(3/\nu)}}\quad\text{and}\quad \beta_r=\sigma_r\sqrt{\cfrac{\Gamma(1/\nu)}{\Gamma(3/\nu)}}.
\end{equation}

An AGGD model has four parameters: $\nu$ controls the shape of the distribution, while $(\sigma_l,\sigma_r)$ are scale parameters that control the spread along each side of the mode; and $\eta$ is the mean of the distribution, given by $\eta=(\beta_r-\beta_l)\frac{\Gamma(2/\eta)}{\Gamma(1/\eta)}$.

Apart from the second-order pair-product statistics, we also extract another supplementary set of features by modeling the log-derivative statistics of neighboring MSCN coefficient pairs as introduced in \cite{zhang2013no}. Specifically, the absolute pixel values of $\hat{Y}(i,j)$ are first logarithmically transformed:
\begin{equation}
\label{eq:log}
Z(i,j)=\log[|\hat{Y}(i,j)|+0.1],
\end{equation}
then seven types of log-derivative statistics (Eqs. (\ref{eq:log-deri})) along six paired orientations - horizontal (H: $\nabla_x Z(i,j)$), vertical (V: $\nabla_y Z(i,j)$), main-diagonal (MD: $\nabla_{xy} Z(i,j)$), secondary-diagonal (SD: $\nabla_{yx} Z(i,j)$), horizontal-vertical (HV: $\nabla_x\nabla_y Z(i,j)$), and two combined-diagonals (CDs: $\nabla_{cx}\nabla_{cy} Z(i,j)_1,\ \nabla_{cx}\nabla_{cy} Z(i,j)_2$), are modeled as GGD, respectively, after which the estimated GGD parameters are gathered as additional statistical features for learning the eventual quality predictor.
\begin{align}
\label{eq:log-deri}
\begin{split}
D_1:& \nabla_x Z(i,j)  =Z(i,j+1)-Z(i,j) \\
D_2:& \nabla_y Z(i,j)  =Z(i+1,j)-Z(i,j)   \\
D_3:& \nabla_{xy} Z(i,j)  =Z(i+1,j+1)-Z(i,j) \\
D_4:& \nabla_{yx} Z(i,j)  =Z(i+1,j-1)-Z(i,j) \\
D_5:& \nabla_x\nabla_y Z(i,j)  = Z(i-1,j)+Z(i+1,j) \\
 & \quad\quad\quad  -Z(i,j-1)-Z(i,j+1) \\ 
D_6:& \nabla_{cx}\nabla_{cy} Z(i,j)_1 =Z(i,j)+Z(i+1,j+1) \\
 & \quad\quad\quad-Z(i,j+1)-Z(i+1,j) \\
D_7:& \nabla_{cx}\!\nabla_{cy} Z(i,j)_2=Z(i\!-\!1,j\!-\!1)+Z(i\!+\!1,j\!+\!1) \\
 & \quad\quad\quad -Z(i-1,j+1)-Z(i+1,j-1) 
\end{split}
\end{align}

For each pair log-derivative feature map, a single scale NSS model is used to derive two parameters $(\alpha,\sigma)$ by fitting a GGD distribution using the same moment-matching procedure, yielding a total of 14 additional features.

The variance field (or `sigma' field) in Eq. (\ref{eq:sigma}) has been previously shown to provide effective quality-aware features deriving from the same NSS/retinal model \cite{ghadiyaram2017perceptual, kundu2017no}. We extract two additional quantities from the variance field (Eq. (\ref{eq:sigma})): the mean $\phi_\sigma$ and square of the reciprocal of the coefficient of variation (CoV):
\begin{equation}
\label{eq:sigma_field_mu}
\phi_\sigma=\frac{1}{MN}\sum_{i=0}^{M-1}\sum_{j=0}^{N-1}\sigma(i,j)
\end{equation}
where the CoV is $\rho=(\phi_\sigma/\omega_\sigma)^2$ and where
\begin{equation}
\label{eq:sigma_field_omega}
\omega_\sigma=\sqrt{\frac{1}{MN}\sum_{i=0}^{M-1}\sum_{j=0}^{N-1}[\sigma(i,j)-\phi_\sigma]^2}.
\end{equation}

In order to visualize how these NSS regularities are perturbed by \emph{UGC} distortions, we selected four pictures ranging from high quality to low quality - \texttt{10004473376.jpg} (MOS=3.82), \texttt{6462096609.jpg} (MOS=2.93) from KoNIQ-10k \cite{hosu2020koniq} and \texttt{t4.bmp} (MOS=35.8), \texttt{12.bmp} (MOS=15.9) from LIVE-IQC \cite{ghadiyaram2015massive}, as shown in Fig. \ref{fig:eg_imgs}. Fig. \ref{fig:mscn_sigma} plots the histograms of the MSCN and variance field coefficients on these images of diverse perceptual quality. Note that it is extremely difficult to isolate specific distortion types on authentically distorted UGC pictures, since several complex, commingled distortions usually co-exist, hence it is difficult to predict how a given GGD histogram will vary in the presence of quality degradations. However, we may still observe in Fig. \ref{fig:mscn_sigma} that MSCN and sigma coefficients can differentiate images having different perceptual qualities. In this regard, the estimated parameters from these distributions are good indicators of perceptual quality for UGC pictures or videos.

\begin{figure}[!t]
\centering
\footnotesize
\def\xwidth{0.242}
\setlength{\tabcolsep}{1pt}
\begin{tabular}{cccc}
\includegraphics[width=\xwidth\linewidth]{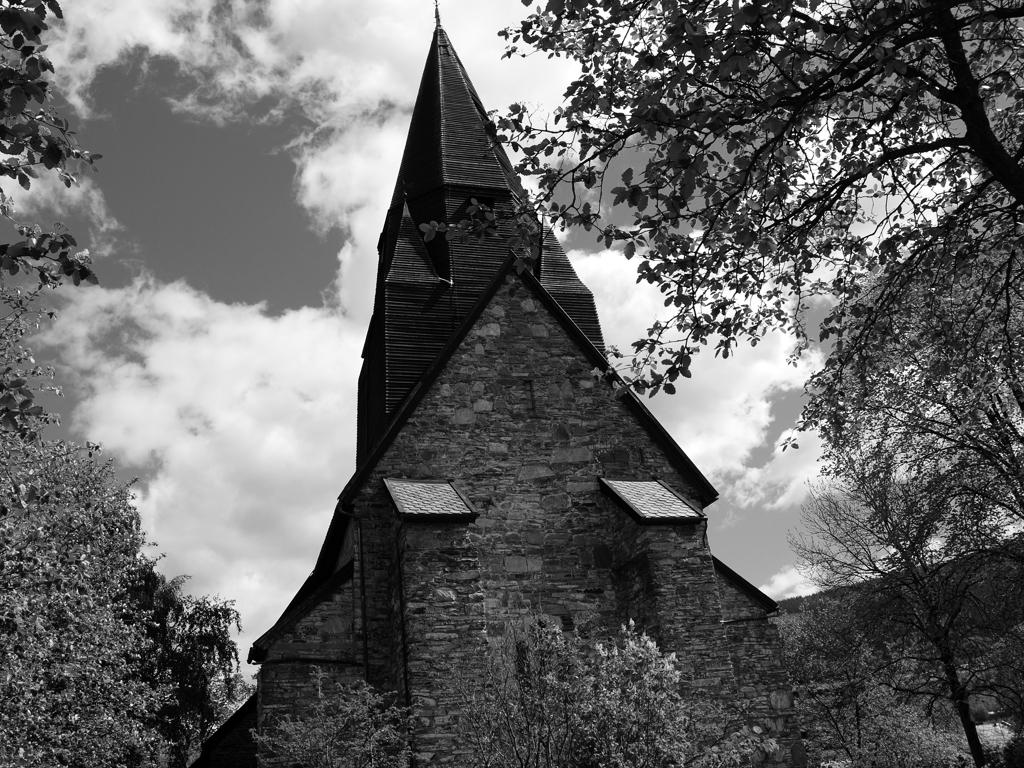} &
\includegraphics[width=\xwidth\linewidth]{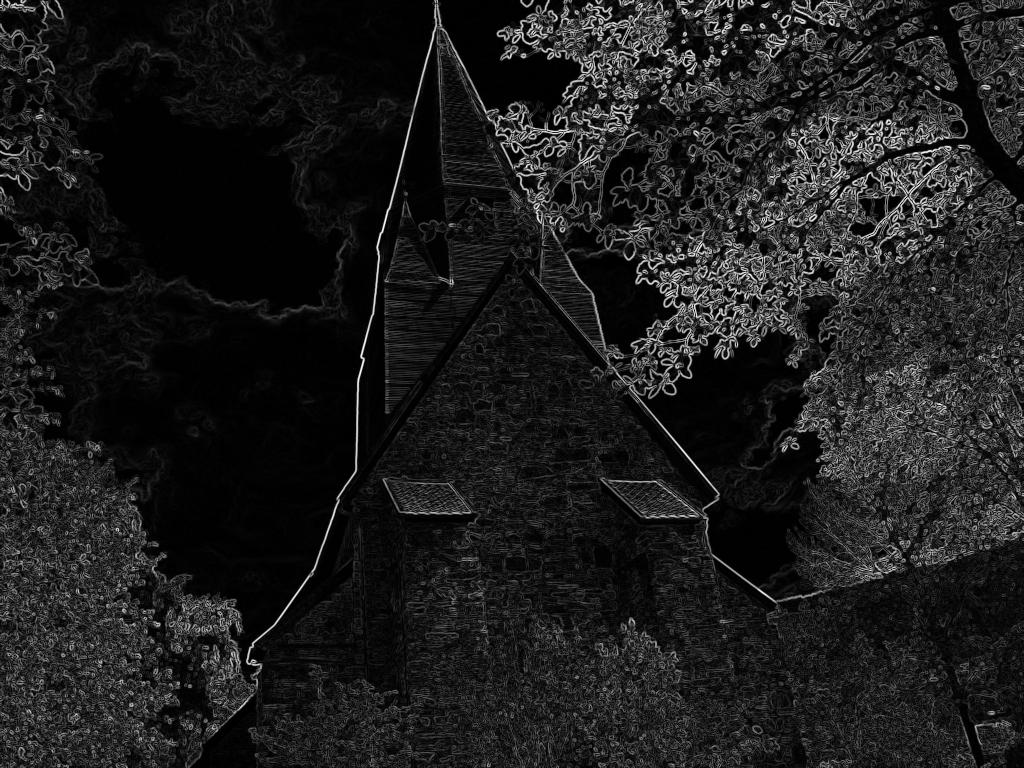} &
\includegraphics[width=\xwidth\linewidth]{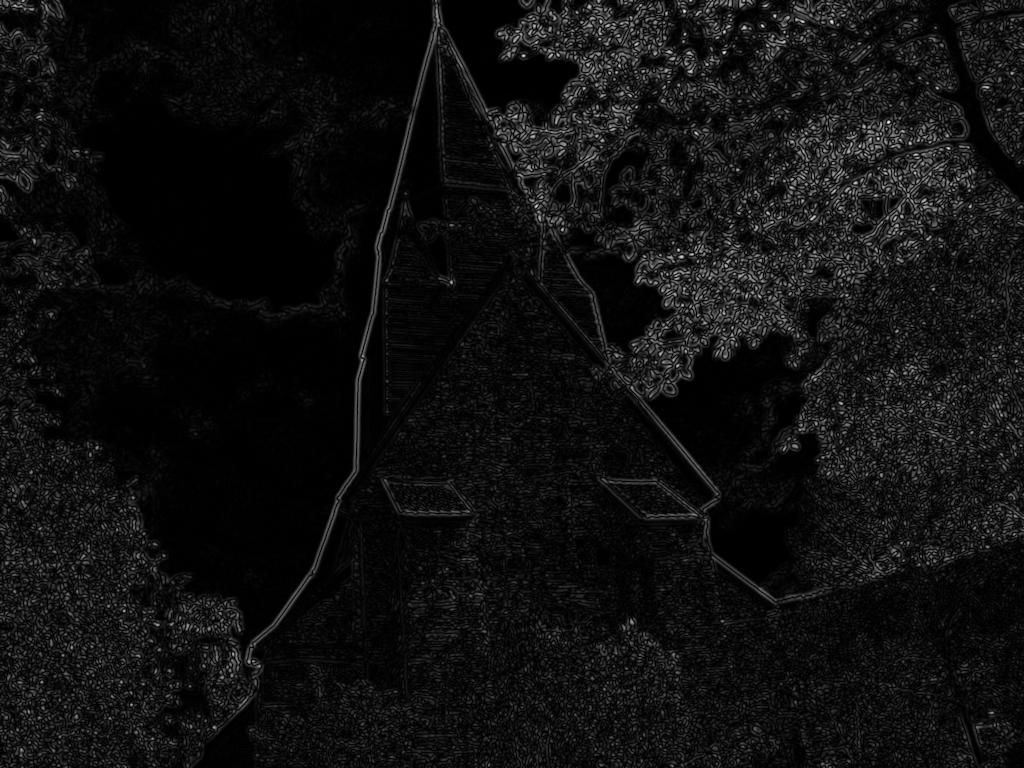} &
\includegraphics[width=\xwidth\linewidth]{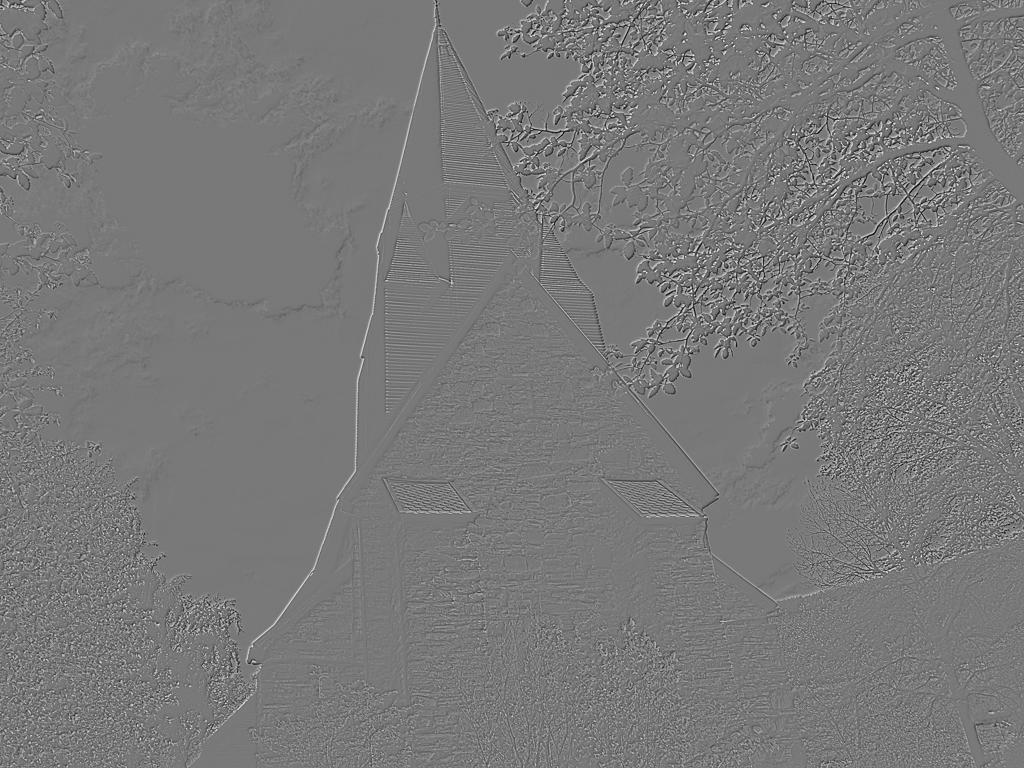} \\
\includegraphics[width=\xwidth\linewidth]{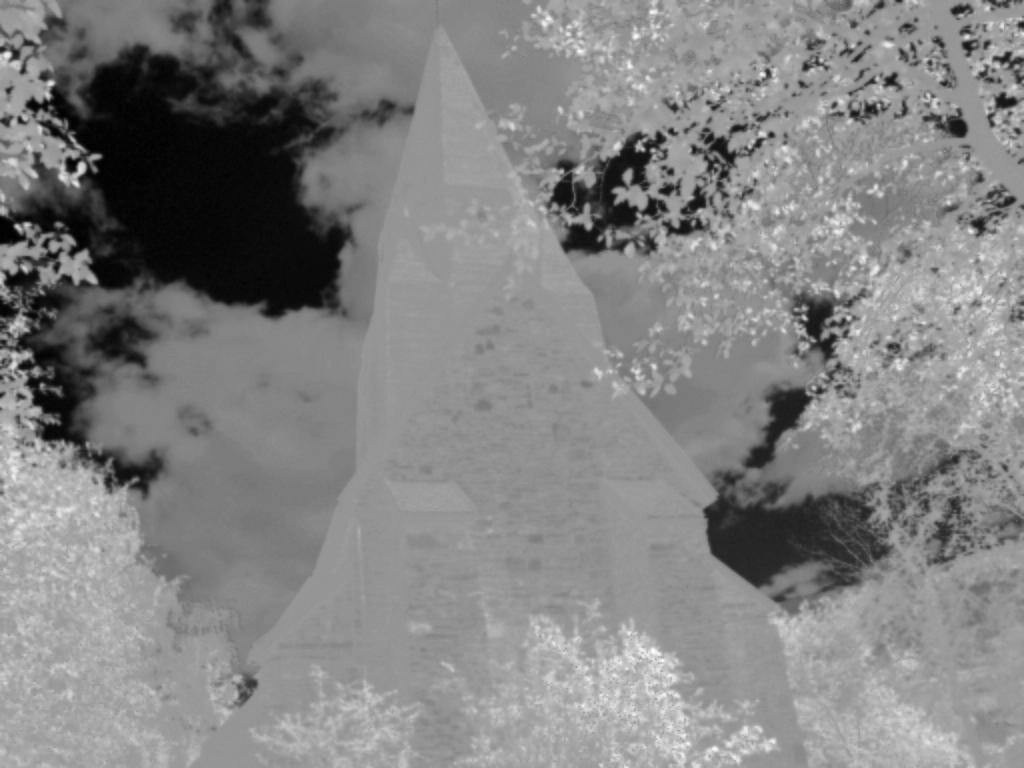} &
\includegraphics[width=\xwidth\linewidth]{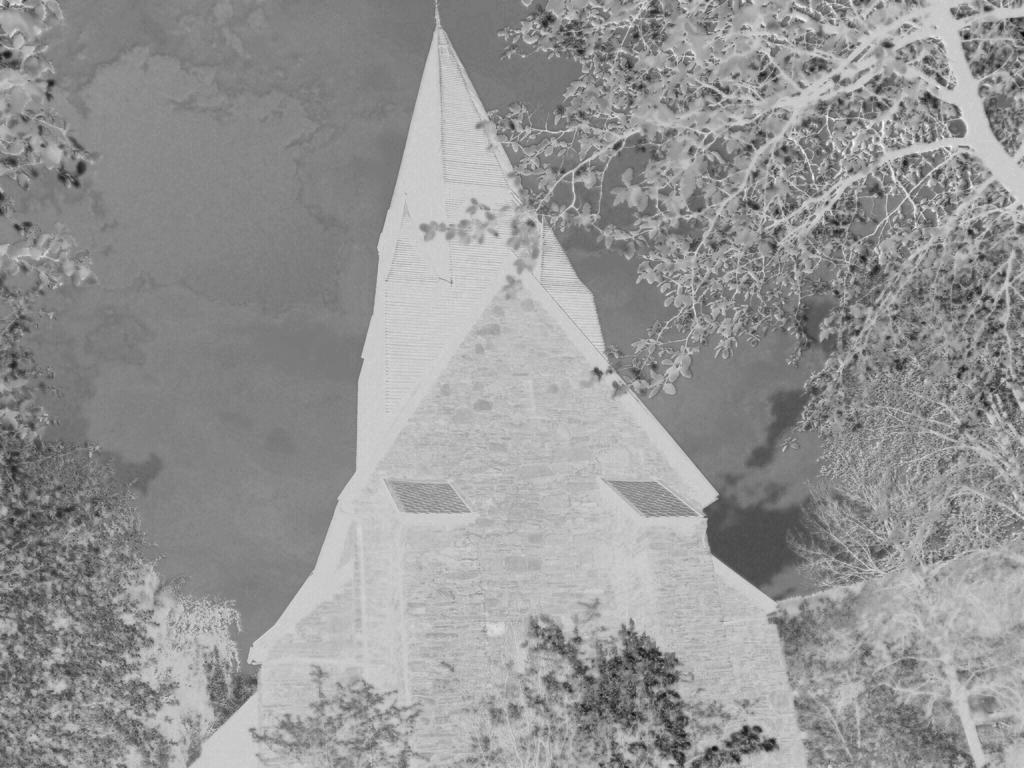} &
\includegraphics[width=\xwidth\linewidth]{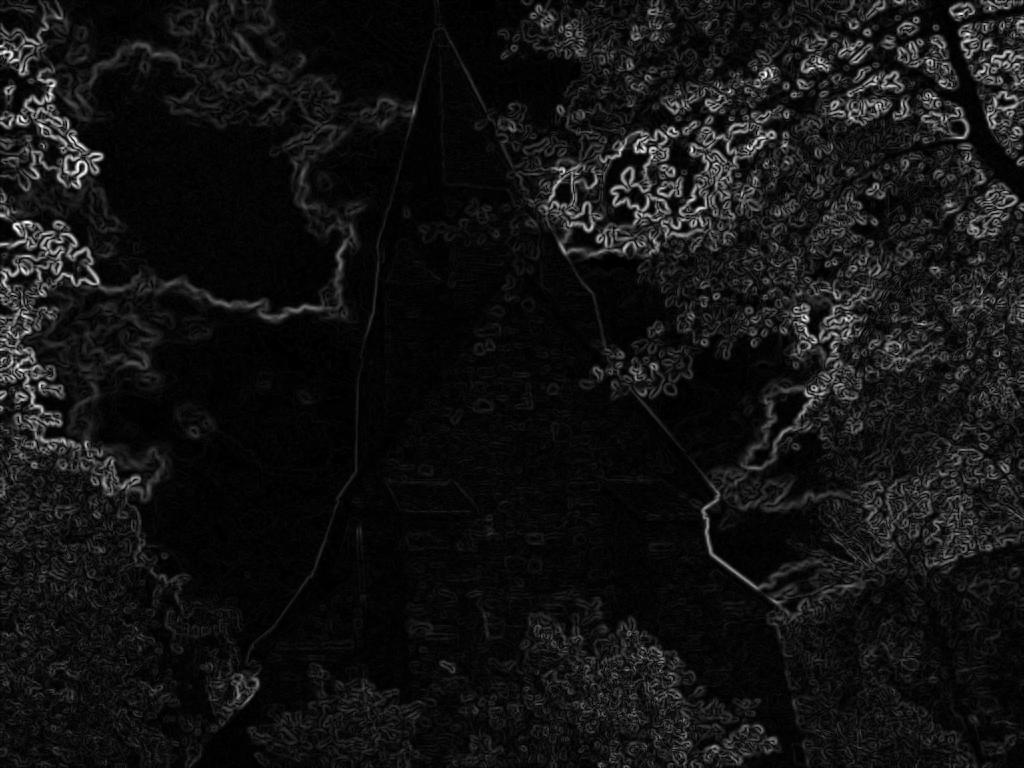} &
\includegraphics[width=\xwidth\linewidth]{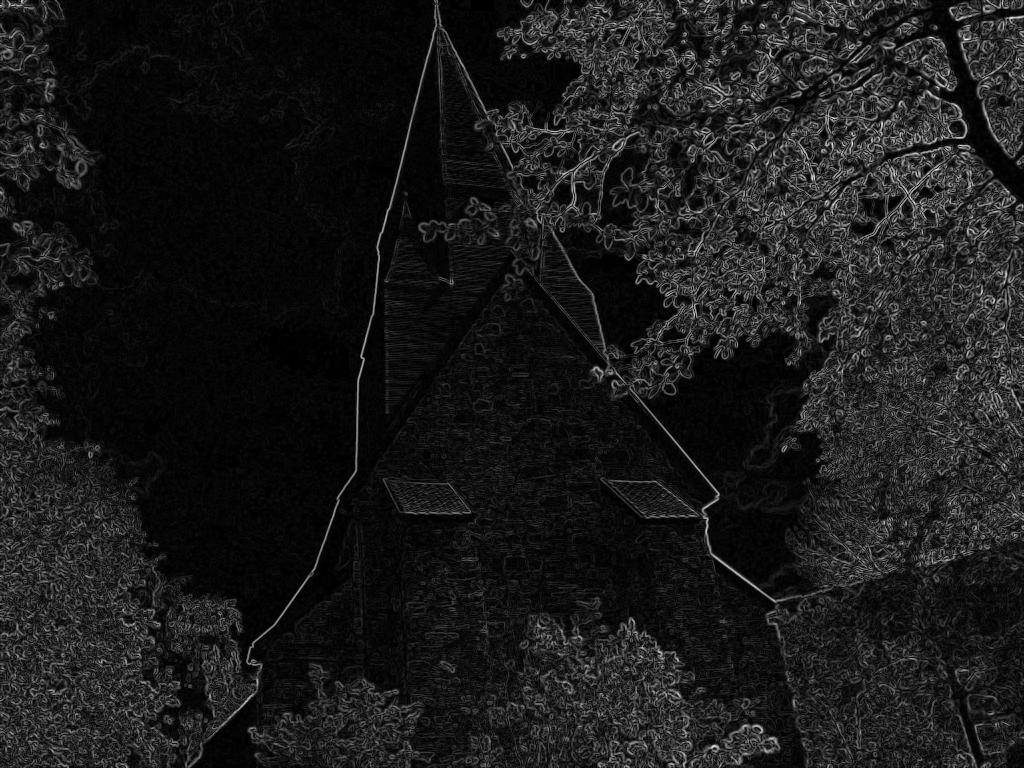} \\
\includegraphics[width=\xwidth\linewidth]{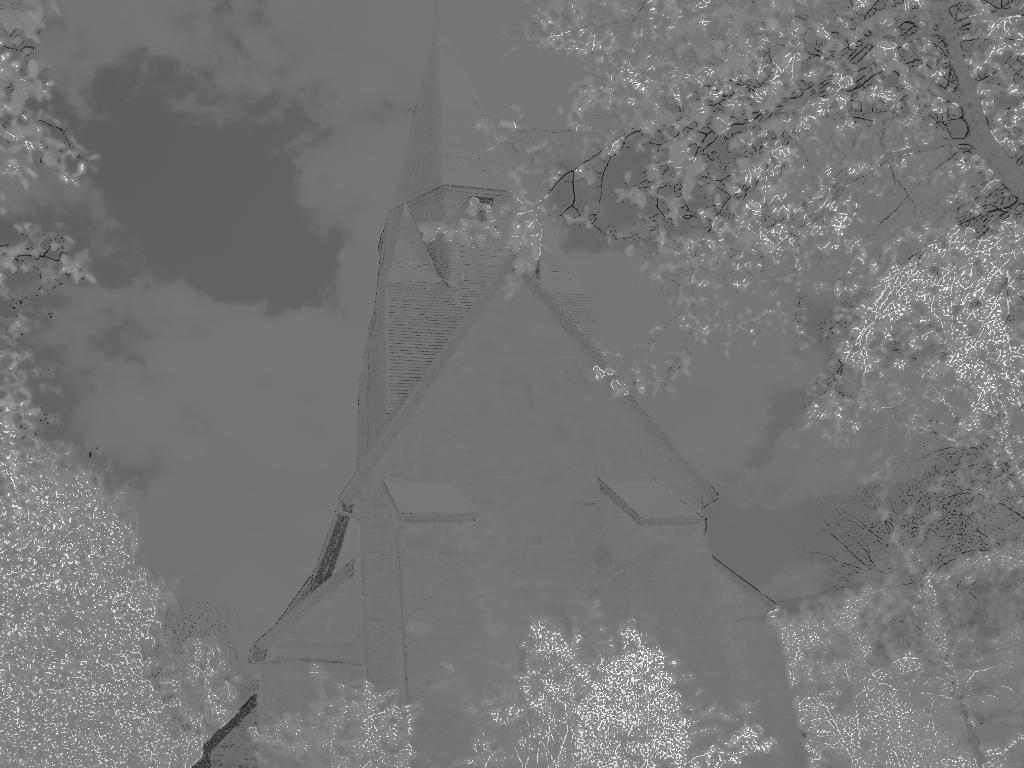} &
\includegraphics[width=\xwidth\linewidth]{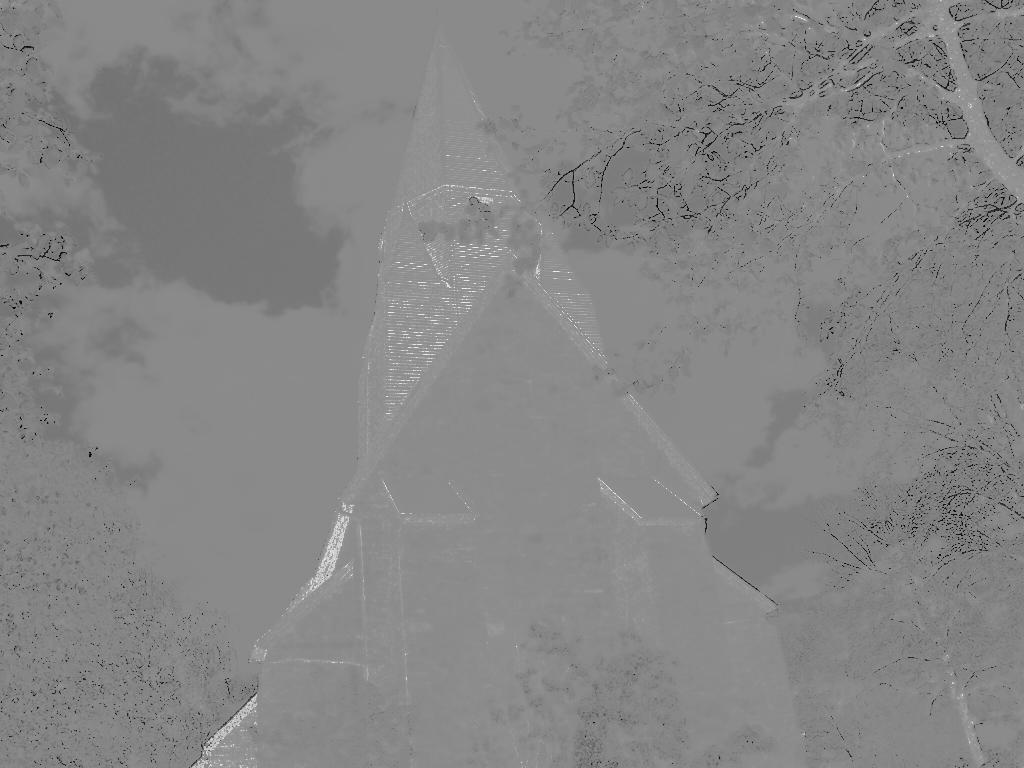} &
\includegraphics[width=\xwidth\linewidth]{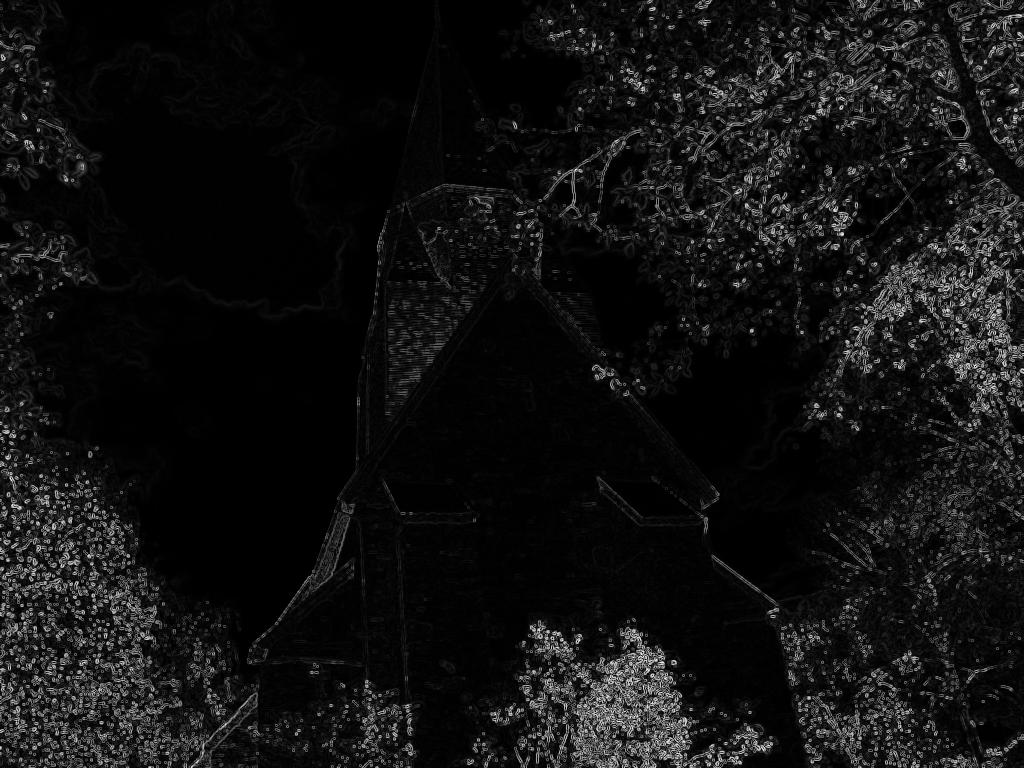} &
\includegraphics[width=\xwidth\linewidth]{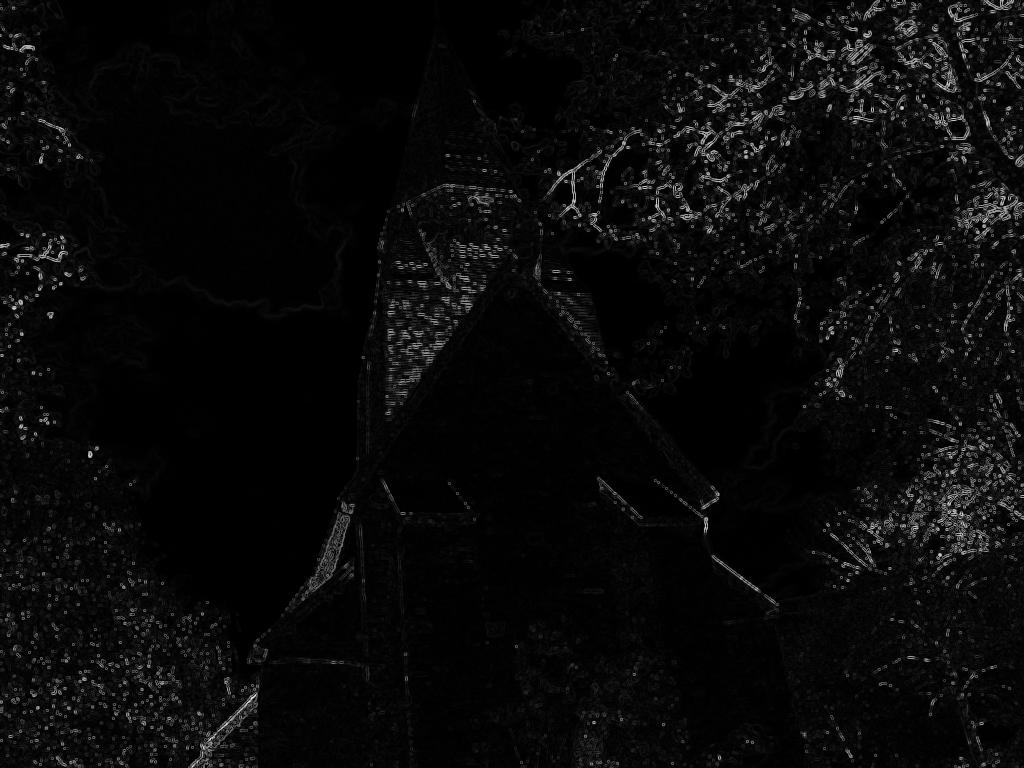} \\
\includegraphics[width=\xwidth\linewidth]{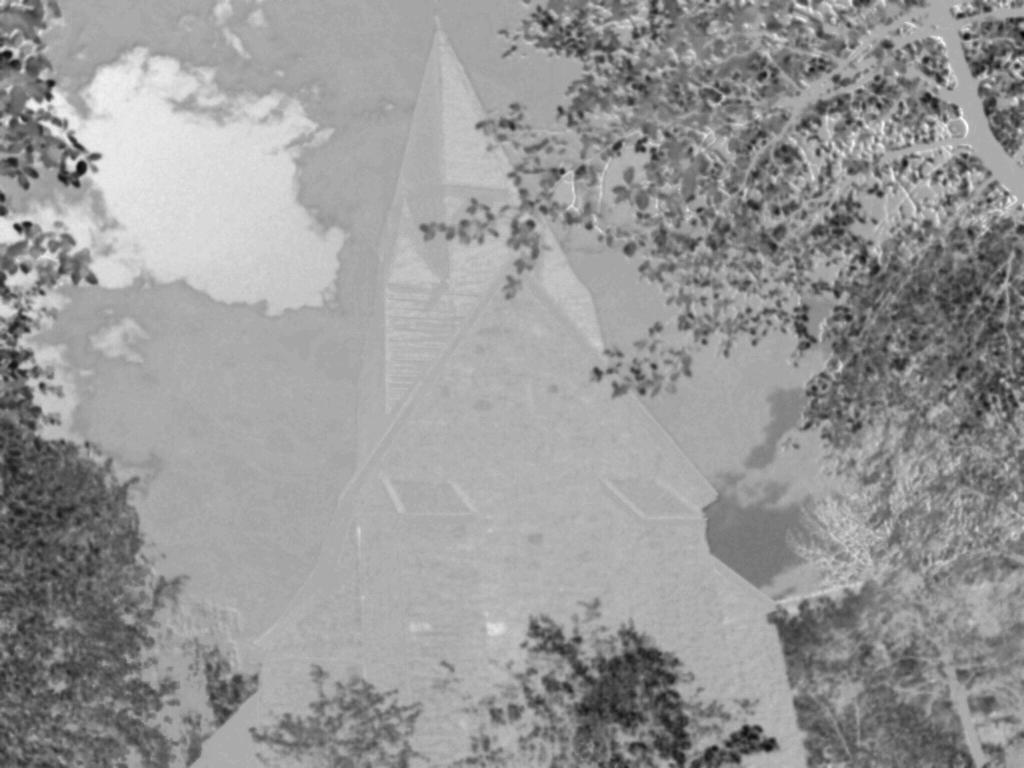} &
\includegraphics[width=\xwidth\linewidth]{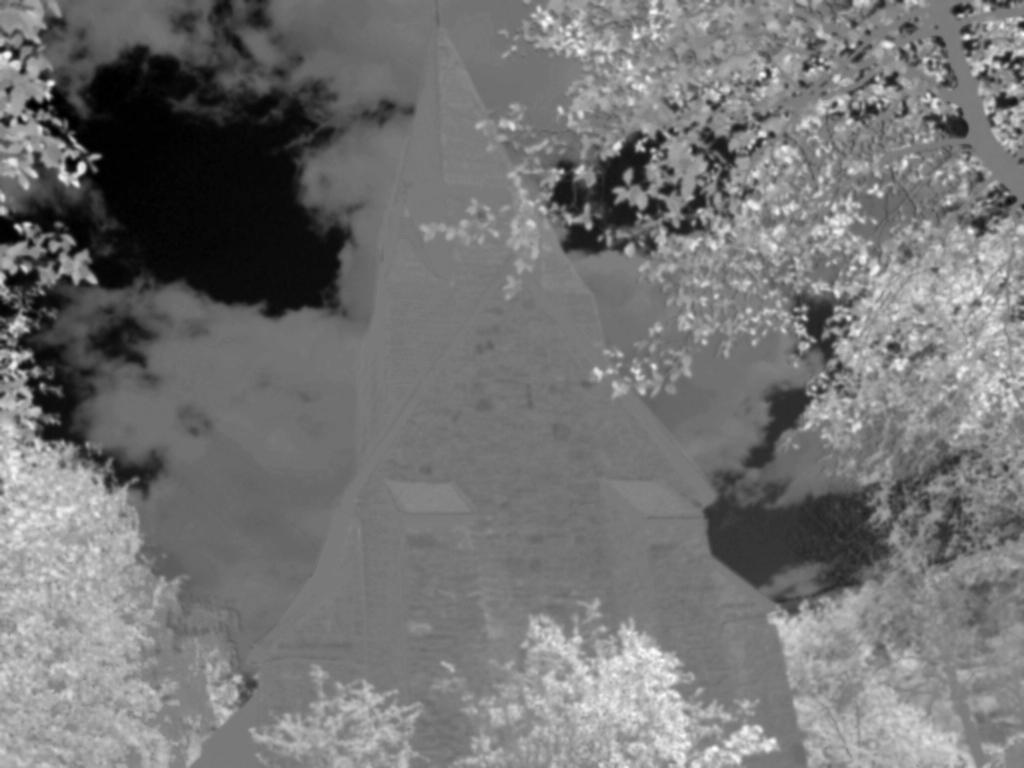} &
\includegraphics[width=\xwidth\linewidth]{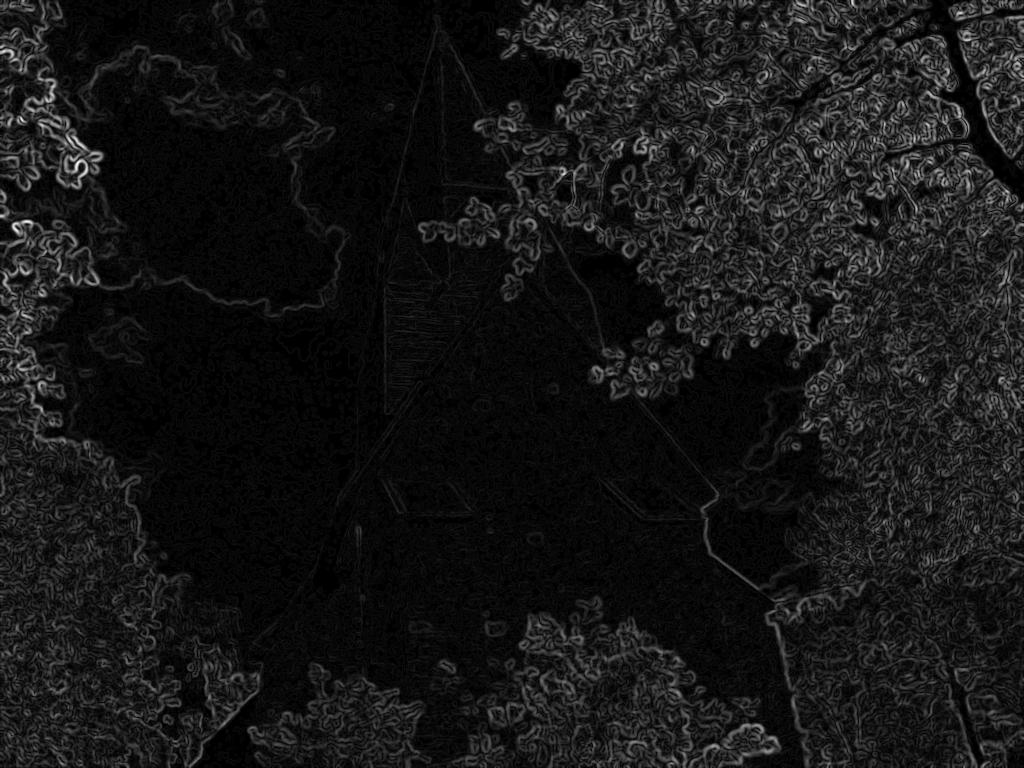} &
\includegraphics[width=\xwidth\linewidth]{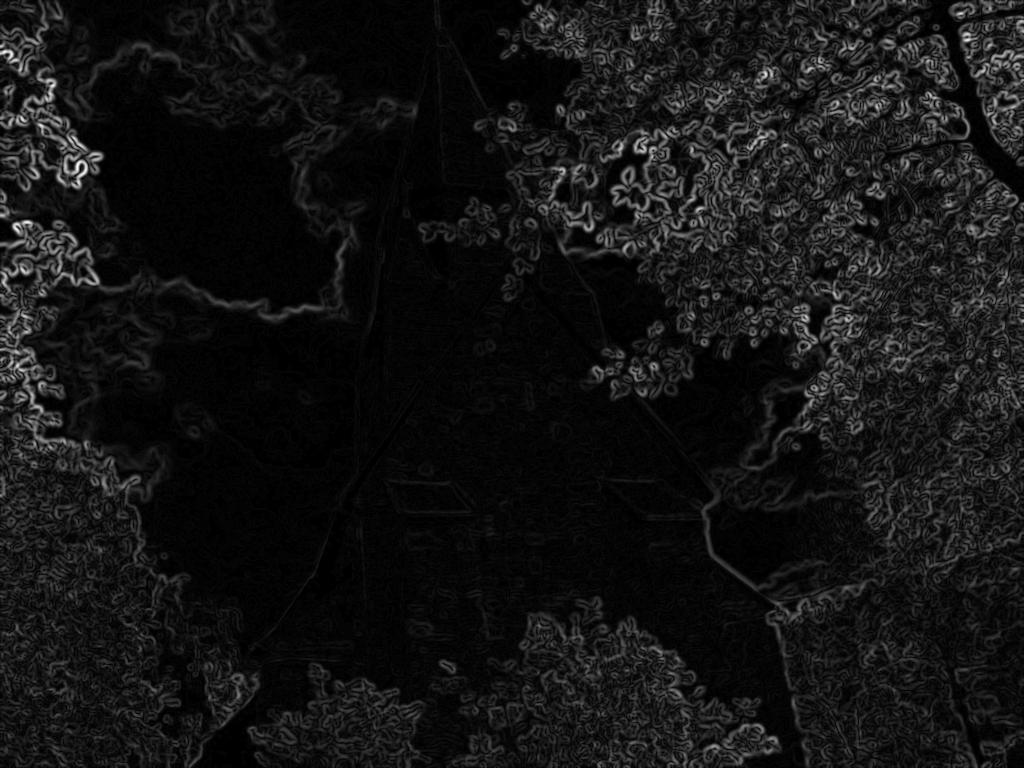} \\
\end{tabular}
\caption{Extracted feature maps defined in Sec. \ref{ssec:spatial_features}. 1st row: $Y$, $GM$, $LoG$, $DoG$; 2nd row: $O_2$, $O_3$, $GMO_2$, $GMO_3$; 3rd row: $BY$, $RG$, $GMBY$, $GMRG$; 4th row: $A$, $B$, $GMA$, $GMB$.}
\label{fig:feat_maps}
\end{figure}

We also plotted the histograms of adjacent pair products and log-derivative statistics in Fig. \ref{fig:pair_prod} and \ref{fig:log_deri}, respectively. It may be observed that the product statistics exhibit similar behavior as the MSCN coefficients, in that their shapes are significantly altered as a function of quality. These statistics better characterize correlations introduced or lost by distortion as compared to first-order MSCN and variance features. The seven types of log-derivative statistics shown in Fig. \ref{fig:log_deri} exhibit distinct deviations against distortion, and all are effective at capturing quality variations on UGC pictures. Therefore, it is informative to include all these statistical features in the final prediction model.

\begin{table}[!t]
\setlength{\tabcolsep}{6pt}
\renewcommand{\arraystretch}{1.}
\centering\footnotesize
\caption{Summary of the proposed NSS-34 feature extraction module.}
\label{table:nss34}
\begin{tabular}{lrr}
\toprule
\textsc{Index} & \textsc{Description} & \textsc{Computation Procedure} \\
\hline\\[-1.em]
$f_1-f_2$ & $(\alpha, \sigma)$ & Fit GGD to MSCN coefficients \\
$f_3-f_4$ & $(\phi_\sigma,\omega_\sigma)$ & Compute statistics on `sigma' map \\
$f_5-f_8$ & $(\nu,\eta,\sigma_l,\sigma_r)$ & Fit AGGD to H pairwise products \\
$f_9-f_{12}$ & $(\nu,\eta,\sigma_l,\sigma_r)$ & Fit AGGD to V pairwise products \\
$f_{13}-f_{16}$ & $(\nu,\eta,\sigma_l,\sigma_r)$ & Fit AGGD to D1 pairwise products\\
$f_{17}-f_{20}$ & $(\nu,\eta,\sigma_l,\sigma_r)$ & Fit AGGD to D2 pairwise products\\
$f_{21}-f_{22}$ & $(\alpha, \sigma)$ & Fit GGD to D\textsubscript{1} pairwise log-derivative \\
$f_{23}-f_{24}$ & $(\alpha, \sigma)$ & Fit GGD to D\textsubscript{2} pairwise log-derivative \\
$f_{25}-f_{26}$ & $(\alpha, \sigma)$ & Fit GGD to D\textsubscript{3} pairwise log-derivative \\
$f_{27}-f_{28}$ & $(\alpha, \sigma)$ & Fit GGD to D\textsubscript{4} pairwise log-derivative \\
$f_{29}-f_{30}$ & $(\alpha, \sigma)$ & Fit GGD to D\textsubscript{5} pairwise log-derivative \\
$f_{31}-f_{32}$ & $(\alpha, \sigma)$ & Fit GGD to D\textsubscript{6} pairwise log-derivative \\
$f_{33}-f_{34}$ & $(\alpha, \sigma)$ & Fit GGD to D\textsubscript{7} pairwise log-derivative \\
\bottomrule
\end{tabular}
\end{table}

\subsection{Spatial Features}
\label{ssec:spatial_features}

We built a basic statistical feature extraction module using the NSS features mentioned in the previous section, as summarized in Table \ref{table:nss34}. Given an input image or feature map, it extracts two features $(\alpha, \sigma)$ from the MSCN transforms, two features $(\phi_\sigma,\omega_\sigma)$ from the variance field, 16 features $4\times(\nu,\eta,\sigma_l,\sigma_r)$ from the AGGD fit of MSCN adjacent pair products along 4 directions, and 14 features $7\times(\alpha, \sigma)$ from GGD fits of paired log-derivatives along 7 directions, yielding a total of 34 features, which we dub NSS-34 for simplicity.

We also hypothesized that the NSS-34 operator is able to extract valuable color quality information if applied in a chromatic space, which we deem to be a more unified and efficient approach than crafting specific, complex features in different feature spaces, as is done in FRIQUEE \cite{ghadiyaram2017perceptual}. The proposed NSS-34 feature set is based on fast, low-level statistics derived from GGD and AGGD models only, which we will show to deliver superior efficiency in the experimental section.

\begin{figure*}[!t]
\centering
\footnotesize
\def\xheight{0.068}
\def\yheihgt{0.12}
\def\rdshift{0.0562\linewidth}
\setlength{\tabcolsep}{1pt}
\begin{tabular}{cccccccc}
frame 0 &
frame 1 &
frame 2 &
frame 3 & 
frame 4 &
frame 5 &
frame 6 &
frame 7 
\\
\includegraphics[height=\xheight\linewidth]{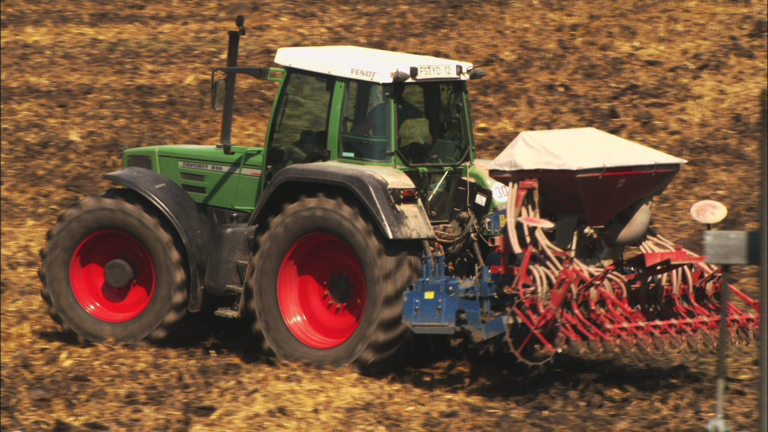}  & 
\includegraphics[height=\xheight\linewidth]{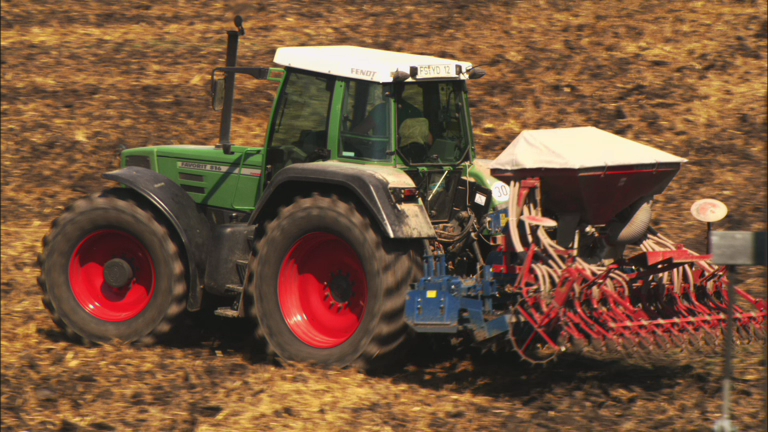}  & 
\includegraphics[height=\xheight\linewidth]{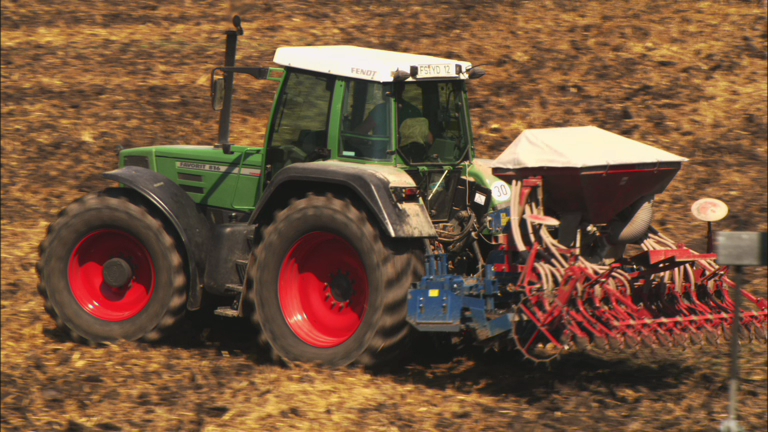} & 
\includegraphics[height=\xheight\linewidth]{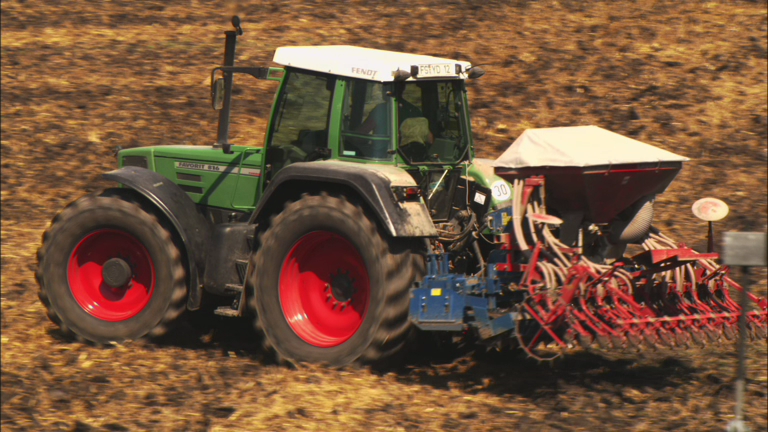} & 
\includegraphics[height=\xheight\linewidth]{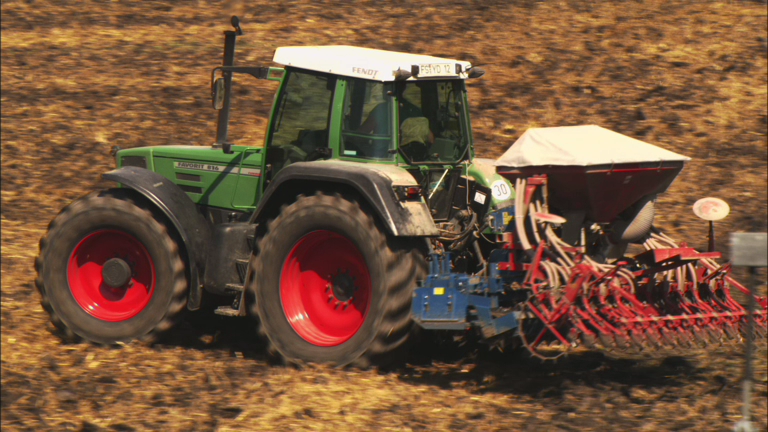} & 
\includegraphics[height=\xheight\linewidth]{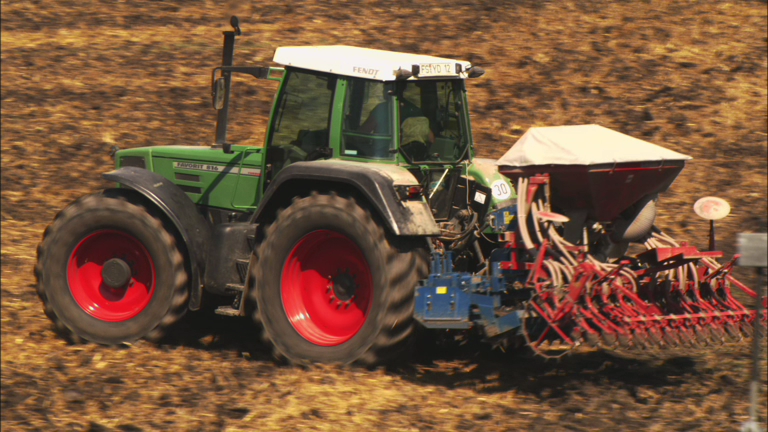} & 
\includegraphics[height=\xheight\linewidth]{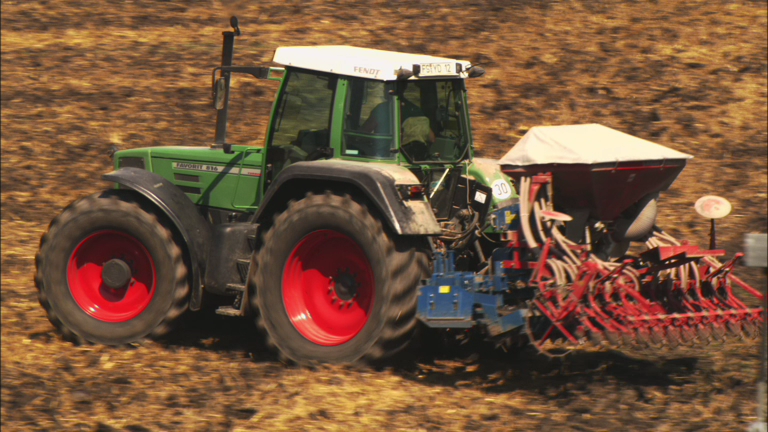} & 
\includegraphics[height=\xheight\linewidth]{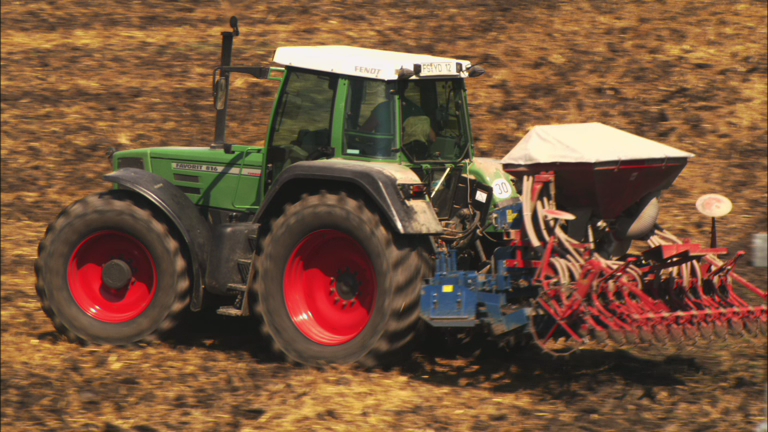}
\\
subband 0 &
subband 1 &
subband 2 &
subband 3 &
subband 4 &
subband 5 &
subband 6 & 
subband 7
\\
\includegraphics[height=\xheight\linewidth]{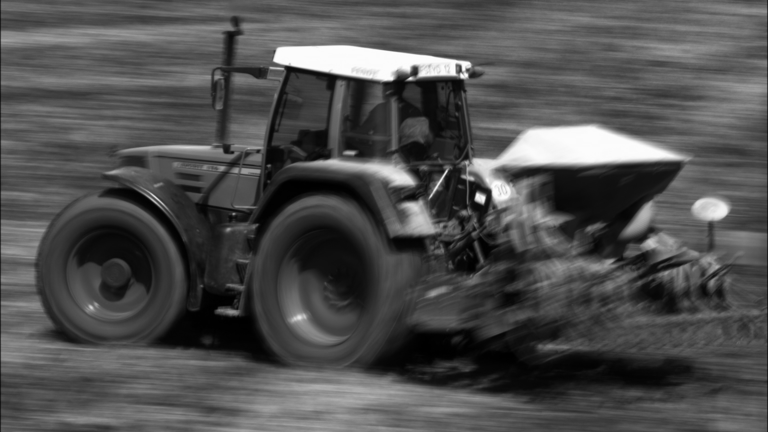}  &
\includegraphics[height=\xheight\linewidth]{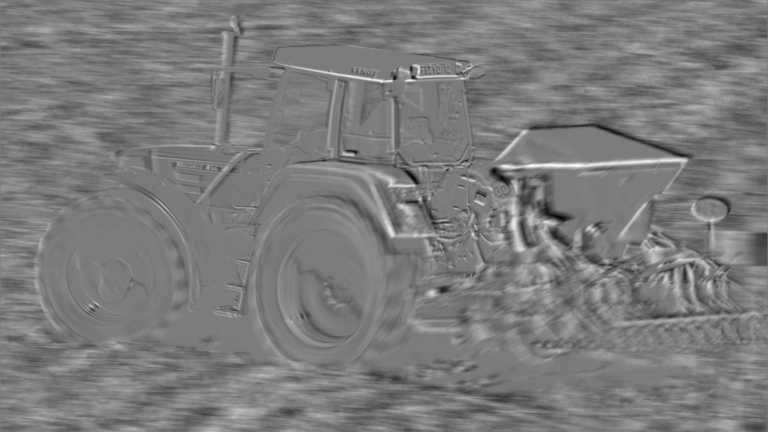}  &
\includegraphics[height=\xheight\linewidth]{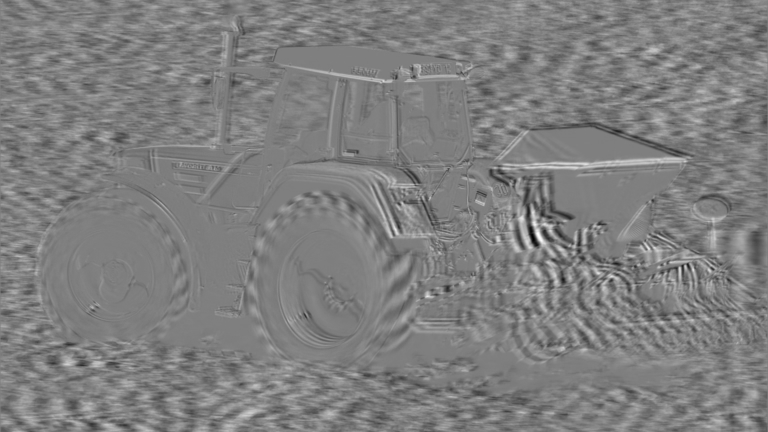}  &
\includegraphics[height=\xheight\linewidth]{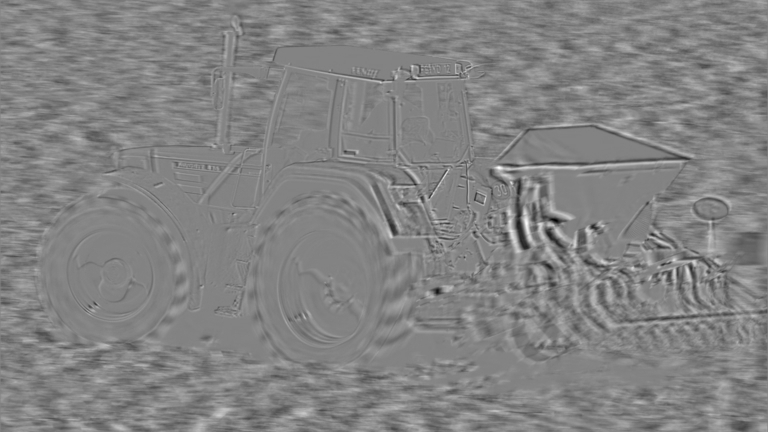}  &
\includegraphics[height=\xheight\linewidth]{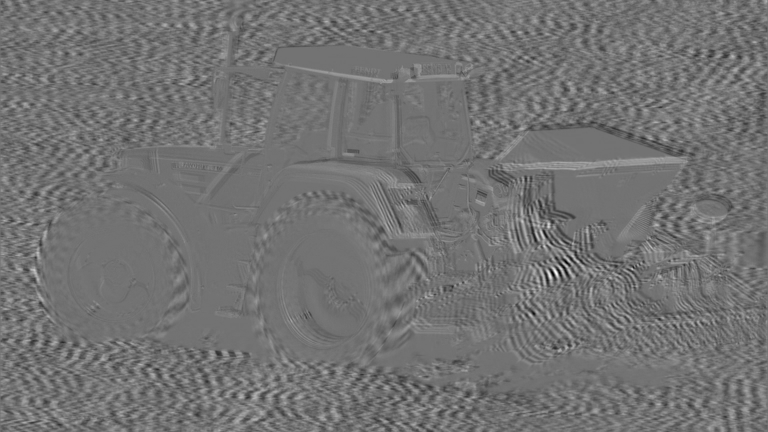}  &
\includegraphics[height=\xheight\linewidth]{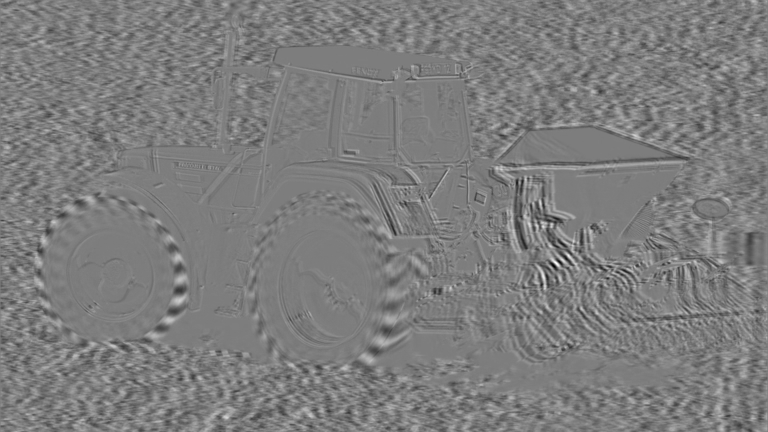}  &
\includegraphics[height=\xheight\linewidth]{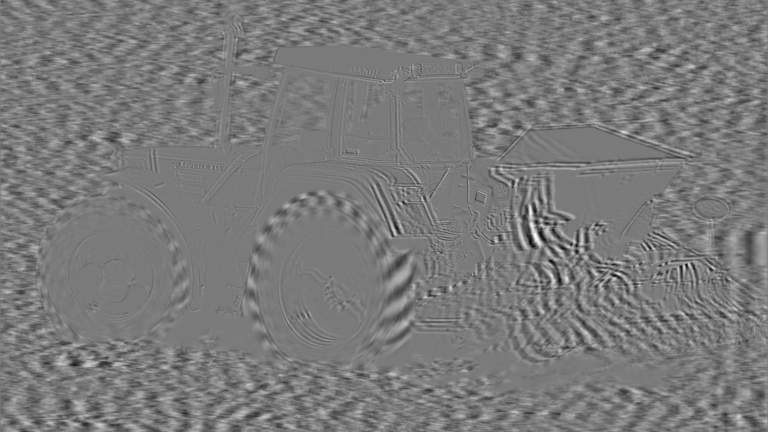}  &
\includegraphics[height=\xheight\linewidth]{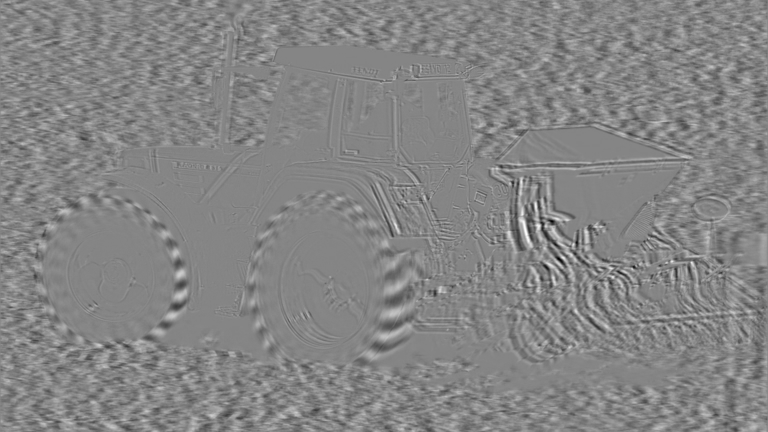}
\\
\multirow{1}{*}[\rdshift]{\includegraphics[width=\yheihgt\linewidth]{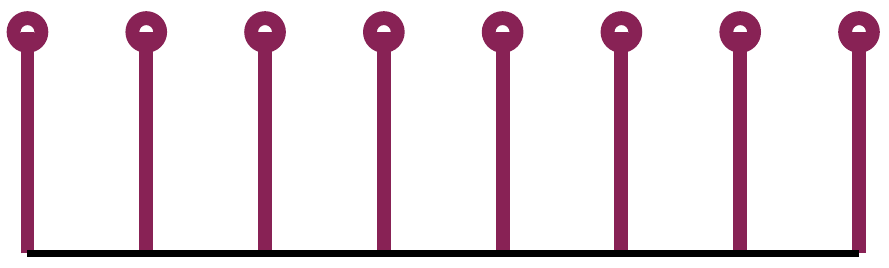}}  &
\includegraphics[width=\yheihgt\linewidth]{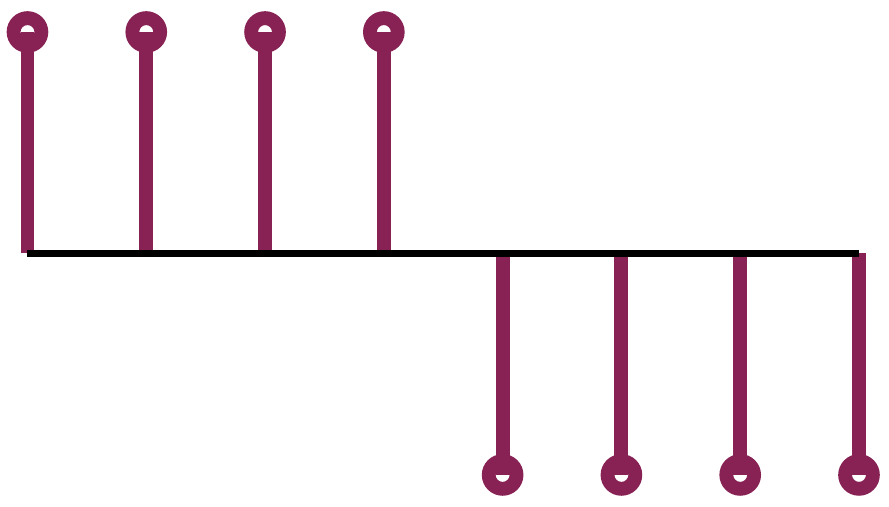}  &
\includegraphics[width=\yheihgt\linewidth]{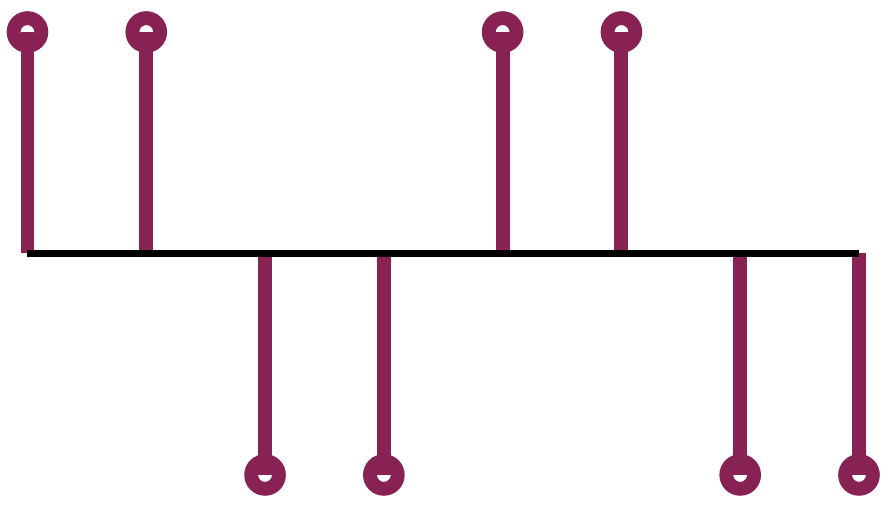}  &
\includegraphics[width=\yheihgt\linewidth]{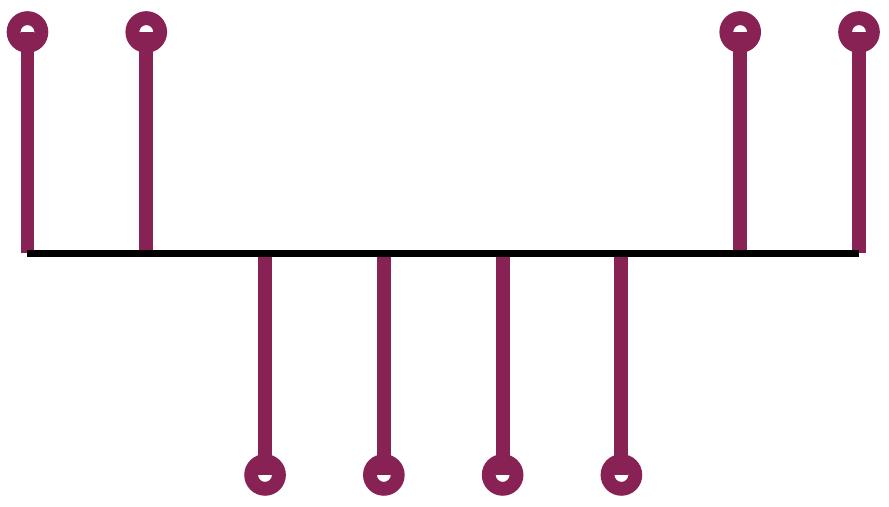}  &
\includegraphics[width=\yheihgt\linewidth]{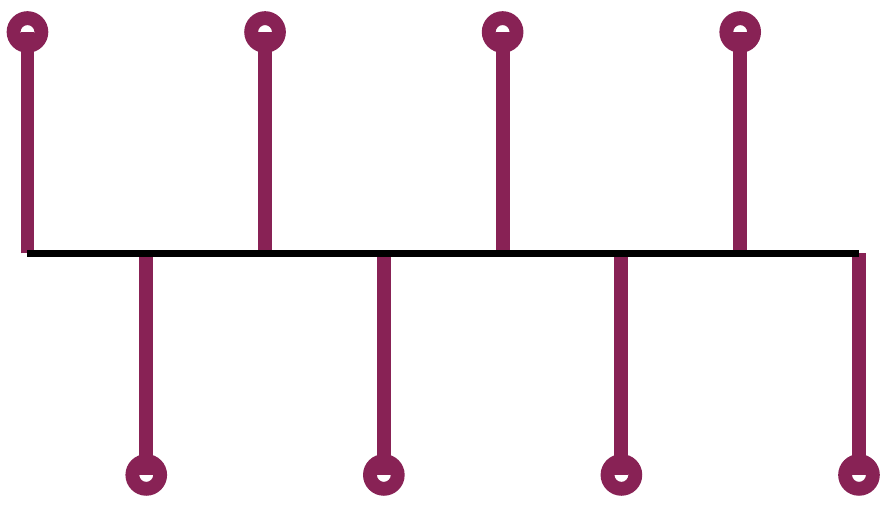}  &
\includegraphics[width=\yheihgt\linewidth]{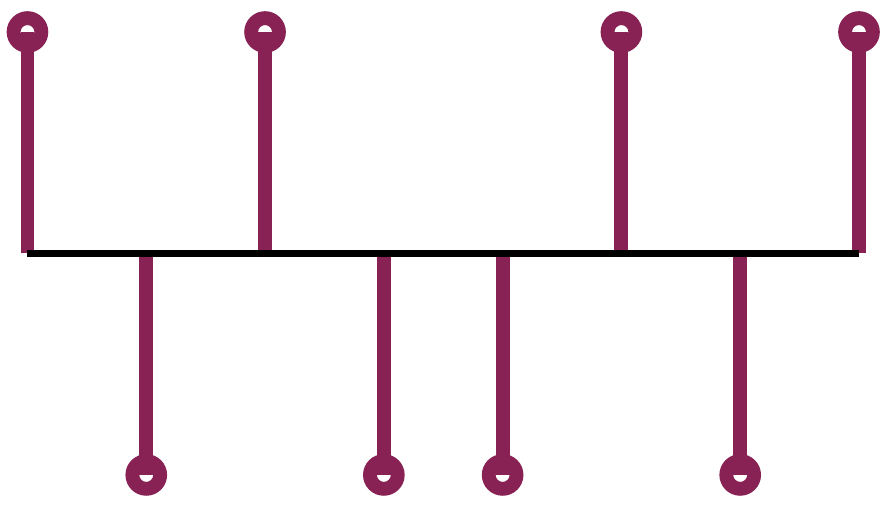}  &
\includegraphics[width=\yheihgt\linewidth]{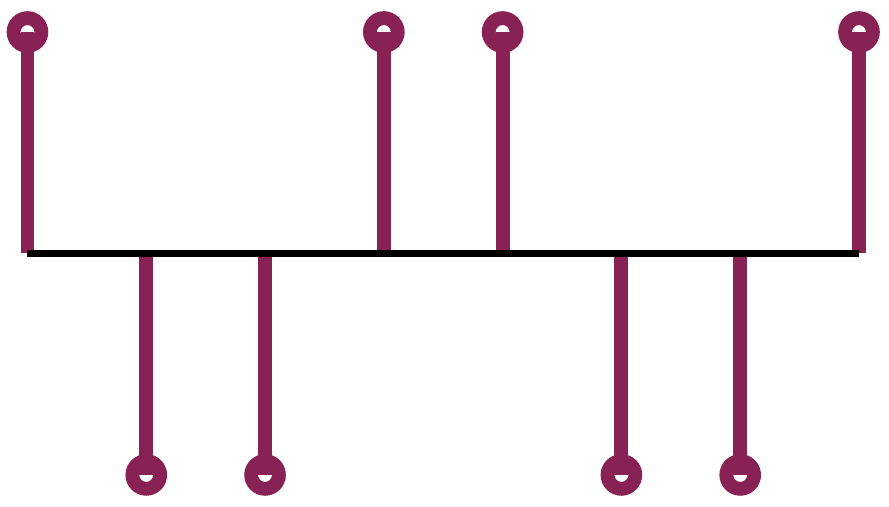}  &
\includegraphics[width=\yheihgt\linewidth]{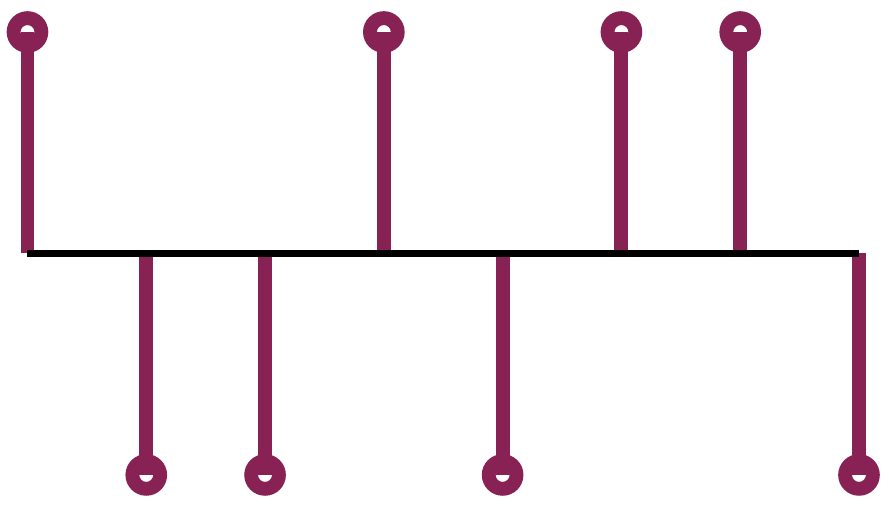}
\\
\end{tabular}
\caption{Top row: eight exemplar consecutive frames sampled from sequence \texttt{Tractor} in LIVE-VQA \cite{seshadrinathan2010study}. Middle row: temporal bandpass-filtered responses by convolving with the filters in an 8-subband Haar-wavelet filter bank, which are shown in the bottom row. The subband frequency increases from left to right: $k=0,...,7$, for both the responses and wavelet functions.}
\label{fig:haar}
\end{figure*}

The gradient magnitude (GM) of a video frame is defined as the root mean square of directional gradients along two orthogonal spatial directions. GM is computed by convolving with a linear filter such as the Roberts, Sobel, or Prewitt. We utilized the Sobel kernels:
\begin{equation}
\label{eq:sobel}
h_x=
\begin{bmatrix}
+1 & 0 & -1 \\
+2 & 0 & -2 \\
+1 & 0 & -1 \\
\end{bmatrix}
\ \text{and}\  
h_y=
\begin{bmatrix}
+1 & +2 & +1 \\
0 & 0 & 0 \\
-1 & -2 & -1 \\
\end{bmatrix}
\end{equation}
whereby the GM of an image or frame $I(i,j)$ is calculated by:
\begin{equation}
\label{eq:gm}
GM=\sqrt{(I\ast h_x)^2+(I\ast h_y)^2},
\end{equation}
where $\ast$ denotes the convolution operator.

It has been observed that two dimensional difference-of-Gaussian (DoG) or Laplacian-of-Gaussian (LoG) operators well-characterize the multiscale receptive fields of retinal ganglion cells \cite{campbell1968application}. We also extract two bandpass maps, using LoG and DoG, and extract their corresponding NSS-34 features, respectively. The LoG of image $I$ is:
\begin{equation}
\label{eq:Lap-of-G}
LoG = I\ast h_{LoG},
\end{equation}
where the LoG kernel is defined as:
\begin{equation}
\label{eq:log_h}
\begin{split}
h_{LoG}&=\left(\frac{\partial^2}{\partial x^2}+\frac{\partial^2}{\partial y^2}\right)g_\sigma(x,y)\\
&=\frac{x^2+y^2-2\sigma^2}{2\pi\sigma^6}\exp{\left(-\frac{x^2+y^2}{2\sigma^2}\right)},
\end{split}
\end{equation}
where $g_\sigma(x,y)$ is an isotropic Gaussian function with scale parameter $\sigma$. We used a window size of $9\times 9$ for LoG filtering. 

While the GM and LoG are used by RAPIQUE to amplify high-frequency responses relating to local frame structures, the DoG is configured to capture mid-frequencies, expressive of structure at larger bandpass scales. The DoG response is defined as the difference of the responses of two Gaussian filters with different standard deviations
\begin{equation}
\label{eq:dog}
DoG=I\ast g_{\sigma_1}-I\ast g_{\sigma_2}=I\ast(g_{\sigma_1}-g_{\sigma_2}).
\end{equation}

To avoid redundant information between the LoG and DoG, only the first level of an $N$-level DoG decomposition with $k=1.6,\ \sigma_i=k^{i-1},i=1,...,N-1$ is utilized. Fig. \ref{fig:feat_maps} shows the differences between the GM, LoG, and DoG responses on a sample video frame. Overall, the four luma channel feature maps $(Y,GM,LoG,DoG)$ (where $Y = 0.299R + 0.587G + 0.114B$) are fed into the NSS-34 module to obtain useful statistical features.

Most previous BVQA models have overlooked the importance of chromatic features, whereas recent work \cite{tu2020ugc, ghadiyaram2017capture, kundu2017no, chen2020chroma} has shown the efficacy of color components for UGC video quality prediction. Previous efforts on the chromatic statistics of quality models involve opponent color spaces such as YIQ/YUV \cite{zhang2011fsim, pei2015image}, $\mathrm{O_1O_2O_3}$ \cite{zhang2015feature}, LMS \cite{zhang2015feature, ghadiyaram2017perceptual}, perceptual color spaces like CIELAB \cite{rajashekar2010perceptual, kundu2017no, ghadiyaram2017perceptual}, HSI \cite{lee2016toward, ghadiyaram2017perceptual}, Yellow color \cite{ghadiyaram2017perceptual}, and ``colorfulness'' features \cite{hasler2003measuring, korhonen2019two, tu2020ugc}. Here we deploy perceptually relevant color transforms from RGB frames (where $R(i,j), G(i,j), B(i,j)$ are red, green, and blue channels) to $\mathrm{O_1O_2O_3}$, red-green (RG), and blue-yellow (BY) as follows:
\begin{equation}
\label{eq:o1o2o3}
\begin{bmatrix}
O_1 \\
O_2 \\
O_3 \\
\end{bmatrix}
= 
\begin{bmatrix}
0.06 & 0.63 & 0.27 \\
0.30 & 0.04 & -0.35 \\
0.34 & -0.60 & 0.17 \\
\end{bmatrix}
\begin{bmatrix}
R \\
G \\
B \\
\end{bmatrix}
\end{equation}
and
\begin{equation}
\label{eq:lms}
\begin{split}
\mathcal{R}(i,j)&=\log[R(i,j)+0.1]-\mu_R \\
\mathcal{G}(i,j)&=\log[G(i,j)+0.1]-\mu_G \\
\mathcal{B}(i,j)&=\log[B(i,j)+0.1]-\mu_B \\
\end{split}
,
\end{equation}
where $\mu_R$, $\mu_G$, and $\mu_B$ are the average values of 
$\log[R(i,j)+0.1]$, $\log[G(i,j)+0.1]$, and $\log[B(i,j)+0.1]$, respectively, over each entire frame. Then the RG and BY opponent color space values are
\begin{equation}
\label{eq:by_rg}
\begin{split}
\hat{L}&=(\mathcal{R}+\mathcal{G}+\mathcal{B})/\sqrt{3} \\
BY&=(\mathcal{R}+\mathcal{G}-2\mathcal{B})/\sqrt{6} \\
RG&=(\mathcal{R}-\mathcal{G})/\sqrt{2} \\
\end{split}
.
\end{equation}

We also included chroma maps $A, B$ from the most widely used CIELAB perceptual color space \cite{kundu2017no, ghadiyaram2017perceptual}, which can be converted from RGB via CIEXYZ \cite{wiki:CIELAB}. Note that we extract both chroma maps as well as their corresponding gradient maps (via Eq. (\ref{eq:gm})) following the suggestions in \cite{kundu2017no}. The above defined luma and chroma feature transforms are visualized in Fig. \ref{fig:feat_maps}.

Images are naturally multiscale, and distortions
affect image structures across scales. Incorporating multiscale information in quality models provides significant performance improvements \cite{ mittal2012no,wang2003multiscale}. Hence, we extract NSS-34 features from the four luma feature maps (Y, GM, LoG, DoG) at two scales: the original image scale and a reduced (by a factor two) resolution. However, we only extract NSS-34 features at the half scale on the twelve chroma feature maps, since frames are often compressed in YUV420 format, which already contain chroma information in reduced scale; additionally, it has also been observed that humans are more sensitive to luma distortions than chroma distortions \cite{chen2020chroma}. To sum up, the entire collection of spatial features is collected by applying two-scales NSS-34 to the luma feature maps and single-scale NSS-34 to the chromatic maps, yielding a total of $34\times 2\times 4+34\times 1\times 12=680$ features. It is worth noting that this 680-dim spatial model is an improved alternative to the SOTA 560-dim FRIQUEE model \cite{ghadiyaram2017perceptual} since it achieves comparable performance as FRIQUEE, but is \textbf{20x} faster.

\subsection{Temporal Features}
\label{ssec:temporal_features}

Prior BVQA methods accounting for temporal distortions, however, either rely on expensive motion estimation \cite{saad2014blind, korhonen2019two}, or underperform on UGC videos by only accounting for simple frame-difference statistics \cite{mittal2015completely, sinno2019spatio, yu2020predicting}, even including complex CNN models \cite{li2019quality, kim2018deep}. Here we attempt to exploit more general temporal scene statistics of natural videos to develop and improve BVQA models. To the best of our knowledge, we propose the \textit{first general, effective and efficient} temporal statistics model based on bandpass regularities of natural videos along the time dimension, going beyond simpler frame-difference models \cite{soundararajan2012video}.

Inspired by the efficacy of temporal bandpass statistics in the prediction of frame rate-dependent video quality \cite{madhusudana2020st}, our proposed temporal model utilizes 1D temporal bandpass representations. Specifically, consider a bank of $K$ temporal bandpass filters denoted $h_k,\ k\in \{0,...,K-1\}$, where $k$ denotes the subband index. The temporal bandpass responses of a video $F(\mathbf{x},t)$ (where $\mathbf{x}=(x,y)$ and $t$ represents spatial and temporal co-ordinates, respectively) is
\begin{equation}
\label{eq:temp_bandpass}
Y_k(\mathbf{x},t)=F(\mathbf{x},t)*h_k(t)\ \ k=0,...,K-1,
\end{equation}
where $*$ and $Y_k$ are 1D temporal convolution operations and the bandpass response of the $k^{th}$ filter, respectively. Note that frame differences are a special case of Eq.~\eqref{eq:temp_bandpass} (the high-pass component of a 2-tap Haar wavelets). Fig. \ref{fig:haar} visually illustrates the bandpass responses of a natural video from the LIVE-VQA dataset \cite{seshadrinathan2010study} using 3-level Haar wavelet filters. 

Attempting to generalize the spatial NSS as mentioned in Sec. \ref{ssec:nss_model} to the temporal domain, we instead analyze the statistics of the temporal bandpass coefficients $Y_k(\mathbf{x},t)$, $k=1,...,7$ (ignoring the lowest band $k=0$) by again applying MSCN transforms, as in Eq.~\eqref{eq:mscn}, to further decorrelate the subband representations, over a set of frame time samples $t\in\{t_0,t_1,...,t_N\}$ (note that $t$ does not need to be densely sampled). Note that in Eq. (\ref{eq:mscn}), $\mu_k(\mathbf{x},t)$ and $\sigma_k(\mathbf{x},t)$ are replaced by the local mean and standard deviation within a spatial window centered at location $(\mathbf{x},t)$, for each subband $k$.

\begin{figure}[!t]
\centering
\footnotesize
\def\xheight{0.485}
\setlength{\tabcolsep}{1pt}
\begin{tabular}{cc}
\includegraphics[width=\xheight\linewidth]{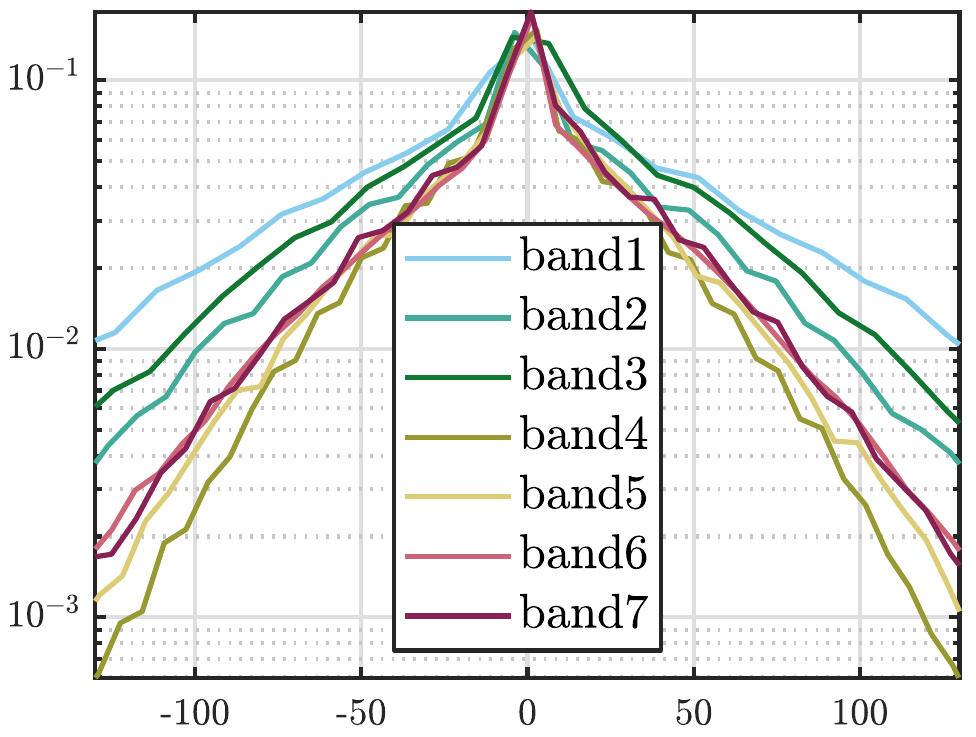} &
\includegraphics[width=\xheight\linewidth]{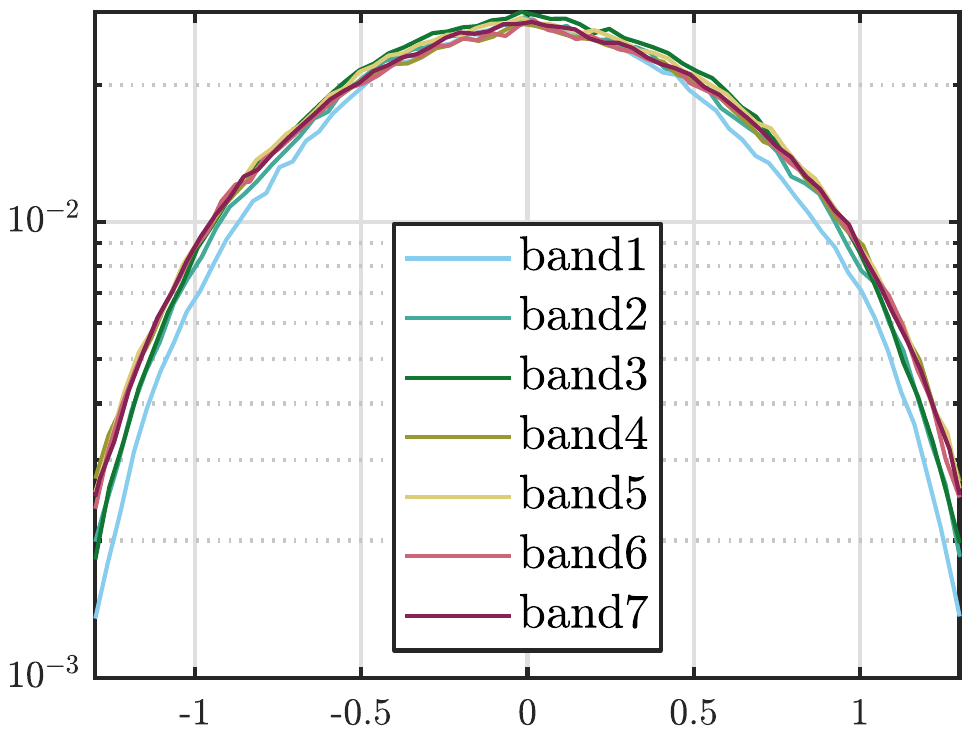} \\
\end{tabular}
\caption{Histograms of raw subband (left) and the corresponding MSCN normalized coefficients (right) of a natural video \texttt{Tractor}, where the normalized coefficients exhibit homogeneous regularities across bands.}
\label{fig:temp_stats}
\end{figure}

We have found that the MSCN coefficients of the temporal bandpass coefficients of natural videos also exhibit a Gaussian-like appearance, as shown in Fig. \ref{fig:temp_stats}, while the regularities are modified by the presence of distortion, strongly suggesting the possibility of quantifying deviations to predict perceived video quality. We model the distributions of subband MSCN coefficients again using the pre-defined GGD and AGGD distributions, by merely passing them into the NSS-34 feature extractor (Sec. \ref{ssec:spatial_features}). Similar to the spatial feature processing, we also extract the temporal statistical features over two scales (original and half scale), yielding a 476-dim feature vector (($34$ features/band)$\times$(7 subbands)$\times$(2 scales) $=476$).

\begin{table}[!t]
\footnotesize
\setlength{\tabcolsep}{4pt}
\caption{Summary of the tested BVQA datasets.}
\label{table:dataset}
\centering
\begin{tabular}{ lrrrlrr }
\toprule
Database & \# Vid  & Reso & Sec &  Label & Range   \\ \midrule
KoNViD-1k'17 \cite{hosu2017konstanz}  & 1,200 & 540p & 8  & MOS+$\sigma$ & [1,5]  \\
 LIVE-VQC'18 \cite{sinno2018large}  & 585   & 1080p-240p & 10  & MOS & [0,100]  \\
  YouTube-UGC'20 \cite{wang2019youtube}  & 1,380  & 4k-360p & 20 & MOS+$\sigma$ & [1,5] \\
 \bottomrule
\end{tabular}
\end{table}

Inspired by the efficacy of standard deviation pooling as first introduced in GMSD \cite{xue2013gradient} and later also shown effective when utilized for temporal pooling in \cite{korhonen2019two, tu2020ugc}, we calculate the 680 spatial features at two frames per second within each non-overlapping one-second chunk, then enrich the feature set by applying average and absolute difference pooling \cite{tu2020comparative} of the frame features within each chunk, based on the hypothesis that the variation of spatial features also correlates with the temporal properties of the video. Finally, all of the chunk-wise feature vectors are average pooled across all chunks to derive the final set of features over the entire video.

\begin{table*}[!t]
\setlength{\tabcolsep}{4pt}
\renewcommand{\arraystretch}{1.}
\centering
\footnotesize
\caption{Performance comparison of the evaluated BVQA models on the four BVQA datasets. The \underline{\textbf{underlined}} and \textbf{boldfaced} entries indicate the best and top three performers on each database for each performance metric, respectively.}
\label{table:eval_svr}
\begin{threeparttable}
\begin{tabular}{rccccccccccccccccccccc}
\toprule
\textsc{Dataset} & & \multicolumn{3}{c}{KoNViD-1k \cite{hosu2017konstanz}} & & \multicolumn{3}{c}{LIVE-VQC \cite{sinno2018large}} & & \multicolumn{3}{c}{YouTube-UGC \cite{wang2019youtube}} & & 
\multicolumn{3}{c}{All-Combined \cite{tu2020ugc}}
\\ \cline{3-5}\cline{7-9}\cline{11-13}\cline{15-17}\\[-1.em]
\textsc{Model} &
& \textsc{SRCC$\uparrow$} & \textsc{PLCC$\uparrow$} & \textsc{RMSE$\downarrow$} &
& \textsc{SRCC$\uparrow$} & \textsc{PLCC$\uparrow$} & \textsc{RMSE$\downarrow$} &
& \textsc{SRCC$\uparrow$} & \textsc{PLCC$\uparrow$} & \textsc{RMSE$\downarrow$} &
& \textsc{SRCC$\uparrow$} & \textsc{PLCC$\uparrow$} & \textsc{RMSE$\downarrow$}
\\ \hline\\[-1.em]
BRISQUE \cite{mittal2012no} &
& 0.6567 & 0.6576 & 0.4813 & & 0.5925 & 0.6380 & 13.100 &
& 0.3820 & 0.3952 & 0.5919 & & 0.5695 & 0.5861 & 0.5617 \\
GM-LOG \cite{xue2014blind} &
& 0.6578 & 0.6636 & 0.4818 & & 0.5881 & 0.6212 & 13.223 &
& 0.3678 & 0.3920 & 0.5896 & & 0.5650 & 0.5942 & 0.5588 \\
HIGRADE \cite{kundu2017no} &
& 0.7206 & 0.7269 & 0.4391 & & 0.6103 & 0.6332 & 13.027 & 
& 0.7376 & 0.7216 & 0.4471 & & 0.7398 & 0.7368 & 0.4674 \\
FRIQUEE \cite{ghadiyaram2017perceptual} &
& 0.7472 & 0.7482 & 0.4252 & & 0.6579 & 0.7000 & 12.198 &
& \textbf{0.7652} & {\textbf{{0.7571}}} & {\textbf{{0.4169}}} & & \textbf{0.7568} & {\textbf{0.7550}} & {\textbf{0.4549}} \\
CORNIA \cite{ye2012unsupervised} &
& 0.7169 & 0.7135 & 0.4486 & & 0.6719 & 0.7183 & 11.832 &
& 0.5972 & 0.6057 & 0.5136 & & 0.6764 & 0.6974 & 0.4946 \\
HOSA \cite{xu2016blind} &
& 0.7654 & 0.7664 & 0.4142 & & 0.6873 & 0.7414 & 11.353 &
& 0.6025 & 0.6047 & 0.5132 & & 0.6957 & 0.7082 & 0.4893 \\
KonCept512 \cite{hosu2020koniq} &
& 0.7349 & 0.7489 & 0.4260 & & 0.6645 & 0.7278 & 11.626 &
& 0.5872 & 0.5940 & 0.5135 & & 0.6608 & 0.6763 & 0.5091 \\
PaQ-2-PiQ \cite{ying2019patches} &
& 0.6130 & 0.6014 & 0.5148 & & 0.6436 & 0.6683 & 12.619 &
& 0.2658 & 0.2935 & 0.6153 & & 0.4727 & 0.4828 & 0.6081 \\\hline\\[-1.em]
V-BLIINDS \cite{saad2014blind} &
& 0.7101 & 0.7037 & 0.4595 & & 0.6939 & 0.7178 & 11.765 &
& 0.5590 & 0.5551 & 0.5356 & & 0.6545 & 0.6599 & 0.5200 \\
TLVQM \cite{korhonen2019two} &
& 0.7729 & 0.7688 & 0.4102 & & {\textbf{\underline{0.7988}}} & {\textbf{\underline{0.8025}}} & {\textbf{\underline{10.145}}} &
& 0.6693 & {0.6590} & {0.4849} & & 0.7271 & 0.7342 & 0.4705  \\
VMEON \cite{liu2018end} & 
&  0.1118 & 0.1958 & 0.6322 &  
&  0.4024 & 0.4088 & 15.524 &
& 0.0634 & 0.1100 & 0.6304 &
& 0.2578 & 0.2594 & 0.6657
\\
VSFA \cite{li2019quality} &
& 0.7728$^*$ & 0.7754$^*$ & 0.4205$^*$ & & 0.6978$^*$ & 0.7426$^*$ & 11.649$^*$ &
& - & - & - & & - & - & - \\
MDVSFA \cite{li2020unified} &
& \textbf{0.7812}$^*$ & \textbf{0.7856}$^*$ & - & & 0.7382$^*$ & \textbf{0.7728}$^*$ & - &
& - & - & - & & - & - & - \\
VIDEVAL \cite{tu2020ugc} &
& \textbf{0.7832} & \textbf{0.7803} & \textbf{0.4026} & & \textbf{0.7522} & 0.7514 & \textbf{11.100} &
& \textbf{\underline{0.7787}} &  \textbf{\underline{0.7733}} & \textbf{\underline{0.4049}} & & {\textbf{{0.7960}}} & {\textbf{{0.7939}}} & {\textbf{0.4268}} \\\hline\\[-1.em]
RAPIQUE & 
& \textbf{\underline{0.8031}} & \textbf{\underline{0.8175}} & \textbf{\underline{0.3623}} & & \textbf{0.7548} & \textbf{0.7863} & \textbf{10.518} &  
& \textbf{0.7591} & \textbf{0.7684} & \textbf{0.4060} & & \textbf{\underline{0.8070}} & \textbf{\underline{0.8229}} & \textbf{\underline{0.3968}} \\
\bottomrule
\end{tabular}
\begin{tablenotes}
    \item $^*$The results are cited from experiments reported in their original papers 
  \end{tablenotes}
\end{threeparttable}
\end{table*}

\begin{figure*}[!t]
\centering
\footnotesize
\def\xheight{0.245}
\setlength{\tabcolsep}{1pt}
\begin{tabular}{cccc}
\includegraphics[width=\xheight\textwidth]{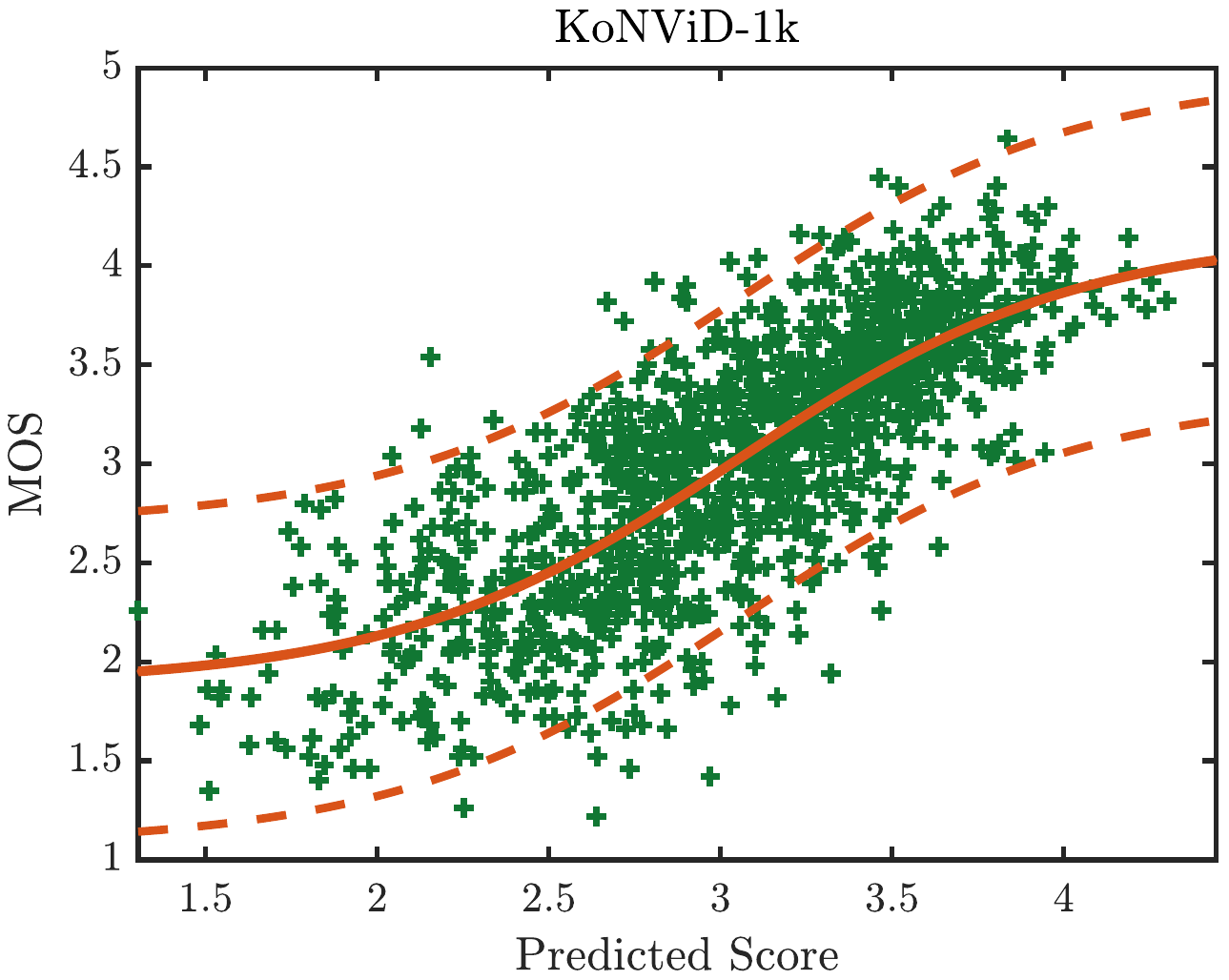} &
\includegraphics[width=\xheight\textwidth]{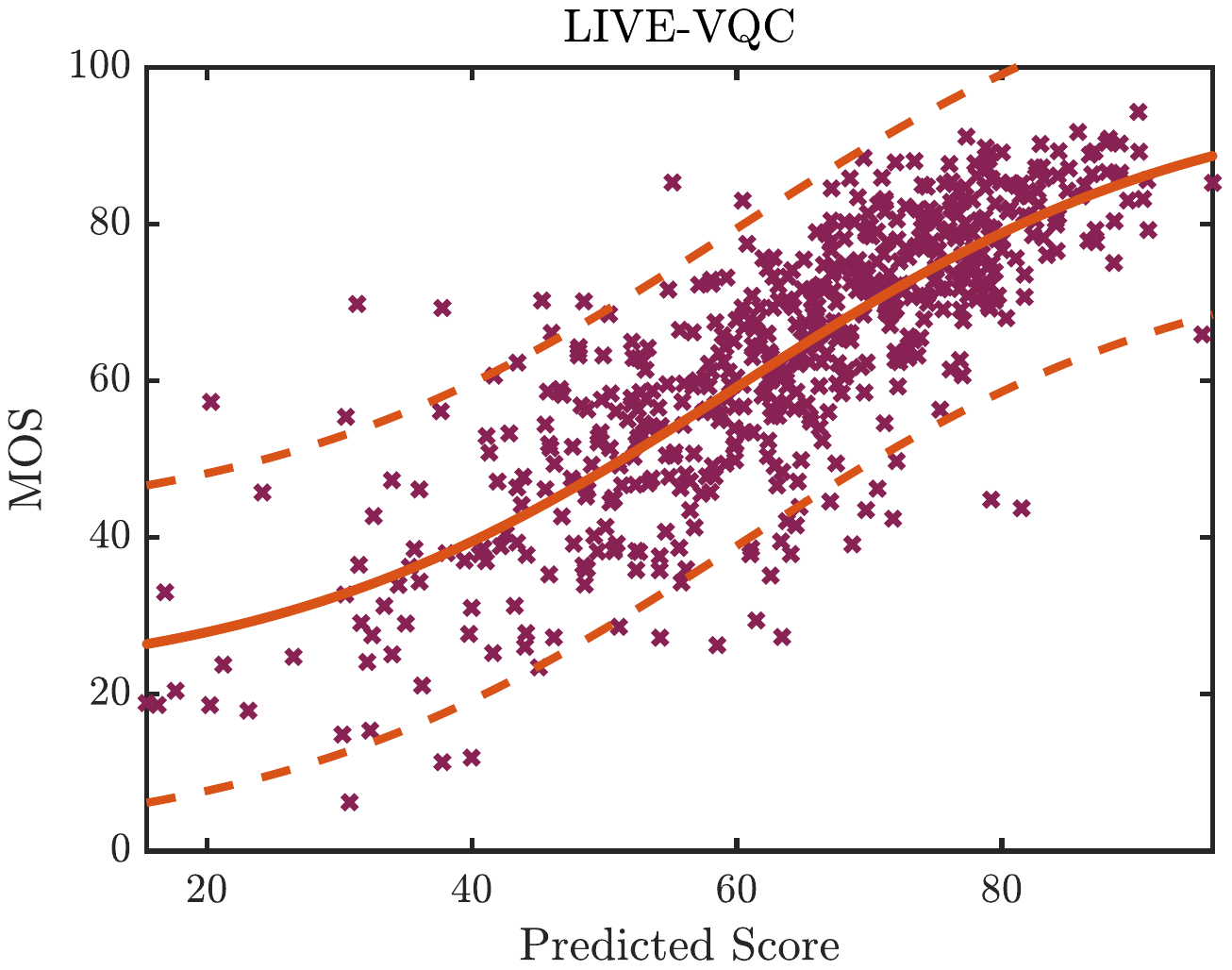} & 
 \includegraphics[width=\xheight\textwidth]{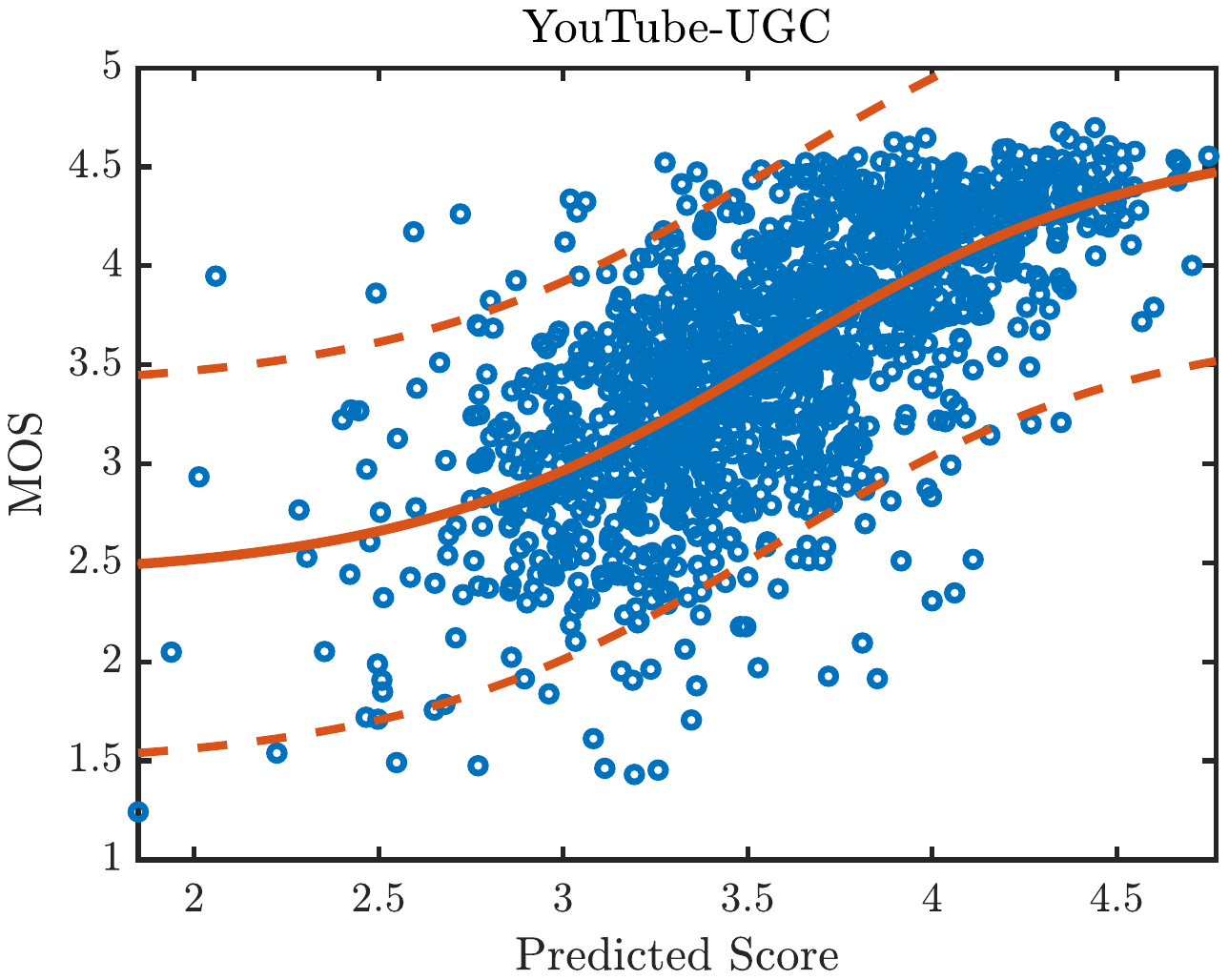} &
\includegraphics[width=\xheight\textwidth]{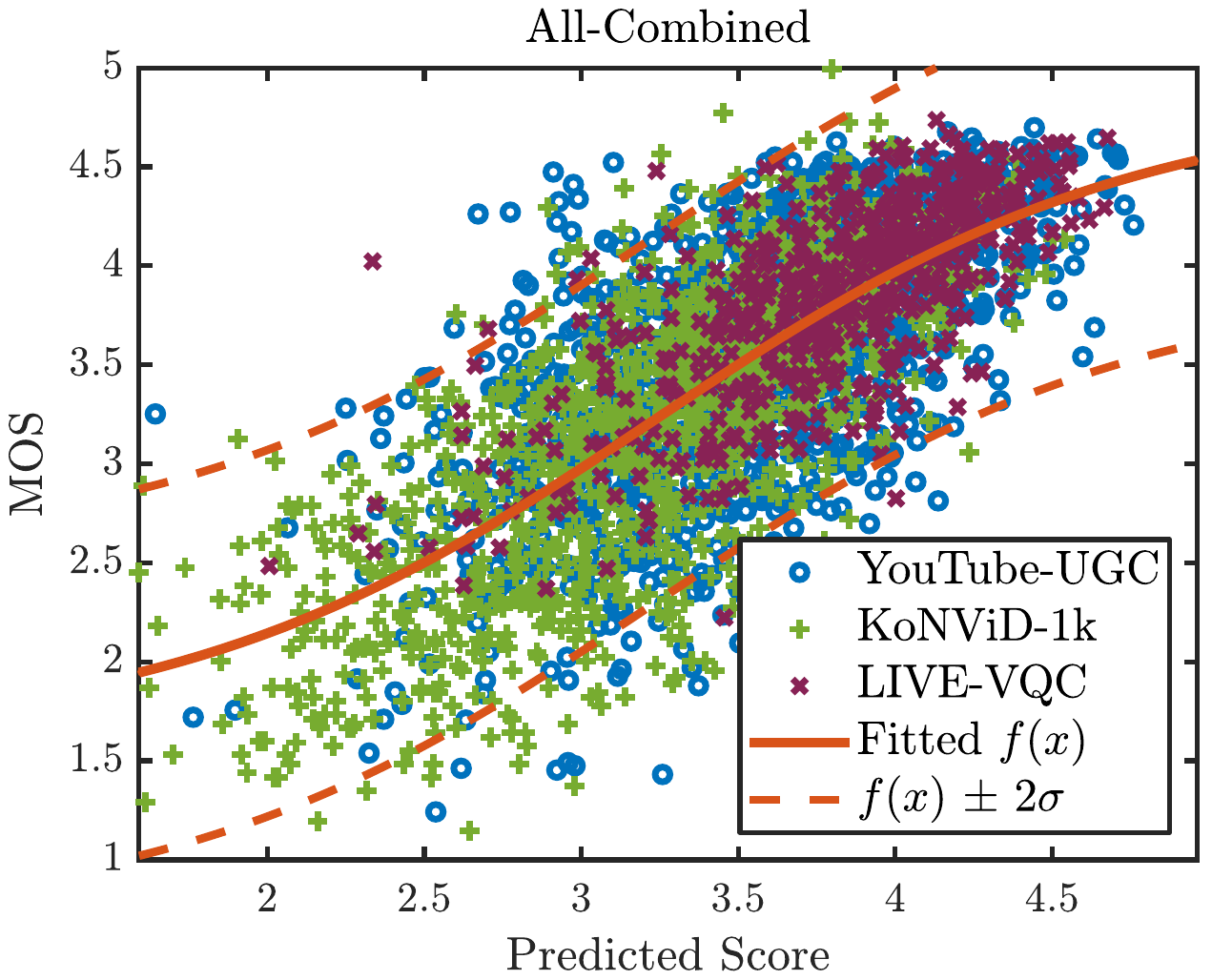} \\
\multicolumn{4}{c}{(a) TLVQM} \\
\includegraphics[width=\xheight\textwidth]{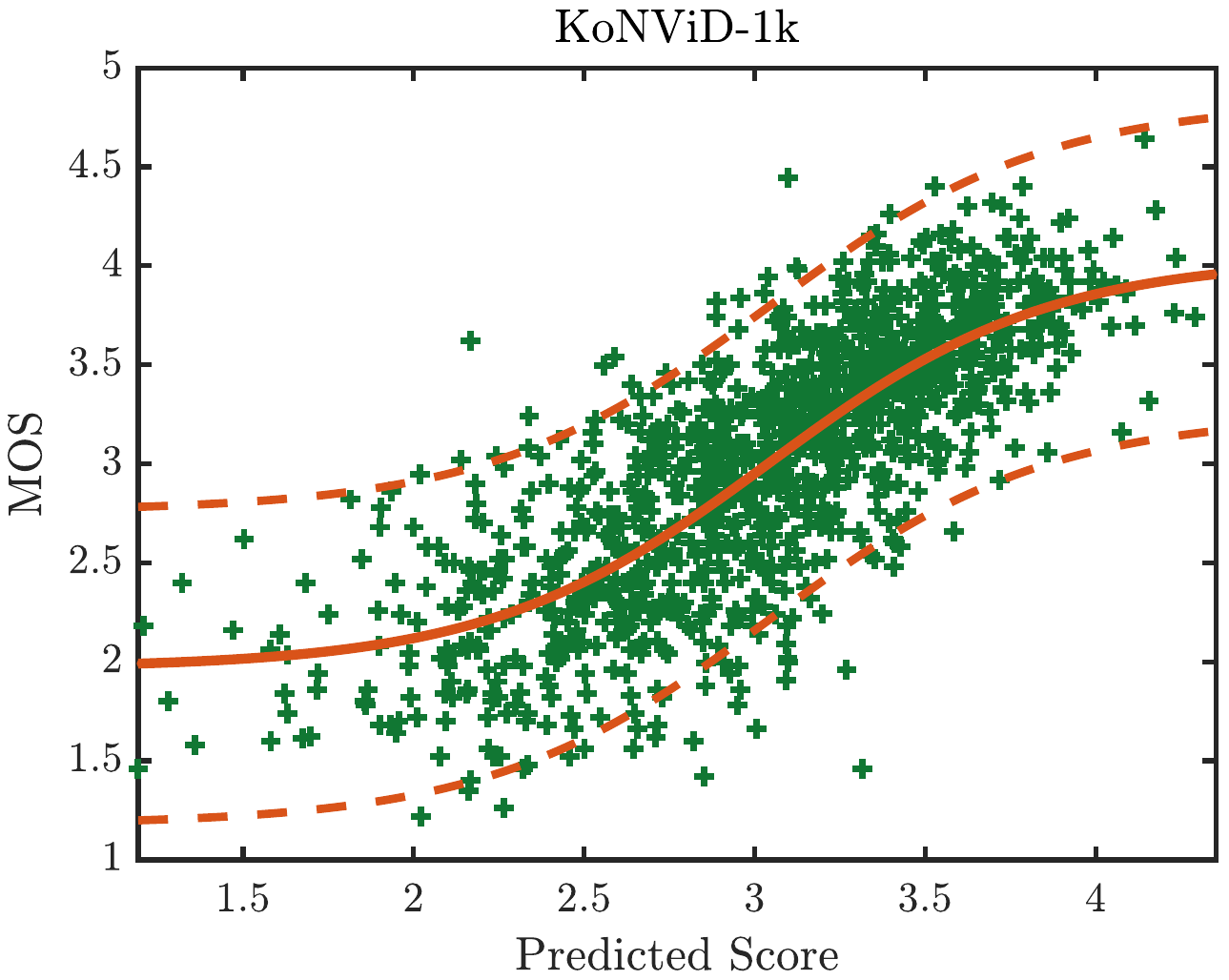} &
\includegraphics[width=\xheight\textwidth]{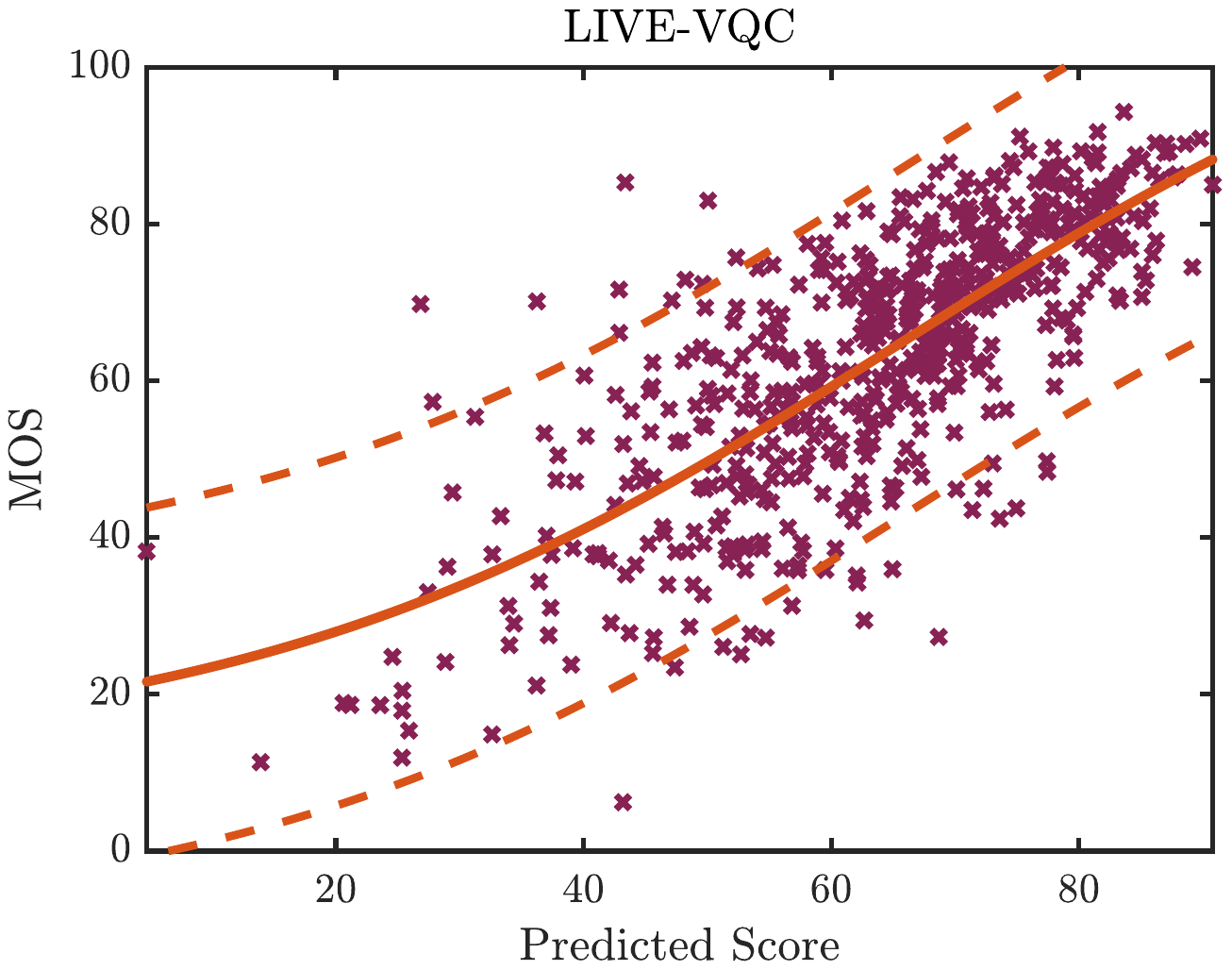} & 
 \includegraphics[width=\xheight\textwidth]{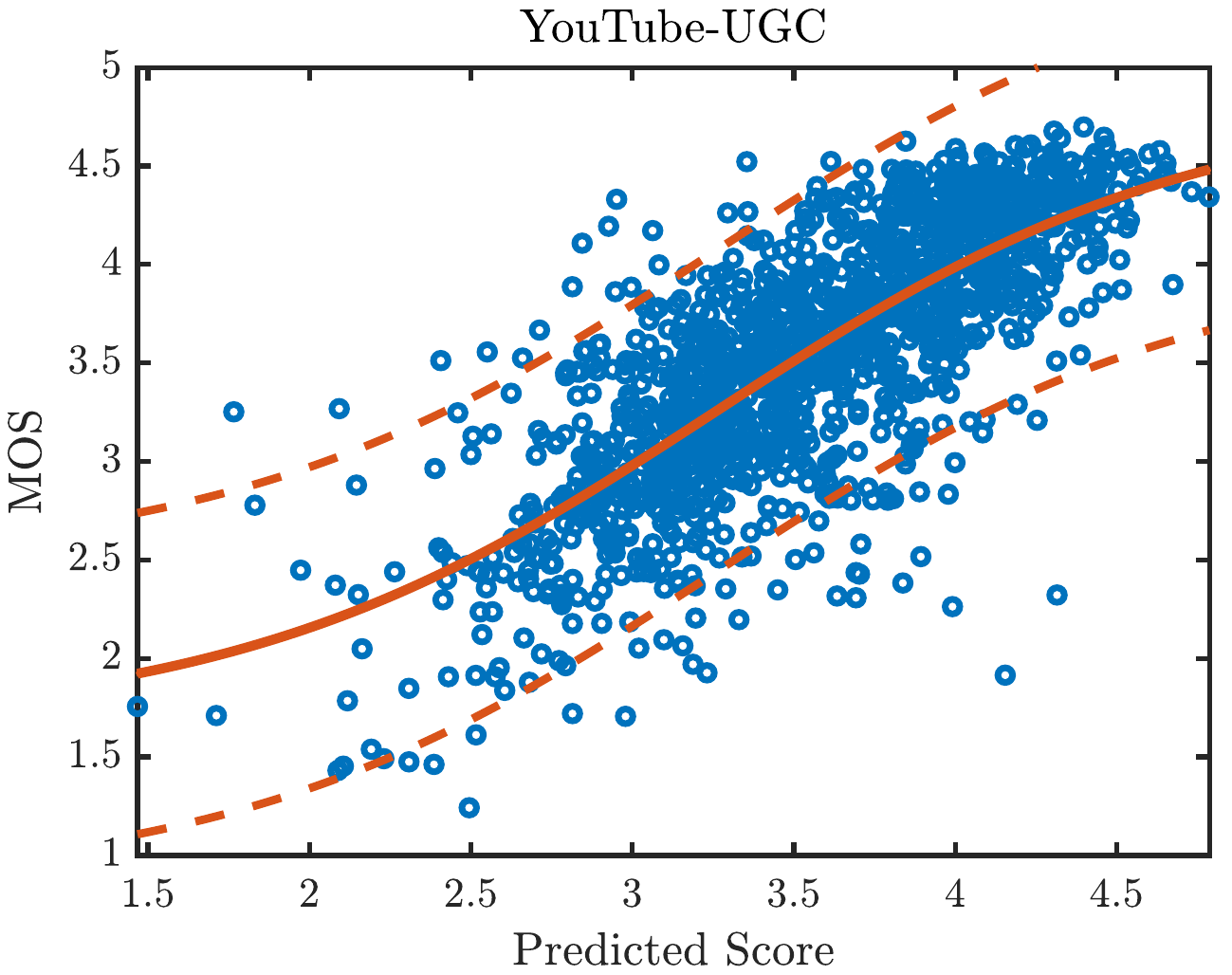} &
\includegraphics[width=\xheight\textwidth]{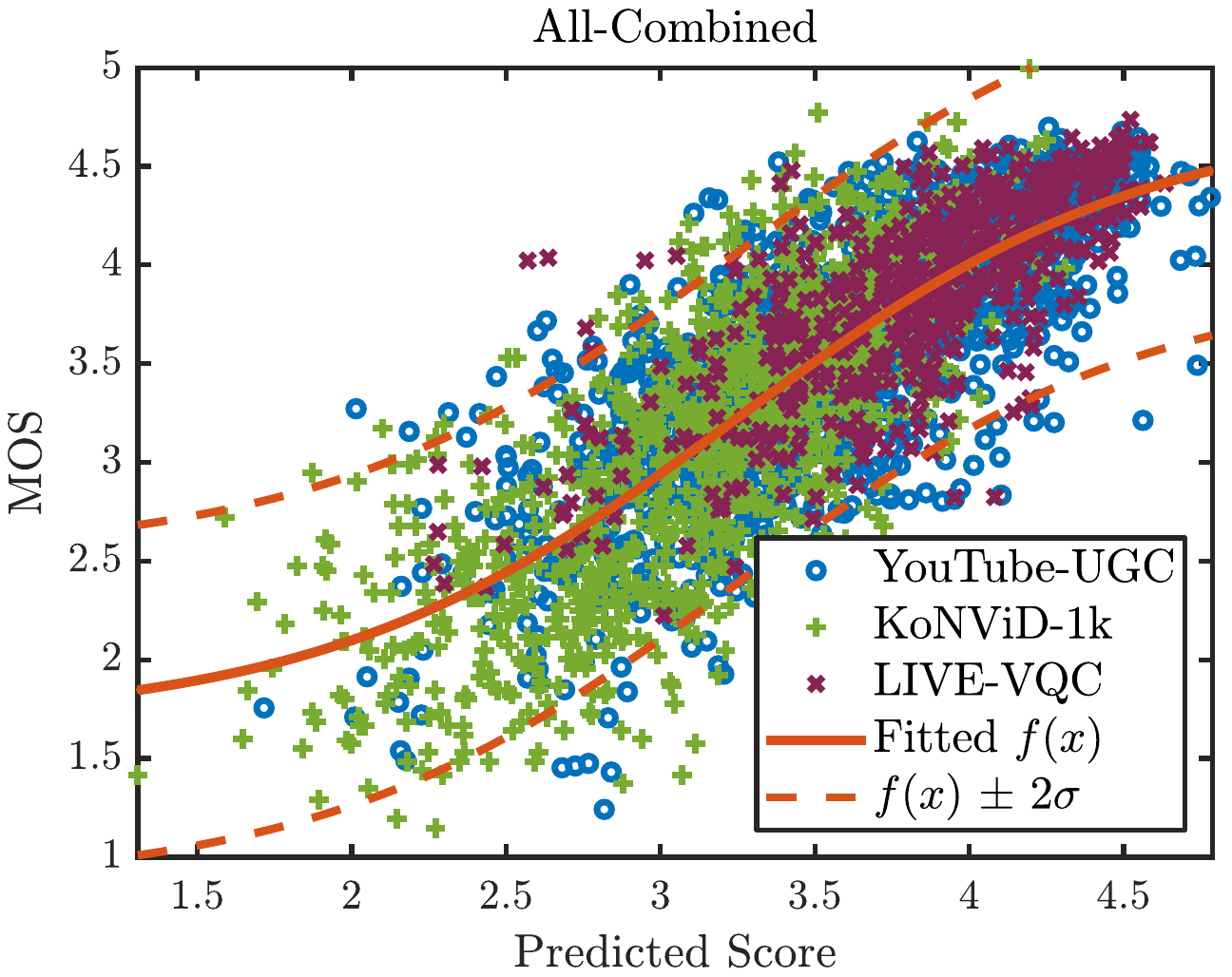} \\
\multicolumn{4}{c}{(b) VIDEVAL} \\
\includegraphics[width=\xheight\textwidth]{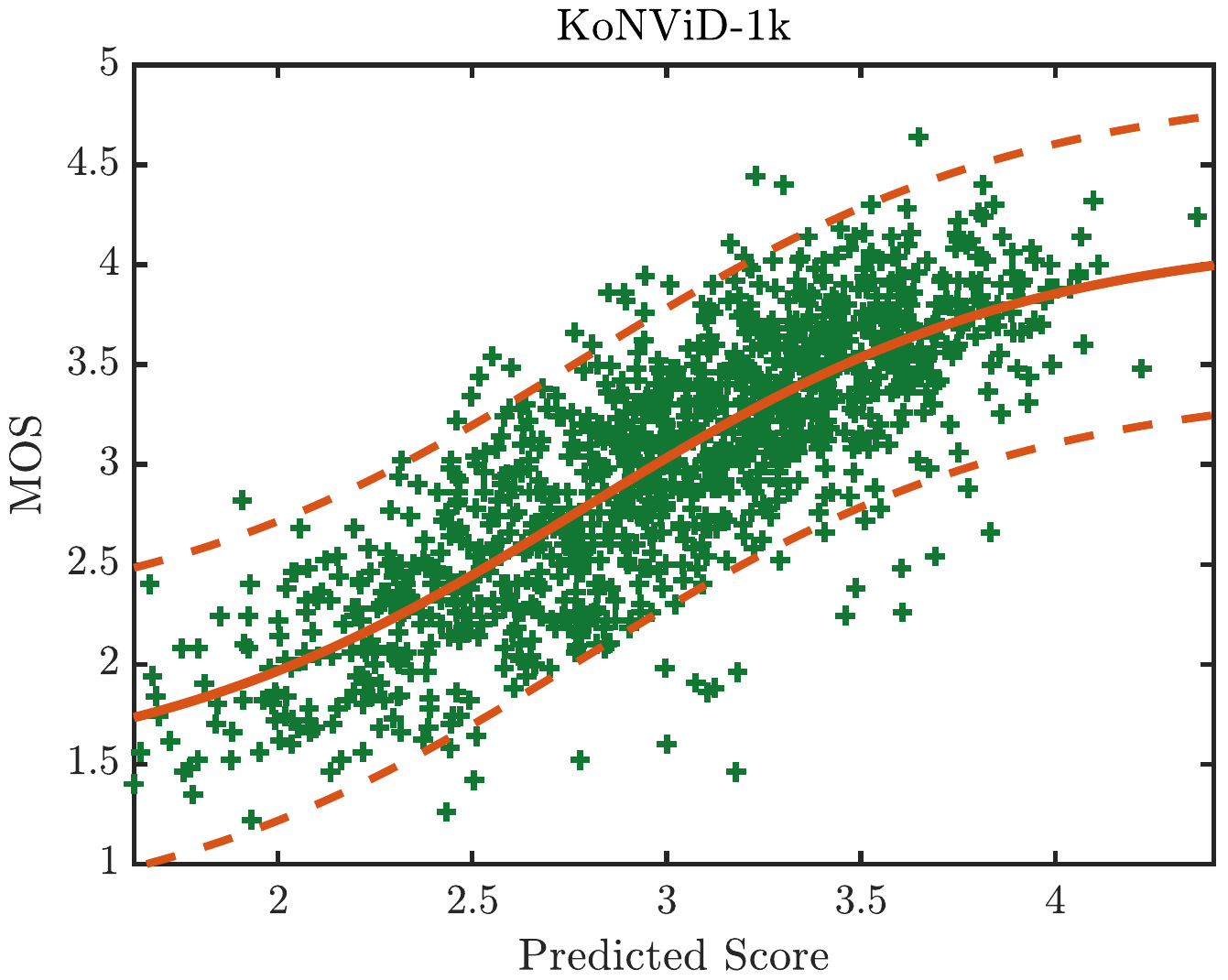} &
\includegraphics[width=\xheight\textwidth]{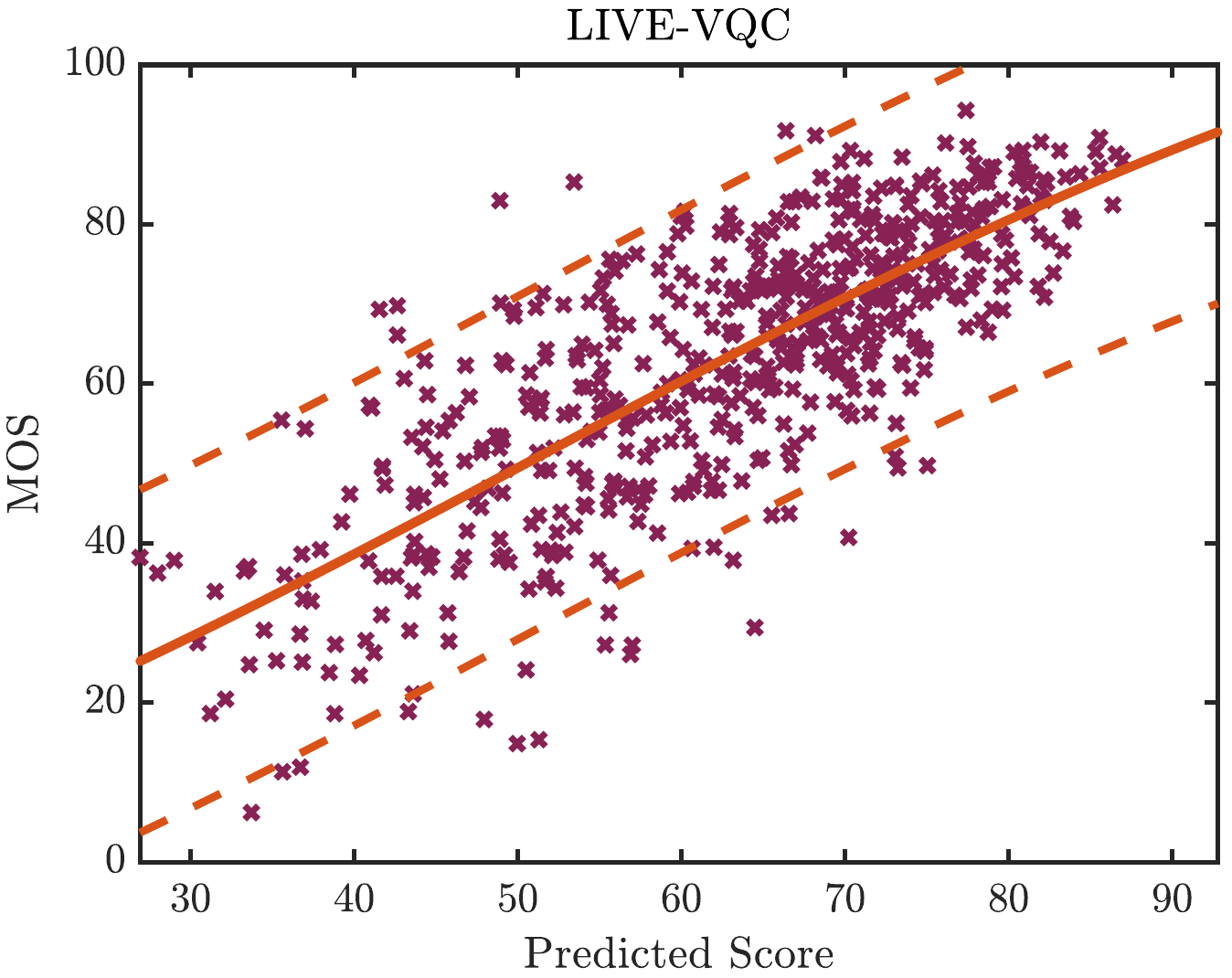} & 
 \includegraphics[width=\xheight\textwidth]{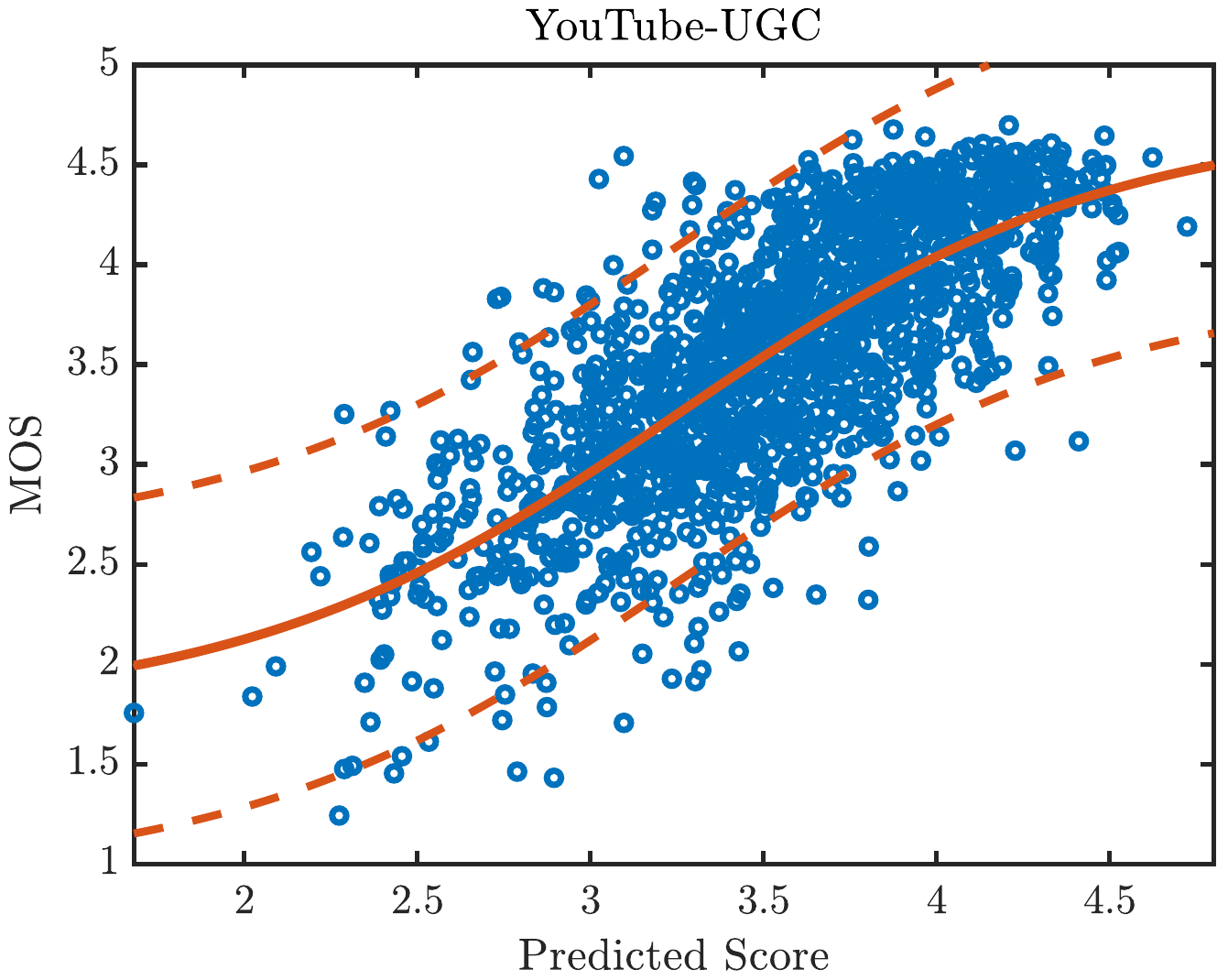} &
\includegraphics[width=\xheight\textwidth]{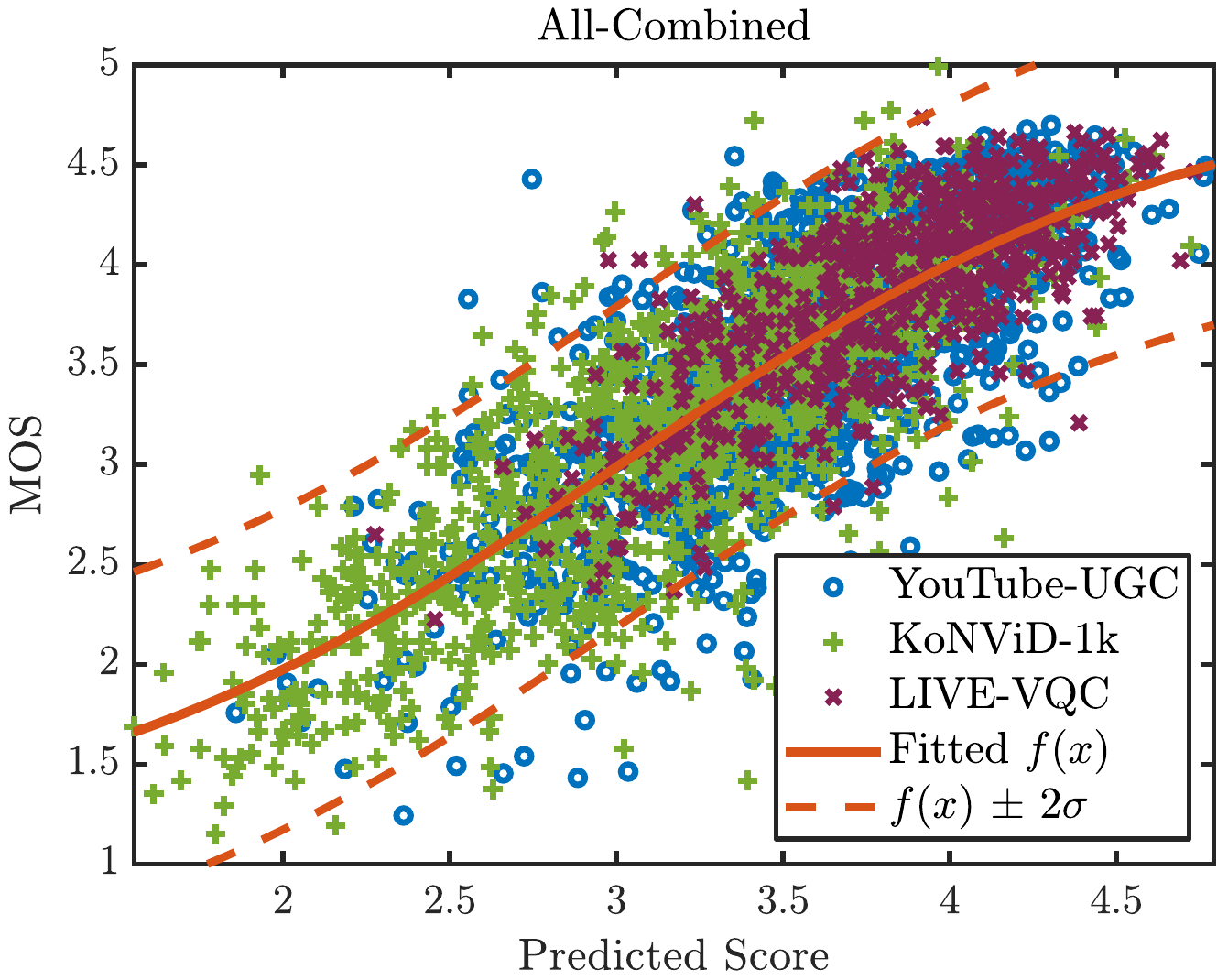} \\
\multicolumn{4}{c}{(c) RAPIQUE} \\
\end{tabular}
\caption{Scatter plots and nonlinear logistic fitted curves of (c) RAPIQUE versus MOS, compared against (a) TLVQM \cite{korhonen2019two} and (b) VIDEVAL \cite{tu2020ugc}, using a grid-search SVR using $k$-fold cross-validation on KoNViD-1k \cite{hosu2017konstanz}, LIVE-VQC \cite{sinno2018large}, YouTube-UGC \cite{wang2019youtube}, and the All-Combined set (Sec. \ref{ssec:experiment_settings}), respectively.}
\label{fig:scatter_plot}
\end{figure*}

\subsection{Deep Learning Features}
\label{ssec:deep_learning_features}

CNN-based solutions have been observed to generally perform well on UGC picture quality problems \cite{hosu2020koniq, ying2019patches, zhang2018blind} thanks to several recently released large-scale picture quality datasets \cite{hosu2020koniq, ying2019patches, fang2020perceptual}. Still, none of them have proven effective on UGC video quality databases \cite{sinno2018large, hosu2017konstanz, wang2019youtube}. However, the authors of \cite{tu2020ugc} have shown that the simple feature vector from an FC-layer, without fine-tuning, to be a useful quality indicator if training a shallow regressor on top. Therefore, we, \textit{for the first time}, propose to leverage the best of both worlds, by combining powerful quality-aware NSS features as described in Sec. \ref{ssec:nss_model}, \ref{ssec:spatial_features}, \ref{ssec:temporal_features}, with pre-trained deep learning features, by jointly training a regressor on them to predict the final quality score.

One issue encountered when dealing with quality prediction problems is the mismatch of picture sizes between the standard inputs of CNN models such as VGG-16 \cite{simonyan2014very}, ResNet-50 \cite{he2016deep}, and IQA-valid high-resolution images. Two possible solutions have been attempted to solve this. The authors of \cite{kim2017deep, ying2019patches} suggested applying a CNN on spatially sampled small patches, then aggregating the locally predicted scores to obtain global quality scores. The authors of \cite{hosu2020koniq} presented a CNN operating on full-sized images, but with either global average pooling (GAP) or spatial pyramid pooling (SPP) \cite{he2015spatial}, feeding FC layers. These two schemes, however, increase the computation load of the CNN models. Since our proposed model is already armed with powerful spatial and temporal quality-aware features, we added CNN features only to exploit its ability to capture high-level semantic information, supplementing the low-level NSS features. In this regard, we aggressively downscaled the frames to fit the CNN model inputs when extracting these semantic-aware features, yielding greater efficiency than previous CNN VQA models. Another reason to use a pre-trained CNN without fine-tuning is to prevent overfitting, since existing video quality datasets are of limited sizes. In our implementation, we used a ResNet-50 (2,048-dim) as a semantic feature extractor.

\subsection{Learning a Video Quality Predictor}
\label{ssec:predictor}

We summarize the feature extraction process as follows. Since our goal is to build an efficient BVQA model, we devised spatial and temporal sampling strategies to further improve its speed. Specifically, given an input video $F(\mathbf{x},t)$, RAPIQUE uniformly samples 2 frames per second, based on which the 680-dim spatial NSS features (in Sec. \ref{ssec:spatial_features}) are extracted, then average and absolute-difference pools these to obtain 680 spatial and 680 temporal variation features, respectively. RAPIQUE also uniformly samples 8 consecutive frames each second, then applies temporal Haar filter (Eq. (\ref{fig:haar})) to extract 7 bandpass responses, from which 476 features are calculated at each time sample. The above features are calculated at a higher resized resolution while maintaining the aspect ratio (we used 512p in our experiments). However, the CNN backbone (ResNet-50) operates on resized frames at a sparse temporal sampling of 1 frame/sec to attain an additional 2,048 features.

After obtaining all the spatial, temporal, and CNN features within each one-second chunk, we adopt a simple approach to concatenate them all into a totally 3884-dimensional feature vector for each video chunk, then average-pool each to obtain a single 3884 feature vector over the entire video. A shallow or deep regressor head can then be trained on the aggregated feature vector to predict the final video quality scores.

\section{Experiments}
\label{sec:experiments}

\subsection{Experiment Settings}
\label{ssec:experiment_settings}

\textbf{Datasets and Baselines:} We used three recent BVQA datasets as testbeds for the performance evaluations: KoNViD-1k \cite{hosu2017konstanz}, LIVE-VQC \cite{sinno2018large}, and YouTube-UGC \cite{wang2019youtube}, as summarized in Table \ref{table:dataset}. We also used the combined set (denoted All-Combined) as introduced in \cite{tu2020ugc} as an additional composite benchmark. The All-Combined dataset is simply the union of KoNViD-1k (1,200), LIVE-VQC (575), and YouTube-UGC (1,380) after MOS calibration:
\begin{equation}
\label{eq:mos_cal_kon}
{y_\mathrm{adj}=5-4\times\left[(5-y_\mathrm{org})/4\times1.1241-0.0993\right]}
\end{equation}
\begin{equation}
\label{eq:mos_cal_live}
{y_\mathrm{adj}=5-4\times\left[(100-y_\mathrm{org})/100\times0.7132+0.0253\right]}
\end{equation}
where equations (\ref{eq:mos_cal_kon}) and (\ref{eq:mos_cal_live}) are used to calibrate KoNViD-1k and LIVE-VQC, respectively (YouTube-UGC does not need to be changed). Here $y_\mathrm{adj}$ denotes the adjusted scores, while $y_\mathrm{org}$ is the original MOS. We refer the reader to \cite{tu2020ugc} for details regarding assumptions and derivations of the calibration process.

The baseline models used for comparison are BRISQUE \cite{mittal2012no}, GM-LOG \cite{xue2014blind}, HIGRADE \cite{kundu2017no}, FRIQUEE \cite{ghadiyaram2017perceptual}, the codebook-based models CORNIA \cite{ye2012unsupervised} and HOSA \cite{xu2016blind}, and the deep learning models, KonCept512 \cite{hosu2020koniq}, PaQ-2-PiQ \cite{ying2019patches} which are all spatial-only models. All the spatial models extract features at 1 fps, which were average-pooled to obtain the final video-level feature vector used for training. We also compared against three feature-based BVQA models, V-BLIINDS \cite{saad2014blind}, TLVQM \cite{korhonen2019two}, and VIDEVAL \cite{tu2020ugc}, and the deep learning-based models, V-MEON \cite{liu2018end} and VSFA \cite{li2019quality} as well as its enhanced version, MDVSFA \cite{li2020unified}. Since `completely blind' models such as NIQE \cite{mittal2012making} and VIIDEO \cite{mittal2015completely} were not observed to perform reasonably well on these natural video datasets \cite{tu2020ugc}, we did not include them.

\textbf{Evaluation Method:} We used a support vector regressor (SVR) as the back-end regression model to learn the feature-to-score mappings \cite{korhonen2019two, saad2014blind, kim2017deep, ghadiyaram2017perceptual, mittal2012no}. We optimized the SVR parameters $(C,\gamma)$ via a randomized grid-search on the training set. Following convention, we randomly split the dataset into training and test sets ($80\%/20\%$) over 20 iterations, and the overall median test performance was reported. All of the evaluated methods were implemented using the original release by the respective authors. Four performance metrics were used: the Spearman Rank-Order Correlation Coefficient (SRCC) and the Kendall Rank-Order Correlation Coefficient (KRCC) are non-parametric measures of prediction monotonicity, while the Pearson Linear Correlation Coefficient (PLCC) with corresponding Root Mean Square Error (RMSE) were computed to assess prediction accuracy. Note that PLCC and RMSE are computed after performing a nonlinear four-parametric logistic regression to linearize the objective predictions to be on the same scale as MOS \cite{seshadrinathan2010study}:
\begin{equation}
\label{eq:logistic}
f(x)=\beta_2+\frac{\beta_1-\beta_2}{1+\exp{(-x+\beta_3/|\beta_4|})}.
\end{equation}

\begin{figure}[!t]
\centering
\includegraphics[width=0.4\textwidth]{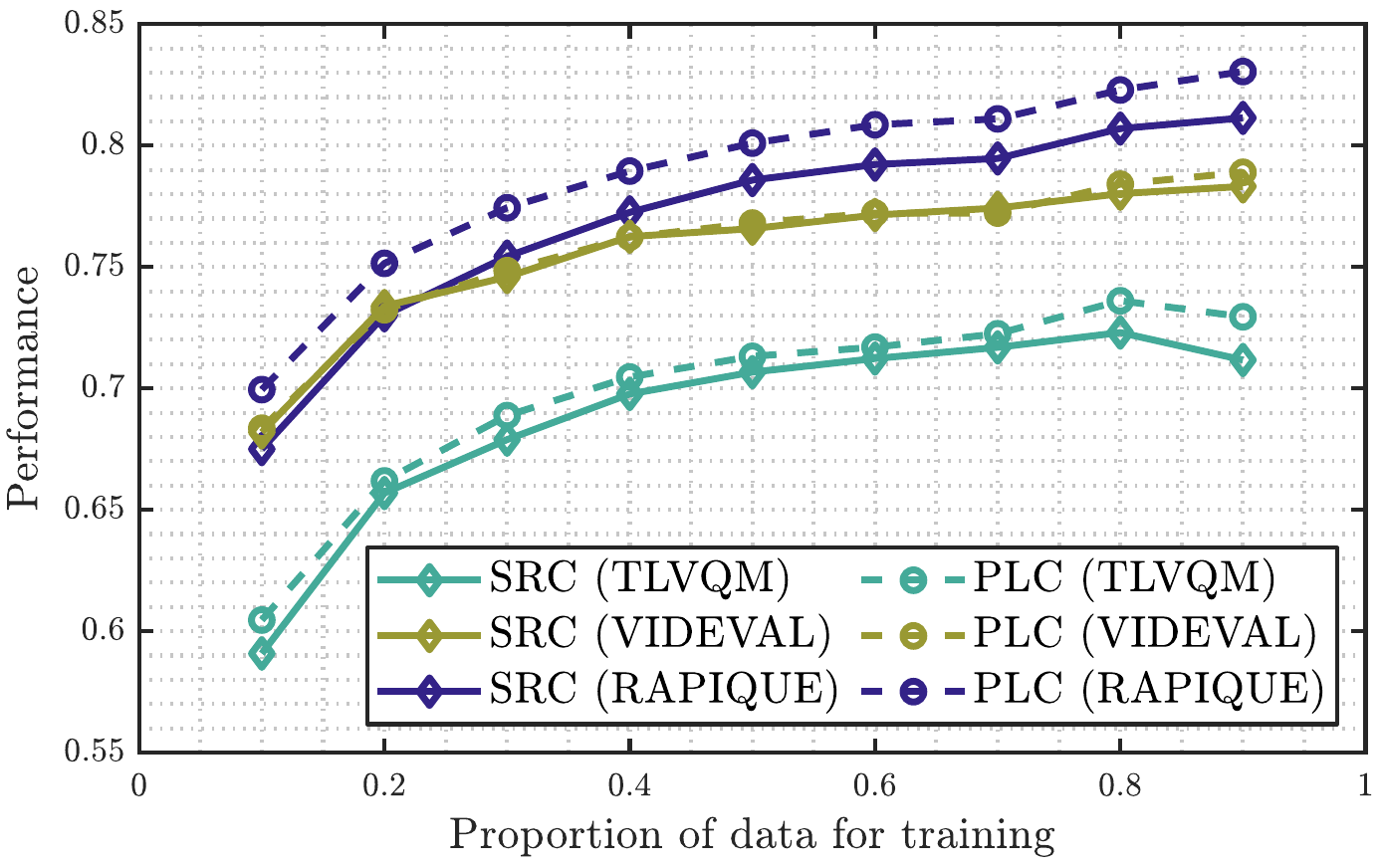}
\caption{Performance comparison of SRCC$/$PLCC as a function of the percentage of the content used to train the compared blind VQA models on the composite All-Combined set. Note that this result is self-explanatory as we used a slightly different evaluation method (20 iterations, SVR with randomized search cross-validation) compared to previous experiments.}
\label{fig:test_size}
\end{figure}

\subsection{Main Evaluation Results}
\label{ssec:main_evaluation_results}

Table \ref{table:eval_svr} shows the main comparison results on the four evaluated datasets. It may be observed that RAPIQUE achieved the best performance on KoNViD-1k, even outperforming the most recent, dense deep learning models such as VSFA and MDVSFA. On LIVE-VQC, which contains many mobile videos exhibiting large camera motions \cite{tu2020ugc}, TLVQM, which contains numerous heavily crafted motion-relevant features, was the best performer. However, RAPIQUE ranked a clear second, indicating that the temporal NSS features in RAPIQUE are powerful indications of temporal and motion-related distortions.

The most recent deep still picture quality models, KonCept512 and PaQ-2-PiQ, have been observed to perform poorly on UGC-VQA datasets. One reason for this is that these models were trained on picture quality datasets \cite{ying2019patches, hosu2020koniq}, containing strictly spatial content and distortions. A leading blind deep video quality model, V-MEON, also does not perform well, likely because it was trained on compression artifacts rather than on complex combinations of UGC distortions.

On the larger datasets, RAPIQUE delivered the second-best correlation against the subjective data on YouTube-UGC, only slightly worse than the current SOTA model VIDEVAL, while RAPIQUE ranked the best on the 3165-video composite set, All-Combined. Since VIDEVAL was created by a supervised feature selection process (using subjective labels) on the composite combined set, wherein YouTube-UGC accounts for a large portion, it would be expected to outperform on these two sets. The RAPIQUE model, on the contrary, is database-agnostic, and also exhibited uniformly well performance on all four test sets. In this regard, RAPIQUE has the potential to perform better on future larger-scale datasets and in real-world application scenarios it has not been exposed to. The scatter plots and fitted curves of RAPIQUE predictions versus MOS in Fig. \ref{fig:scatter_plot} visually demonstrate that the performance of RAPIQUE remains stable on video sequences from different databases, achieving smaller RMSE on larger databases.

\begin{figure}[!t]
\centering
\footnotesize
\begin{tabular}{c}
     \includegraphics[width=0.4\textwidth]{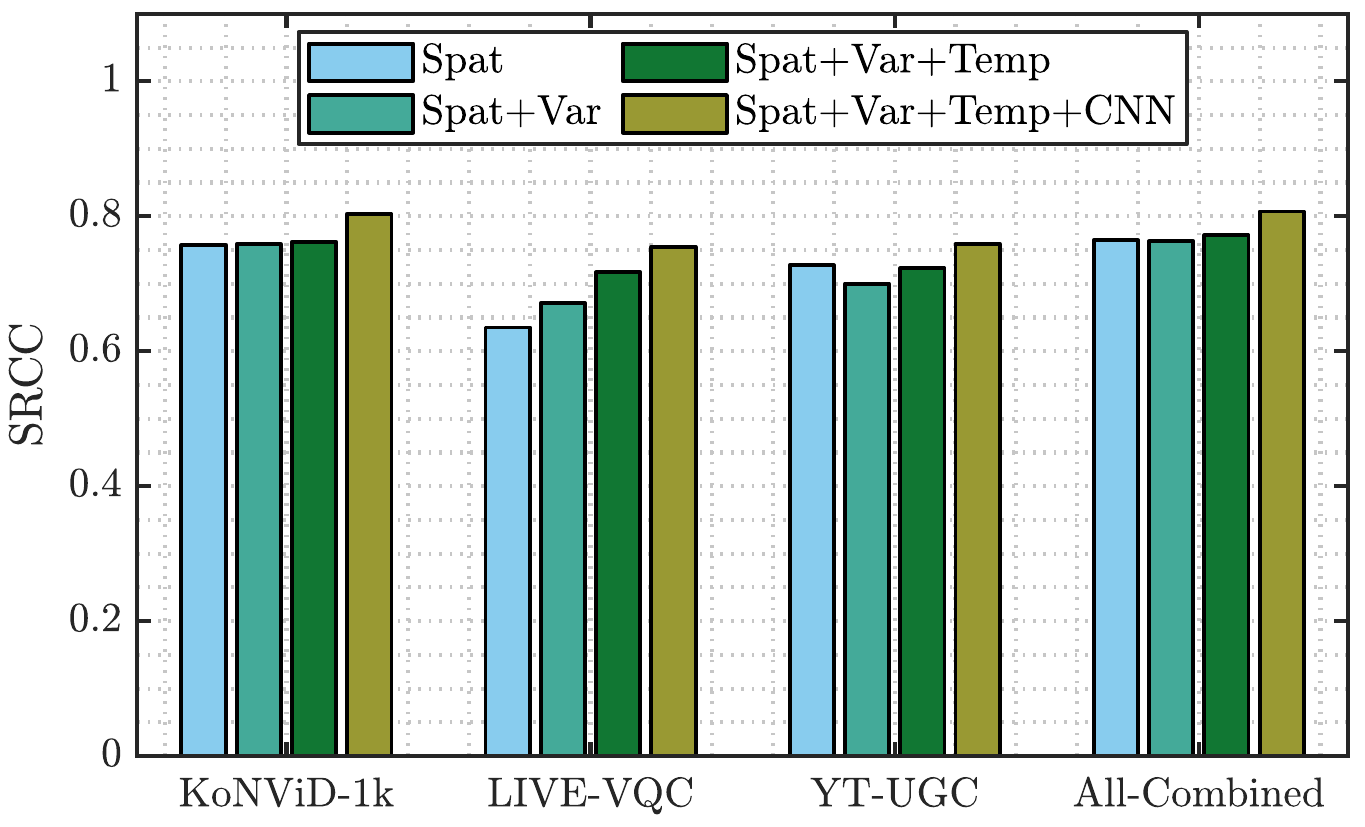}  \\
     (a)
     \\
     \includegraphics[width=0.32\textwidth]{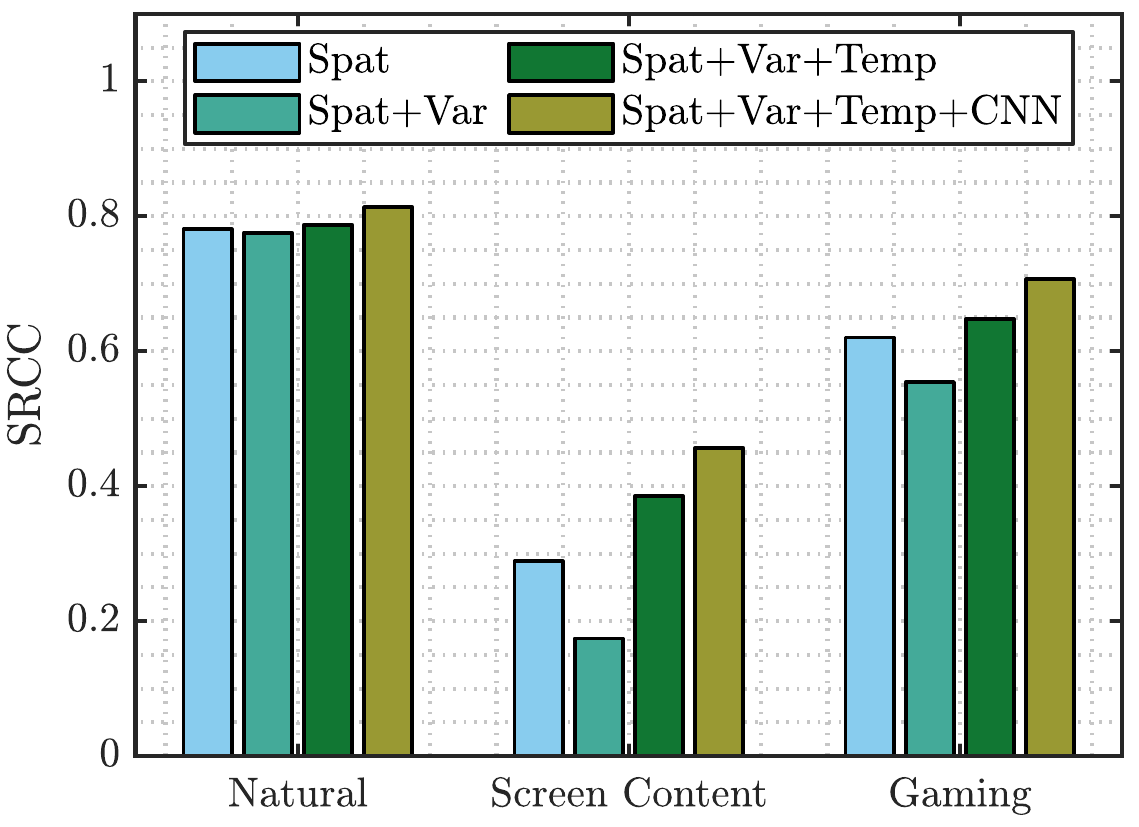} \\
     (b) 
     \\
\end{tabular}

\caption{Ablation study of RAPIQUE on (a) four benchmarks and (b) different content types - \textit{Spat} denotes the spatial model (Sec. \ref{ssec:spatial_features}), \textit{Var} is the temporal difference-pooled spatial models (Sec. \ref{ssec:temporal_features}), \textit{CNN} represents deep features (Sec. \ref{ssec:deep_learning_features}), and \textit{Temp} presents the Haar bandpass-filtered features introduced in Sec. \ref{ssec:temporal_features}.}
\label{fig:ablation}
\end{figure}

\subsection{Effects of Training Data Size}
\label{ssec:training_size}

To study the degree of performance variation by the compared algorithms, we vary the training-test splits from 10\% to 90\% of the content used for training, using the rest for testing on the composite combined set. As might be seen in Fig. \ref{fig:test_size}, the RAPIQUE model was able to already achieve better than 0.8 in PLCC provided only 50\% of the data for training. When compared to SOTA methods, although RAPIQUE was not observed to outperform VIDEVAL when the fraction of training data was less than $40\%$, it delivered improved performances relative to VIDEVAL and TLVQM as the proportion of training data was increased, as shown in Fig.~\ref{fig:test_size}. This suggests that RAPIQUE is very data-efficient, with the potential to achieve even better results when larger-scale datasets become available.

\begin{table}[!t]
\setlength{\tabcolsep}{3.5pt}
\renewcommand{\arraystretch}{1.1}
\centering
\caption{Performance of RAPIQUE combined with different deep learning features. RAPIQUE ((w/ ResNet-50) is the default version proposed in this paper.}
\label{table:fusion}
\begin{tabular}{llcccc}
\toprule
\textsc{Dataset}  & \textsc{Model} / \textsc{Metric} & SRCC$\uparrow$      & PLCC$\uparrow$     & RMSE$\downarrow$    \\
\hline\\[-1.em]
\multirow{4}{*}{KoNViD-1k} & 
RAPIQUE (w/ ResNet-50)  & \textbf{0.8031} & \textbf{0.8175} & \textbf{0.3623} \\
& RAPIQUE (w/ VGG-19) & 0.7554 & 0.7389 & 0.4238  \\
& RAPIQUE (w/ KonCept512) & 0.7802 & 0.7793 & 0.3975 \\
& RAPIQUE (w/ PaQ-2-PiQ)  & 0.7726 & 0.7672 & 0.4026 \\
\hline\\[-1.em]
\multirow{4}{*}{LIVE-VQC}   & 
RAPIQUE (w/ ResNet-50)  & \textbf{0.7548} & \textbf{0.7863} & \textbf{10.518} \\
& RAPIQUE (w/ VGG-19) & 0.6888 & 0.7048 & 12.228 \\
& RAPIQUE (w/ KonCept512) & 0.7497 & 0.7611 & 11.231 \\
& RAPIQUE (w/ PaQ-2-PiQ)  & 0.7147 & 0.7308 & 11.599 \\
\hline\\[-1.em]
\multirow{4}{*}{YouTube-UGC} & 
RAPIQUE (w/ ResNet-50)  & 0.7591 & \textbf{0.7684} & \textbf{0.4060} \\
& RAPIQUE (w/ VGG-19) & 0.7379 & 0.7398 & 0.4365 \\
& RAPIQUE (w/ KonCept512) & \textbf{0.7668} & 0.7678 & 0.4190 \\
& RAPIQUE (w/ PaQ-2-PiQ)  & 0.7596 & 0.7606 & 0.4200 \\

\hline\\[-1.em]
\multirow{4}{*}{All-Combined} &
RAPIQUE (w/ ResNet-50)  & \textbf{0.8070} & \textbf{0.8229} & \textbf{0.3968} \\
& RAPIQUE (w/ VGG-19) & 0.6888 & 0.7048 & 12.228 \\
& RAPIQUE (w/ KonCept512) & 0.7924 & 0.7976 & 0.4169 \\
& RAPIQUE (w/ PaQ-2-PiQ)  & 0.7742 & 0.7809 & 0.4312 \\

\toprule
\end{tabular}
\end{table}

\subsection{Ablation Study}
\label{ssec:ablation_study}

To analyze the importance of each module in RAPIQUE, we conducted an ablation study. Fig. \ref{fig:ablation} shows the incremental performance attained when adding each module sequentially. It is worth mentioning the dataset biases of the evaluated benchmarks. For example, the authors of \cite{tu2020ugc} observed that the LIVE-VQC videos generally contain more (camera) motions and temporal distortions than other databases, while spatial distortions predominate on KoNViD-1k and YouTube-UGC. It may be observed in Fig.~\ref{fig:ablation}(a) that the spatial NSS module (Sec. \ref{ssec:spatial_features}) performs quite well on the UGC databases that mainly present spatial distortions, like KoNViD-1k and YouTube-UGC, indicating its efficacy in capturing authentic spatial distortions. LIVE-VQC, which mainly contains videos with large motions, challenges the spatial NSS module, aligning with the empirical observations made above \cite{tu2020ugc}. Adding spatial variation and temporal NSS features (Sec. \ref{ssec:temporal_features}) improves the performance of RAPIQUE on LIVE-VQC, indicating that these two types of temporal features capture important attributes of motion-intensive videos. Interestingly, we also noticed that including the SpatialNSS-Var features degraded performance on YouTube-UGC. It is possible that the SpatialNSS-Var features are redundant with SpatialNSS features on YouTube-UGC, causing the training algorithm to underperform. We also observed that temporal statistics did not contribute much to the assessment of Internet UGC videos from YouTube and KoNViD-1k (Flickr).

It is also important to note that including deep learning features (Sec. \ref{ssec:deep_learning_features}) significantly boosts the performance over only using NSS features on all these UGC datasets, further validating our assumptions expressed in Sec. \ref{ssec:deep_learning_bvqa_models}, that high-level semantic features are also informative when conducting UGC video quality prediction. To better understand which types of videos are advantageously analyzed by the CNN features, we divided the combined set into three subsets of differing contents: 2,667 natural videos, 163 screen contents, and 209 gaming videos, as shown in Fig.~\ref{fig:ablation}(b). Notably, we observed that the CNN features provided more benefits on screen content and gaming videos than on natural videos. The new temporal statistical features yielded noticeable improvements relative to using only spatial features. Lastly, our deployment of CNN modules is essentially different from other methods \cite{hosu2020koniq, ying2019patches} in that RAPIQUE only requires a single pass of the resized frames (224x224), making it highly advantageous in application scenarios having high-speed requirements.

\begin{table}[!t]
\setlength{\tabcolsep}{10pt}
\renewcommand{\arraystretch}{1.}
\centering\footnotesize
\caption{Feature dimensionality and average (CPU/GPU) runtime comparison (in seconds) evaluated on $1080p$ videos.}
\label{table:complexity}
\begin{tabular}{lrrr}
\toprule
\textsc{Model} & \textsc{Dim} & \multicolumn{2}{c}{\textsc{Runtime}} \\\cline{3-4}\\[-1.em]
& & CPU & GPU \\
\hline\\[-1.em]
BRISQUE (1 fr/sec) & 36 & 1.7 & - \\
GM-LOG (1 fr/sec) & 40 & 2.1 & - \\
HIGRADE (1 fr/sec) & 216 & 11.6 & - \\
FRIQUEE (1 fr/sec) & 560 & 701.2 & - \\
CORNIA (1 fr/sec) &  10k & 14.3 & - \\
HOSA (1 fr/sec) &  14.7k & 1.2 & -  \\
KonCept512 (1 fr/sec) & - & 2.8 & 0.3 \\
PaQ-2-PiQ (1 fr/sec) & - & 6.9 & 0.8 \\
\hline\\[-1.em]
V-BLIINDS & 47 & 1989.9 & - \\
V-MEON & - & 16.4  & 2.6 \\
TLVQM & 75 & 183.8 & - \\
VSFA & - & 1288.7 & 157.9 \\
MDVSFA & - & 1319.4 & 162.5 \\
VIDEVAL & 60 & 305.8 & - \\
\hline\\[-1.em]
RAPIQUE (proposed) & 3.8k & 17.3 & - \\
\bottomrule
\end{tabular}
\end{table}

\subsection{Performance on different deep features}
\label{ssec:perf-deep}

To determine which kinds of deep features most effectively complement the proposed NSS features, we conducted another ablation study. We compared the performance of RAPIQUE variants that use features from different backbones: VGG-19, ResNet-50, PaQ-2-PiQ (trained on LIVE-FB \cite{ying2019patches}), and KonCept512 (trained on KonIQ-10k \cite{hosu2020koniq}). Since PaQ-2-PiQ was designed for local quality prediction, we included the predicted $3\times 5$ local quality scores along with the single global score. For KonCept512, the 256-dim feature vector immediately before the last linear layer in the fully connected head was included. We also included VGG-19 and ResNet-50, except for they were pre-trained on ImageNet classification.

\begin{table}[!t]
\setlength{\tabcolsep}{10pt}
\renewcommand{\arraystretch}{1.}
\centering\footnotesize
\caption{Complexity analysis of RAPIQUE. Tabulated values reflect the partial time devoted to each sub-component in RAPIQUE.}
\label{table:partial_complexity}
\begin{tabular}{lrrc}
\toprule
\textsc{Module} & \textsc{Dim} & \textsc{Runtime} \\
\hline\\[-1.em]
SpatialNSS (Sec. \ref{ssec:spatial_features})  & 680 & \multirow{2}{*}{11.1} \\
SpatialNSS-Var (Sec. \ref{ssec:temporal_features}) & 680  \\
TemporalNSS (Sec. \ref{ssec:temporal_features}) & 476 & 5.8 \\
CNN (Sec. \ref{ssec:deep_learning_features}) & 2.0k & 0.4 \\
\hline\\[-1.em]
RAPIQUE (Full model) & 3.8k & 17.3 \\
\bottomrule
\end{tabular}
\end{table}

The overall performance results are tabulated in Table~\ref{table:fusion}. It may be observed that combining NSS features with ResNet-50 yielded the best or top performances on all benchmarks, slightly better than KonCept512, suggesting that features pre-trained on classification tasks provide valuable high-level semantic information to the quality assessment process. Moreover, using features pre-trained on a specific IQA dataset may limit model generalizability to future, unseen distortions. Another important reason why we prefer ResNet-50 over KonCept512 is the gigantic model size of the InceptionResNetV2 \cite{szegedy2017inception}, used as the backbone of KonCept512.

\begin{figure}[!t]
\centering
\includegraphics[width=0.41\textwidth]{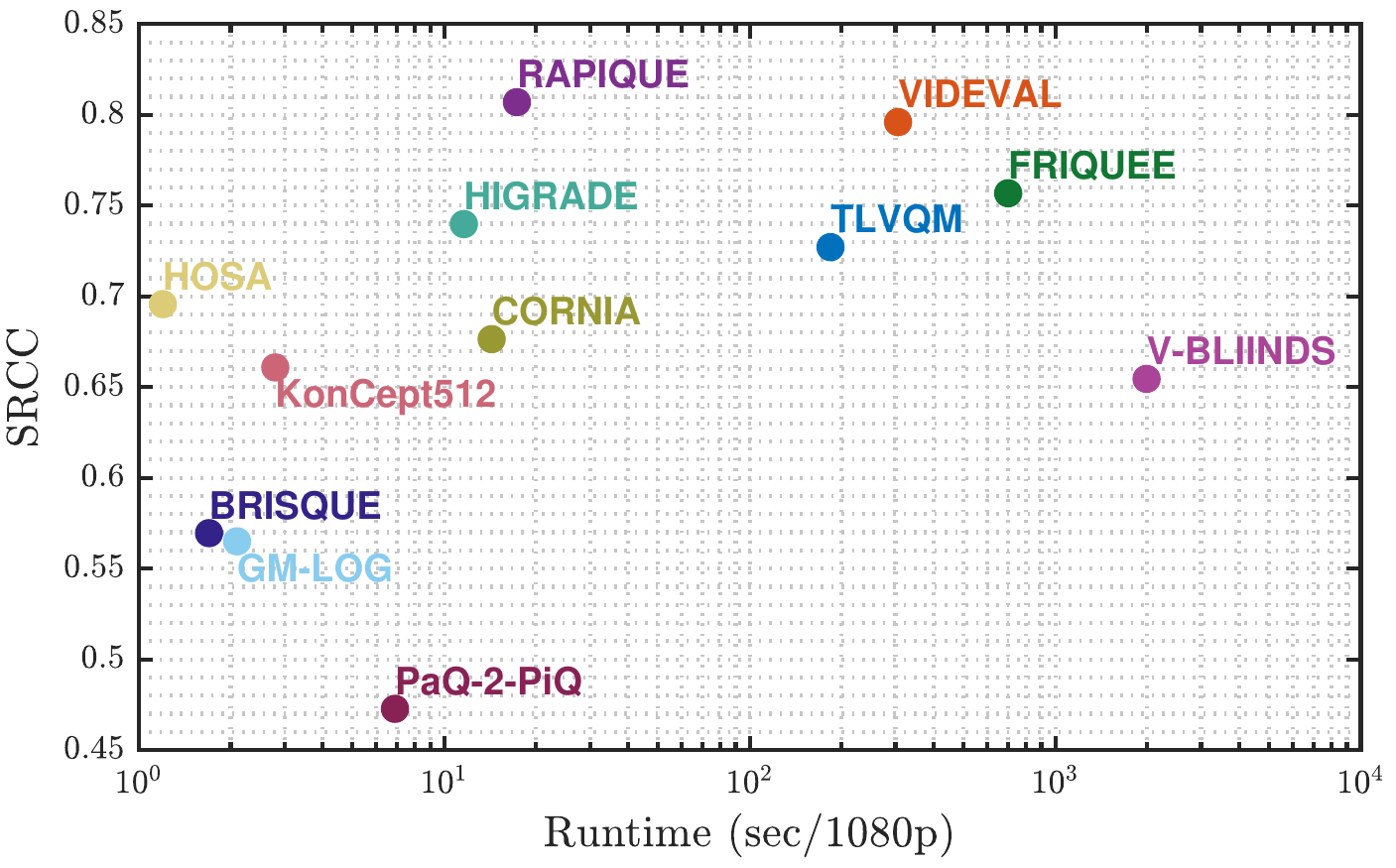}
\caption{Scatter plots of SRCC (on All-Combined) of selected BVQA algorithms versus CPU runtime (per 1080p video on average). Purple indicates the proposed RAPIQUE model.}
\label{fig:perf_n_speed}
\end{figure}

\begin{figure}[!t]
\centering
\includegraphics[width=0.4\textwidth]{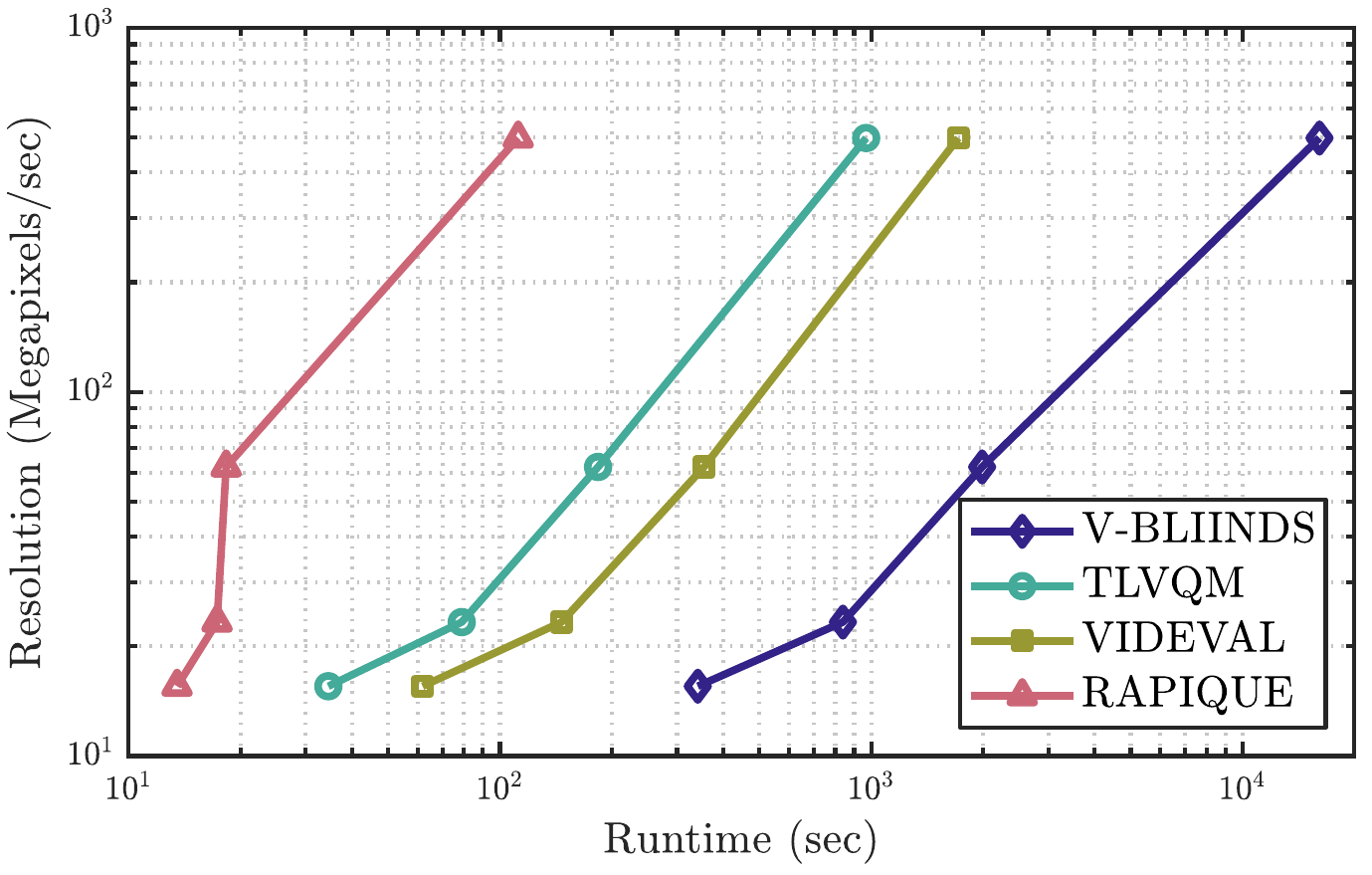}
\caption{Our proposed RAPIQUE model enables high-resolution video quality prediction at significantly lower runtimes than existing BVQA methods. Particularly, as
seen in the plot our model is 2-150x faster than baselines, depending on resolution, and the higher, the faster.}
\label{fig:speed_scales}
\end{figure}

\subsection{Complexity and Runtime Comparison}
\label{ssec:runtime_comparison}

Apart from performance analysis, computational efficiency is also of great importance for BVQA models. Thus, we also study the model (feature) dimension and runtime comparisons in Table \ref{table:complexity}. For a fair comparison, all the experiments were carried out in the same desktop computer, a Dell OptiPlex 7080 Desktop with Intel Core i7-8700 CPU@3.2GHz, 32G RAM, and GeForce GTX 1050 Graphics Cards. The models were implemented using their original releases on MATLAB R2018b and Python 3.6.7 under Ubuntu 18.04.3 LTS system. It should be noted that our comparison of computing complexity involves methods that use different sampling rate (or FPS), which is critical to model efficiency. However, we regard the design of FPS itself as an important aspect of BVQA algorithms, and thus our comparison still provides insights on developing more efficient BVQA models.

It may be seen that RAPIQUE is extremely efficient as compared to other complex top-performing BVQA models like TLVQM and VIDEVAL. Specifically, RAPIQUE is \textbf{10x} faster than TLVQM, which also aims to efficiency. Fig. \ref{fig:perf_n_speed} shows the scatter plots of SRCC versus runtime, which indicates that RAPIQUE achieves comparable prediction accuracy, but with \textbf{20x} less computational expense as compared to VIDEVAL, the current SOTA model on the UGC-VQA problem \cite{tu2020ugc}. We observe however that, CNN models that benefit from optimized low-level implementations are generally faster than NSS models executed in MATLAB; we have observed a $\sim$10x speedup by switching from CPU to GPU on the CNN-based models, KonCept512, PaQ-2-PiQ, V-MEON, VSFA, and MDVSFA.

Predicting the quality of videos having multiple diverse resolutions is also a pressing problem, but has barely been discussed, since most video datasets only contain single-resolution contents. Thanks to the large-scale dataset, YouTube-UGC \cite{wang2019youtube}, which contains videos at five different resolutions, we were able to extend the complexity analysis to videos ranging from 540p to 4k, to study computational scalability with respect to video size. Fig. \ref{fig:speed_scales} compares computation time as a function of video resolution. We may observe that RAPIQUE has superior computational scalability in terms of data sizes, making it attractive and preferable for potential real-time, low-latency, and light-weight applications requiring high-resolution video inputs. Particularly, as seen in the plot our model is 2-150x faster than baselines, depending on resolution, and the higher, the faster. In Table \ref{table:partial_complexity} we list the partial compute time of each sub-module in RAPIQUE on 1080p videos. Since all of the NSS-based features are implemented in MATLAB, a high-level prototyping tool, we would expect further accelerations to be possible (by orders-of-magnitude) if implemented in low-level languages like C/C\texttt{++}, or GPU-friendly frameworks such as Tensorflow or PyTorch.

\section{Conclusion}
\label{conclusion}

We have proposed an effective and efficient model for predicting the subjective quality of user-generated videos, which we call the Rapid and Accurate Video Quality Evaluator (RAPIQUE). The model, for the first time, leverages a composite of spatio-temporal scene statistics features and deep CNN-based high-level features in a two-branch framework, then jointly learns a regressor head for video quality prediction. Within the model, we developed new spatial scene statistics models in an efficient way and further extended the overall model to include normalized temporal bandpass responses, yielding the first general efficacious temporal NSS model for UGC video quality problems. Experiments on recent large-scale UGC video databases show the superior accuracy and efficiency of the proposed model in that it achieves competitive or substantially higher accuracy than both SOTA conventional as well as deep learning video quality models. RAPIQUE is computationally less expensive by orders-of-magnitude than the most accurate benchmark methods and scales remarkably well with video resolution. To support reproducible research, an implementation of RAPIQUE is available on \url{https://github.com/vztu/RAPIQUE}. 






\bibliographystyle{IEEEtran}
\bibliography{refs}

\begin{thebibliography}{10}
\providecommand{\url}[1]{#1}
\csname url@samestyle\endcsname
\providecommand{\newblock}{\relax}
\providecommand{\bibinfo}[2]{#2}
\providecommand{\BIBentrySTDinterwordspacing}{\spaceskip=0pt\relax}
\providecommand{\BIBentryALTinterwordstretchfactor}{4}
\providecommand{\BIBentryALTinterwordspacing}{\spaceskip=\fontdimen2\font plus
\BIBentryALTinterwordstretchfactor\fontdimen3\font minus
  \fontdimen4\font\relax}
\providecommand{\BIBforeignlanguage}[2]{{%
\expandafter\ifx\csname l@#1\endcsname\relax
\typeout{** WARNING: IEEEtran.bst: No hyphenation pattern has been}%
\typeout{** loaded for the language `#1'. Using the pattern for}%
\typeout{** the default language instead.}%
\else
\language=\csname l@#1\endcsname
\fi
#2}}
\providecommand{\BIBdecl}{\relax}
\BIBdecl

\bibitem{seshadrinathan2010study}
K.~Seshadrinathan, R.~Soundararajan, A.~C. Bovik, and L.~K. Cormack, ``Study of
  subjective and objective quality assessment of video,'' \emph{IEEE Trans.
  Image Process.}, vol.~19, no.~6, pp. 1427--1441, 2010.

\bibitem{wang2004image}
Z.~Wang, A.~C. Bovik, H.~R. Sheikh, E.~P. Simoncelli \emph{et~al.}, ``Image
  quality assessment: From error visibility to structural similarity,''
  \emph{IEEE Trans. Image Process.}, vol.~13, no.~4, pp. 600--612, 2004.

\bibitem{li2016toward}
Z.~Li, A.~Aaron, I.~Katsavounidis, A.~Moorthy, and M.~Manohara, ``Toward a
  practical perceptual video quality metric,'' \emph{The Netflix Tech Blog},
  vol.~6, p.~2, 2016.

\bibitem{yu2020predicting}
X.~Yu, N.~Birkbeck, Y.~Wang, C.~G. Bampis, B.~Adsumilli, and A.~C. Bovik,
  ``Predicting the quality of compressed videos with pre-existing
  distortions,'' \emph{arXiv preprint arXiv:2004.02943}, 2020.

\bibitem{Wang2020}
Y.~Wang, H.~Talebi, F.~Yang, J.~G. Yim, N.~Birkbeck, B.~Adsumilli, and
  P.~Milanfar, ``Video transcoding optimization based on input perceptual
  quality,'' in \emph{Appl. Digital Image Process. {XLIII}}, A.~G. Tescher and
  T.~Ebrahimi, Eds., Aug. 2020.

\bibitem{tu2020ugc}
Z.~Tu, Y.~Wang, N.~Birkbeck, B.~Adsumilli, and A.~C. Bovik, ``{UGC}-{VQA}:
  Benchmarking blind video quality assessment for user generated content,''
  \emph{IEEE Trans. Image Process.}, vol.~30, pp. 4449--4464, 2021.

\bibitem{moorthy2011blind}
A.~K. Moorthy and A.~C. Bovik, ``Blind image quality assessment: From natural
  scene statistics to perceptual quality,'' \emph{IEEE Trans. Image Process.},
  vol.~20, no.~12, pp. 3350--3364, 2011.

\bibitem{mittal2012no}
A.~Mittal, A.~K. Moorthy, and A.~C. Bovik, ``No-reference image quality
  assessment in the spatial domain,'' \emph{IEEE Trans. Image Process.},
  vol.~21, no.~12, pp. 4695--4708, 2012.

\bibitem{saad2014blind}
M.~A. Saad, A.~C. Bovik, and C.~Charrier, ``Blind prediction of natural video
  quality,'' \emph{IEEE Trans. Image Process.}, vol.~23, no.~3, pp. 1352--1365,
  2014.

\bibitem{kundu2017no}
D.~Kundu, D.~Ghadiyaram, A.~C. Bovik, and B.~L. Evans, ``No-reference quality
  assessment of tone-mapped {HDR} pictures,'' \emph{IEEE Trans. Image
  Process.}, vol.~26, no.~6, pp. 2957--2971, 2017.

\bibitem{xue2014blind}
W.~Xue, X.~Mou, L.~Zhang, A.~C. Bovik, and X.~Feng, ``Blind image quality
  assessment using joint statistics of gradient magnitude and laplacian
  features,'' \emph{IEEE Trans. Image Process.}, vol.~23, no.~11, pp.
  4850--4862, 2014.

\bibitem{ghadiyaram2017perceptual}
D.~Ghadiyaram and A.~C. Bovik, ``Perceptual quality prediction on authentically
  distorted images using a bag of features approach,'' \emph{J. Vision},
  vol.~17, no.~1, pp. 32--32, 2017.

\bibitem{korhonen2019two}
J.~Korhonen, ``Two-level approach for no-reference consumer video quality
  assessment,'' \emph{IEEE Trans. Image Process.}, vol.~28, no.~12, pp.
  5923--5938, 2019.

\bibitem{ye2012unsupervised}
P.~Ye, J.~Kumar, L.~Kang, and D.~Doermann, ``Unsupervised feature learning
  framework for no-reference image quality assessment,'' in \emph{Proc. IEEE
  Conf. Comput. Vis. Pattern Recognit. (CVPR)}, 2012, pp. 1098--1105.

\bibitem{pei2015image}
S.-C. Pei and L.-H. Chen, ``Image quality assessment using human visual {DOG}
  model fused with random forest,'' \emph{IEEE Trans. Image Process.}, vol.~24,
  no.~11, pp. 3282--3292, 2015.

\bibitem{ebenezer2020no}
J.~P. Ebenezer, Z.~Shang, Y.~Wu, H.~Wei, and A.~C. Bovik, ``No-reference video
  quality assessment using space-time chips,'' \emph{arXiv preprint
  arXiv:2008.00031}, 2020.

\bibitem{ying2019patches}
Z.~Ying, H.~Niu, P.~Gupta, D.~Mahajan, D.~Ghadiyaram, and A.~Bovik, ``From
  patches to pictures ({PaQ}-2-{PiQ}): Mapping the perceptual space of picture
  quality,'' in \emph{Proc. IEEE Conf. Comput. Vis. Pattern Recognit. (CVPR)},
  2020.

\bibitem{ying2020patch}
Z.~Ying, M.~Mandal, D.~Ghadiyaram, and A.~Bovik, ``Patch-{VQ}:'patching up'the
  video quality problem,'' \emph{arXiv preprint arXiv:2011.13544}, 2020.

\bibitem{li2019quality}
D.~Li, T.~Jiang, and M.~Jiang, ``Quality assessment of in-the-wild videos,'' in
  \emph{Proc. ACM Multimedia Conf. (MM)}, 2019, pp. 2351--2359.

\bibitem{feng2006measurement}
X.~Feng and J.~P. Allebach, ``Measurement of ringing artifacts in {JPEG}
  images,'' in \emph{Digit. Pub.}, vol. 6076, 2006, p. 60760A.

\bibitem{tu2020bband}
Z.~Tu, J.~Lin, Y.~Wang, B.~Adsumilli, and A.~C. Bovik, ``Bband index: a
  no-reference banding artifact predictor,'' in \emph{Proc. IEEE Int. Conf.
  Acoust., Speech, Signal Process. (ICASSP)}, 2020, pp. 2712--2716.

\bibitem{norkin2018film}
A.~Norkin and N.~Birkbeck, ``Film grain synthesis for {AV1} video codec,'' in
  \emph{Data Compress. Conf. (DCC)}, 2018, pp. 3--12.

\bibitem{chen2020proxiqa}
L.-H. Chen, C.~G. Bampis, Z.~Li, A.~Norkin, and A.~C. Bovik, ``{ProxIQA}: A
  proxy approach to perceptual optimization of learned image compression,''
  \emph{IEEE Trans. Image Process.}, vol.~30, pp. 360--373, 2020.

\bibitem{wang2019going}
H.~Wang, T.~Chen, Z.~Wang, and K.~Ma, ``I am going mad: Maximum discrepancy
  competition for comparing classifiers adaptively,'' in \emph{Int. Conf.
  Learn. Represent. (ICLR)}, 2019.

\bibitem{chen2020learning}
L.-H. Chen, C.~Bampis, Z.~Li, and A.~C. Bovik, ``Learning to distort images
  using generative adversarial networks,'' \emph{IEEE Signal Process. Lett.},
  2020.

\bibitem{kim2018deep}
W.~Kim, J.~Kim, S.~Ahn, J.~Kim, and S.~Lee, ``Deep video quality assessor: From
  spatio-temporal visual sensitivity to a convolutional neural aggregation
  network,'' in \emph{Proc. Eur. Conf. Comput. Vis. (ECCV)}, 2018, pp.
  219--234.

\bibitem{zhang2018blind}
Y.~Zhang, X.~Gao, L.~He, W.~Lu, and R.~He, ``Blind video quality assessment
  with weakly supervised learning and resampling strategy,'' \emph{IEEE Trans.
  Circuits Syst. Video Technol.}, vol.~29, no.~8, pp. 2244--2255, 2018.

\bibitem{lin2018koniq}
H.~Lin, V.~Hosu, and D.~Saupe, ``{KonIQ-10K}: Towards an ecologically valid and
  large-scale {IQA} database,'' \emph{arXiv preprint arXiv:1803.08489}, 2018.

\bibitem{ghadiyaram2015massive}
D.~Ghadiyaram and A.~C. Bovik, ``Massive online crowdsourced study of
  subjective and objective picture quality,'' \emph{IEEE Trans. Image
  Process.}, vol.~25, no.~1, pp. 372--387, 2015.

\bibitem{sinno2018large}
Z.~Sinno and A.~C. Bovik, ``Large-scale study of perceptual video quality,''
  \emph{IEEE Trans. Image Process.}, vol.~28, no.~2, pp. 612--627, 2018.

\bibitem{hosu2017konstanz}
V.~Hosu, F.~Hahn, M.~Jenadeleh, H.~Lin, H.~Men, T.~Szir{\'a}nyi, S.~Li, and
  D.~Saupe, ``The {Konstanz} natural video database ({KoNViD-1k}),'' in
  \emph{Proc. 9th Int. Conf. Qual. Multimedia Exper. (QoMEX)}, 2017, pp. 1--6.

\bibitem{wang2019youtube}
Y.~Wang, S.~Inguva, and B.~Adsumilli, ``{YouTube UGC} dataset for video
  compression research,'' \emph{arXiv preprint arXiv:1904.06457}, 2019.

\bibitem{mittal2012making}
A.~Mittal, R.~Soundararajan, and A.~C. Bovik, ``Making a “completely blind”
  image quality analyzer,'' \emph{IEEE Signal Process. Lett.}, vol.~20, no.~3,
  pp. 209--212, 2012.

\bibitem{he2016deep}
K.~He, X.~Zhang, S.~Ren, and J.~Sun, ``Deep residual learning for image
  recognition,'' in \emph{Proc. IEEE Conf. Comput. Vis. Pattern Recognit.
  (CVPR)}, 2016, pp. 770--778.

\bibitem{wang2000blind}
Z.~Wang, A.~C. Bovik, and B.~L. Evan, ``Blind measurement of blocking artifacts
  in images,'' in \emph{Proc. IEEE Int. Conf. Image Process. (ICIP)}, vol.~3,
  2000, pp. 981--984.

\bibitem{marziliano2002no}
P.~Marziliano, F.~Dufaux, S.~Winkler, and T.~Ebrahimi, ``A no-reference
  perceptual blur metric,'' in \emph{Proc. IEEE Int. Conf. Image Process.
  (ICIP)}, vol.~3, 2002, pp. III--III.

\bibitem{wang2016perceptual}
Y.~Wang, S.-U. Kum, C.~Chen, and A.~Kokaram, ``A perceptual visibility metric
  for banding artifacts,'' in \emph{Proc. IEEE Int. Conf. Image Process.
  (ICIP)}, 2016, pp. 2067--2071.

\bibitem{tu2020adaptive}
{Z. Tu, J. Lin, Y. Wang, B. Adsumilli, and A. C. Bovik}, ``Adaptive debanding
  filter,'' \emph{IEEE Signal Process. Lett.}, vol.~27, pp. 1715--1719, 2020.

\bibitem{amer2005fast}
A.~Amer and E.~Dubois, ``Fast and reliable structure-oriented video noise
  estimation,'' \emph{IEEE Trans. Circuits Syst. Video Technol.}, vol.~15,
  no.~1, pp. 113--118, 2005.

\bibitem{ruderman1994statistics}
D.~L. Ruderman, ``The statistics of natural images,'' \emph{Network: Comput.
  Neural Syst.}, vol.~5, no.~4, pp. 517--548, 1994.

\bibitem{sheikh2006image}
H.~R. Sheikh and A.~C. Bovik, ``Image information and visual quality,''
  \emph{IEEE Trans. Image Process.}, vol.~15, no.~2, pp. 430--444, 2006.

\bibitem{moorthy2010two}
A.~K. Moorthy and A.~C. Bovik, ``A two-step framework for constructing blind
  image quality indices,'' \emph{IEEE Signal Process. Lett.}, vol.~17, no.~5,
  pp. 513--516, 2010.

\bibitem{zhang2014c}
Y.~Zhang, A.~K. Moorthy, D.~M. Chandler, and A.~C. Bovik, ``{C-DIIVINE}:
  No-reference image quality assessment based on local magnitude and phase
  statistics of natural scenes,'' \emph{Signal Process.: Image Commun.},
  vol.~29, no.~7, pp. 725--747, 2014.

\bibitem{saad2010dct}
M.~A. Saad, A.~C. Bovik, and C.~Charrier, ``A {DCT} statistics-based blind
  image quality index,'' \emph{IEEE Signal Process. Lett.}, vol.~17, no.~6, pp.
  583--586, 2010.

\bibitem{saad2012blind}
------, ``Blind image quality assessment: A natural scene statistics approach
  in the dct domain,'' \emph{IEEE Trans. Image Process.}, vol.~21, no.~8, pp.
  3339--3352, 2012.

\bibitem{li2016spatiotemporal}
X.~Li, Q.~Guo, and X.~Lu, ``Spatiotemporal statistics for video quality
  assessment,'' \emph{IEEE Trans. Image Process.}, vol.~25, no.~7, pp.
  3329--3342, 2016.

\bibitem{mittal2015completely}
A.~Mittal, M.~A. Saad, and A.~C. Bovik, ``A completely blind video integrity
  oracle,'' \emph{IEEE Trans. Image Process.}, vol.~25, no.~1, pp. 289--300,
  2015.

\bibitem{sinno2019spatio}
Z.~Sinno and A.~C. Bovik, ``Spatio-temporal measures of naturalness,'' in
  \emph{Proc. IEEE Int. Conf. Image Process. (ICIP)}, 2019, pp. 1750--1754.

\bibitem{zhang2013no}
Y.~Zhang and D.~M. Chandler, ``No-reference image quality assessment based on
  log-derivative statistics of natural scenes,'' \emph{J. Electron. Imag.},
  vol.~22, no.~4, p. 043025, 2013.

\bibitem{nuutinen2016cvd2014}
M.~Nuutinen, T.~Virtanen, M.~Vaahteranoksa, T.~Vuori, P.~Oittinen, and
  J.~H{\"a}kkinen, ``{CVD}2014—a database for evaluating no-reference video
  quality assessment algorithms,'' \emph{IEEE Trans. Image Process.}, vol.~25,
  no.~7, pp. 3073--3086, 2016.

\bibitem{hosu2020koniq}
V.~Hosu, H.~Lin, T.~Sziranyi, and D.~Saupe, ``Koniq-10k: An ecologically valid
  database for deep learning of blind image quality assessment,'' \emph{IEEE
  Trans. Image Process.}, vol.~29, pp. 4041--4056, 2020.

\bibitem{born2005structure}
R.~T. Born and D.~C. Bradley, ``Structure and function of visual area {MT},''
  \emph{Annu. Rev. Neurosci.}, vol.~28, pp. 157--189, 2005.

\bibitem{dendi2020no}
S.~V.~R. Dendi and S.~S. Channappayya, ``No-reference video quality assessment
  using natural spatiotemporal scene statistics,'' \emph{IEEE Trans. Image
  Process.}, vol.~29, pp. 5612--5624, 2020.

\bibitem{madhusudana2020st}
P.~C. Madhusudana, N.~Birkbeck, Y.~Wang, B.~Adsumilli, and A.~C. Bovik,
  ``{ST}-{GREED}: Space-time generalized entropic differences for frame rate
  dependent video quality prediction,'' \emph{arXiv preprint arXiv:2010.13715},
  2020.

\bibitem{chen2020chroma}
L.-H. {Chen}, C.~G. {Bampis}, Z.~{Li}, J.~{Sole}, and A.~C. {Bovik},
  ``Perceptual video quality prediction emphasizing chroma distortions,''
  \emph{IEEE Trans. Image Process.}, pp. 1--1, 2020.

\bibitem{lee2020ipas}
D.~Y. Lee, H.~Ko, J.~Kim, and A.~C. Bovik, ``Video quality model for space-time
  resolution adaptation,'' in \emph{Proc. IEEE Int. Conf. Image Process. Appl.
  Syst.}, 2020.

\bibitem{lee2020josa}
------, ``On the space-time statistics of motion pictures,'' \emph{J. Optical
  Society America A}, 2020, submitted.

\bibitem{jiang2019enlightengan}
Y.~Jiang, X.~Gong, D.~Liu, Y.~Cheng, C.~Fang, X.~Shen, J.~Yang, P.~Zhou, and
  Z.~Wang, ``{EnlightenGAN}: Deep light enhancement without paired
  supervision,'' \emph{arXiv preprint arXiv:1906.06972}, 2019.

\bibitem{kang2014convolutional}
L.~Kang, P.~Ye, Y.~Li, and D.~Doermann, ``Convolutional neural networks for
  no-reference image quality assessment,'' in \emph{Proc. IEEE Conf. Comput.
  Vis. Pattern Recognit. (CVPR)}, 2014, pp. 1733--1740.

\bibitem{bosse2016deep}
S.~Bosse, D.~Maniry, T.~Wiegand, and W.~Samek, ``A deep neural network for
  image quality assessment,'' in \emph{Proc. IEEE Int. Conf. Image Process.
  (ICIP)}, 2016, pp. 3773--3777.

\bibitem{kim2017deep}
J.~Kim, H.~Zeng, D.~Ghadiyaram, S.~Lee, L.~Zhang, and A.~C. Bovik, ``Deep
  convolutional neural models for picture-quality prediction: Challenges and
  solutions to data-driven image quality assessment,'' \emph{IEEE Signal
  Process. Mag.}, vol.~34, no.~6, pp. 130--141, 2017.

\bibitem{deng2009imagenet}
J.~Deng, W.~Dong, R.~Socher, L.-J. Li, K.~Li, and L.~Fei-Fei, ``Imagenet: A
  large-scale hierarchical image database,'' in \emph{Proc. IEEE Conf. Comput.
  Vis. Pattern Recognit. (CVPR)}, 2009, pp. 248--255.

\bibitem{sheikh2006statistical}
H.~R. Sheikh, M.~F. Sabir, and A.~C. Bovik, ``A statistical evaluation of
  recent full reference image quality assessment algorithms,'' \emph{IEEE
  Trans. Image Process.}, vol.~15, no.~11, pp. 3440--3451, 2006.

\bibitem{ponomarenko2013color}
N.~Ponomarenko, O.~Ieremeiev, V.~Lukin, K.~Egiazarian, L.~Jin, J.~Astola,
  B.~Vozel, K.~Chehdi, M.~Carli, F.~Battisti \emph{et~al.}, ``Color image
  database {TID2013}: Peculiarities and preliminary results,'' in \emph{Proc.
  4th Eur. Workshop Vis. Inf. Process. (EUVIP)}, 2013, pp. 106--111.

\bibitem{liu2018end}
W.~Liu, Z.~Duanmu, and Z.~Wang, ``End-to-end blind quality assessment of
  compressed videos using deep neural networks,'' in \emph{Proc. ACM Multimedia
  Conf. (MM)}, 2018, pp. 546--554.

\bibitem{ghadiyaram2017capture}
D.~Ghadiyaram, J.~Pan, A.~C. Bovik, A.~K. Moorthy, P.~Panda, and K.-C. Yang,
  ``In-capture mobile video distortions: A study of subjective behavior and
  objective algorithms,'' \emph{IEEE Trans. Circuits Syst. Video Technol.},
  vol.~28, no.~9, pp. 2061--2077, 2017.

\bibitem{li2020unified}
D.~Li, T.~Jiang, and M.~Jiang, ``Unified quality assessment of in-the-wild
  videos with mixed datasets training,'' \emph{Int. J. Comput. Vis.}, 2021.

\bibitem{vu2014vis3}
P.~V. Vu and D.~M. Chandler, ``{ViS3}: An algorithm for video quality
  assessment via analysis of spatial and spatiotemporal slices,'' \emph{J.
  Electron. Imag.}, vol.~23, no.~1, p. 013016, 2014.

\bibitem{sharifi1995estimation}
K.~Sharifi and A.~Leon-Garcia, ``Estimation of shape parameter for generalized
  gaussian distributions in subband decompositions of video,'' \emph{IEEE
  Trans. Circuits Syst. Video Technol.}, vol.~5, no.~1, pp. 52--56, 1995.

\bibitem{campbell1968application}
F.~W. Campbell and J.~G. Robson, ``Application of fourier analysis to the
  visibility of gratings,'' \emph{J. Physiology}, vol. 197, no.~3, p. 551,
  1968.

\bibitem{zhang2011fsim}
L.~Zhang, L.~Zhang, X.~Mou, and D.~Zhang, ``{FSIM}: A feature similarity index
  for image quality assessment,'' \emph{IEEE Trans. Image Process.}, vol.~20,
  no.~8, pp. 2378--2386, 2011.

\bibitem{zhang2015feature}
L.~Zhang, L.~Zhang, and A.~C. Bovik, ``A feature-enriched completely blind
  image quality evaluator,'' \emph{IEEE Trans. Image Process.}, vol.~24, no.~8,
  pp. 2579--2591, 2015.

\bibitem{rajashekar2010perceptual}
U.~Rajashekar, Z.~Wang, and E.~P. Simoncelli, ``Perceptual quality assessment
  of color images using adaptive signal representation,'' in \emph{Human Vis.
  Electron. Imaging XV}, vol. 7527.\hskip 1em plus 0.5em minus 0.4em\relax
  International Society for Optics and Photonics, 2010, p. 75271L.

\bibitem{lee2016toward}
D.~Lee and K.~N. Plataniotis, ``Toward a no-reference image quality assessment
  using statistics of perceptual color descriptors,'' \emph{IEEE Trans. Image
  Process.}, vol.~25, no.~8, pp. 3875--3889, 2016.

\bibitem{hasler2003measuring}
D.~Hasler and S.~E. Suesstrunk, ``Measuring colorfulness in natural images,''
  in \emph{Proc. SPIE Human Vis. Electron. Imag.}, vol. 5007, 2003, pp. 87--96.

\bibitem{wiki:CIELAB}
\BIBentryALTinterwordspacing
``Cielab color space --- {Wikipedia}{,} the free encyclopedia,'' [Accessed
  5-October-2020]. [Online]. Available:
  \url{https://en.wikipedia.org/wiki/CIELAB_color_space}
\BIBentrySTDinterwordspacing

\bibitem{wang2003multiscale}
Z.~Wang, E.~P. Simoncelli, and A.~C. Bovik, ``Multiscale structural similarity
  for image quality assessment,'' in \emph{Proc. IEEE Asilomar Conf. Signals,
  Syst. Comput.}, vol.~2, 2003, pp. 1398--1402.

\bibitem{soundararajan2012video}
R.~Soundararajan and A.~C. Bovik, ``Video quality assessment by reduced
  reference spatio-temporal entropic differencing,'' \emph{IEEE Trans. Circuits
  Syst. Video Technol.}, vol.~23, no.~4, pp. 684--694, 2012.

\bibitem{xue2013gradient}
W.~Xue, L.~Zhang, X.~Mou, and A.~C. Bovik, ``Gradient magnitude similarity
  deviation: A highly efficient perceptual image quality index,'' \emph{IEEE
  Trans. Image Process.}, vol.~23, no.~2, pp. 684--695, 2013.

\bibitem{tu2020comparative}
Z.~{Tu}, C.~J. {Chen}, L.~H. {Chen}, N.~{Birkbeck}, B.~{Adsumilli}, and A.~C.
  {Bovik}, ``A comparative evaluation of temporal pooling methods for blind
  video quality assessment,'' in \emph{Proc. IEEE Int. Conf. Image Process.
  (ICIP)}, 2020, pp. 141--145.

\bibitem{xu2016blind}
J.~Xu, P.~Ye, Q.~Li, H.~Du, Y.~Liu, and D.~Doermann, ``Blind image quality
  assessment based on high order statistics aggregation,'' \emph{IEEE Trans.
  Image Process.}, vol.~25, no.~9, pp. 4444--4457, 2016.

\bibitem{fang2020perceptual}
Y.~Fang, H.~Zhu, Y.~Zeng, K.~Ma, and Z.~Wang, ``Perceptual quality assessment
  of smartphone photography,'' in \emph{Proc. IEEE Conf. Comput. Vis. Pattern
  Recognit. (CVPR)}, 2020, pp. 3677--3686.

\bibitem{simonyan2014very}
K.~Simonyan and A.~Zisserman, ``Very deep convolutional networks for
  large-scale image recognition,'' \emph{arXiv preprint arXiv:1409.1556}, 2014.

\bibitem{he2015spatial}
K.~He, X.~Zhang, S.~Ren, and J.~Sun, ``Spatial pyramid pooling in deep
  convolutional networks for visual recognition,'' \emph{IEEE Trans. Pattern
  Anal. Mach. Intell.}, vol.~37, no.~9, pp. 1904--1916, 2015.

\bibitem{szegedy2017inception}
C.~Szegedy, S.~Ioffe, V.~Vanhoucke, and A.~Alemi, ``Inception-v4,
  inception-resnet and the impact of residual connections on learning,'' in
  \emph{Proc. AAAI Conf. Artificial Intelligence}, vol.~31, no.~1, 2017.

\end{thebibliography}

\end{document}